\documentclass[10pt,twocolumn,letterpaper]{article}

\usepackage{iccv}
\usepackage{times}
\usepackage{epsfig}
\usepackage{graphicx}
\usepackage{amsmath}
\usepackage{amssymb}

\usepackage{booktabs}
\usepackage{caption}
\usepackage{enumitem}
\usepackage[symbol]{footmisc}

\usepackage[breaklinks=true, bookmarks=false, pagebackref=true,colorlinks]{hyperref}

\iccvfinalcopy 

\def\iccvPaperID{****} 

\ificcvfinal\fi

\begin{document}

\title{Internal Video Inpainting by Implicit Long-range Propagation}

\author{Hao Ouyang\footnotemark[1] \qquad  Tengfei Wang\footnotemark[1] \qquad   Qifeng Chen\\

The Hong Kong University of Science and Technology\\

}

\ificcvfinal\thispagestyle{empty}\fi


\twocolumn[{
\renewcommand\twocolumn[1][]{#1}%
\maketitle
\begin{center}
\vspace{-2mm}
    \centering
\includegraphics[width=0.99\linewidth]{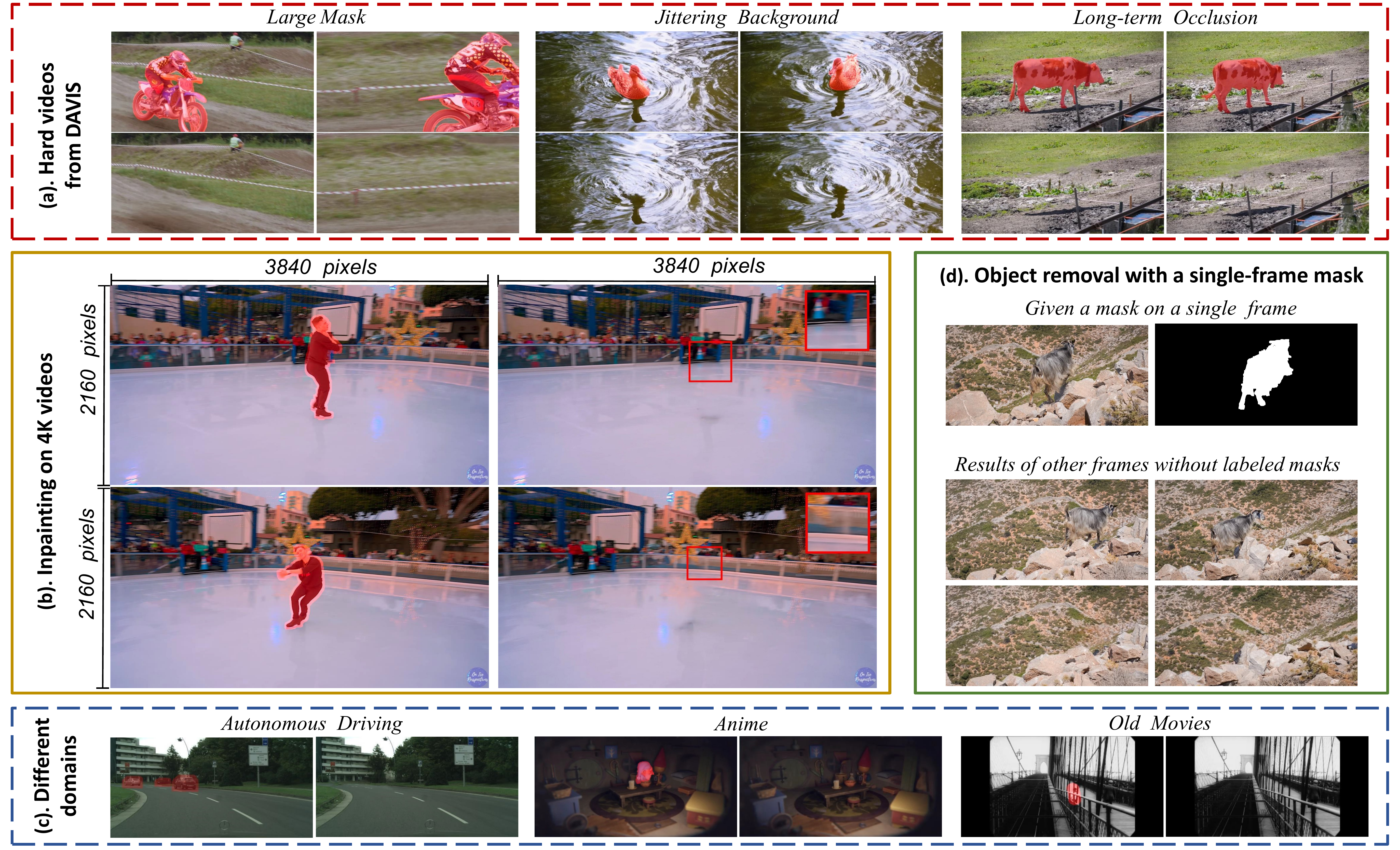}\\
\captionof{figure}{Ours results on (a) hard video sequences in DAVIS~\cite{perazzi2016benchmark}, (b) a 4K-resolution video,  (c) videos from different domains~\cite{cordts2016cityscapes}, and (d) a video with a single-frame mask. Zoom in for details. }
\label{fig:teaser}
\end{center}}]

 \footnotetext[1]{Equal contribution}
 \ificcvfinal\thispagestyle{empty}\fi
\begin{abstract}
 
We propose a novel framework for video inpainting by adopting an internal learning strategy. Unlike previous methods that use optical flow for cross-frame context propagation to inpaint unknown regions, we show that this can be achieved implicitly by fitting a convolutional neural network to known regions.  Moreover, to handle challenging sequences with ambiguous backgrounds or long-term occlusion, we design two regularization terms to preserve high-frequency details and long-term temporal consistency. Extensive experiments on the DAVIS dataset demonstrate that the proposed method achieves state-of-the-art inpainting quality quantitatively and qualitatively. We further extend the proposed method to another challenging task: learning to remove an object from a video giving a single object mask in only one frame in a 4K video. Our source code is available at \url{https://tengfei-wang.github.io/Implicit-Internal-Video-Inpainting/}.
\end{abstract}

\section{Introduction} \label{introduction}
Video inpainting is the problem of filling in missing regions in a video sequence with both spatial and temporal consistency. Video inpainting is beneficial for video editing, such as removing watermarks and unwanted objects. With the explosion of multimedia content in daily life, there are growing needs for inpainting sequences from multiple domains and real-world high-resolution videos. It is also expected to alleviate the human workload of labor-intensive mask labeling for semi-automatic object removal.

The video inpainting task is still unsolved yet because the existing approaches cannot consistently produce visually pleasing inpainted videos with long-range consistency. Most traditional methods~\cite{newson2014video,huang2016temporally, granados2012background, wexler2004space} adopt patch-based optimization strategies. These methods have limited ability to capture complex motion or synthesize new content. Recent flow-guided methods~\cite{gao2020flow, xu2019deep} propagate context information with the optical flow to achieve temporally consistent results. However, obtaining accurate optical flow in the missing region is non-trivial, especially when there is constantly blocked regions or the motion is complicated.  Recent deep models~\cite{wang2019video,kim2019deep,lee2019copy,oh2019onion,chang2019free,li2020short,zeng2020learning} trained on large video datasets achieve more promising performance. However, the dataset collection process is time-consuming and labor-intensive, and these methods may suffer from performance drop when test videos are in different domains from training videos. Most recently, Zhang et al.~\cite{zhang2019internal} propose an internal learning approach to video inpainting, which avoids the domain gap problem as the model training is completed on the test video.
While internal video inpainting is a promising direction, their approach sometimes generates incorrect or inconsistent results as this approach still depends on the externally-trained optical flow estimation to propagate context information.  Therefore, we propose a new internal learning method for video inpainting that can overcome the aforementioned issues with implicit long-range propagation, as shown in Fig.~\ref{fig:teaser}.

Instead of propagating information across frames via explicit correspondences like optical flow, we show that the information propagation process can be implicitly addressed by the intrinsic properties of natural videos and convolutional neural networks.  We will analyze these properties in detail and focus on handling two special challenging cases by imposing regularization. In the end, we manage to restore the missing region with cross-frame correlation and ensure temporal consistency by enforcing gradient constraints.  We train a convolutional neural network on the test video so that the trained model can propagate the information on known pixels (not in masks) to the whole video.

We first evaluate the proposed method on the DAVIS dataset. The proposed approach achieves  state-of-the-art performance both quantitatively and qualitatively, and the results of the user study indicate that our method is mostly preferred. We also apply our method to different video domains such as autonomous driving scenes, old films, and animations, and obtain promising results, as illustrated in Fig.~\ref{fig:teaser} (c). In addition, our formulation possesses great flexibility to extend to more challenging settings: 
1) A video with the mask on a single frame. We adopt a similar training strategy by switching the input and output in the above formulation to propagate masks.  2) Super high-resolution image sequences, such as a video with 4K resolution.  We design a progressive learning scheme, and the finer scale demands an additional prior, which is the output from the coarse-scale. We show examples of the extended method in  Fig.~\ref{fig:teaser} (b) and Fig.~\ref{fig:teaser} (d).

Our contribution can be summarized as follows:

\begin{itemize}[noitemsep,topsep=0pt]
  \item We propose an internal video inpainting method, which implicitly propagates information from the known regions to the unknown parts. Our method achieves state-of-the-art performance on  DAVIS  and can be applied to videos in various domains.
  \item We design two regularization terms  to address the ambiguity and deficiency problems for challenging video sequences. The anti-ambiguity term benefits the details generation, and the gradient regularization term reduces temporal inconsistency.
  \item To the best of our knowledge, our approach is the first deep internal learning model that demonstrates the feasibility of removing the objects in a 4K video with only a single frame mask.
\end{itemize}

\section{Related work} \label{relatedwork}

\noindent\textbf{Image inpainting.}
Traditional image inpainting methods, including the diffusion-based~\cite{ashikhmin2001synthesizing,ballester2001filling,esedoglu2002digital} or patch-based~\cite{barnes2009patchmatch} approaches, usually generate low-quality contents in complex scenes or with large inpainting masks. The recent development of deep learning has greatly improved inpainting performance. Pathak et al.~\cite{pathak2016context} are the first to   use the encoder-decoder network to extract features and reconstruct the missing region. The follow-up works~\cite{iizuka2017globally,yu2018generative,liu2018image,yu2019free} improve the network designs to handle the free-form mask and adopted a coarse-to-fine structure ~\cite{nazeri2019edgeconnect,li2019progressive,ren2019structureflow,wang2021external,zhang2020pixel} to use additional prior (e.g. edges and structures) as guidance.  Our work utilizes a similar generation structure with ~\cite{yu2019free} while adopts different training strategies and losses to improve the temporal consistency.

\begin{figure*}[t]
\centering
\vspace{-1.5mm}
\includegraphics[width=0.97\linewidth]{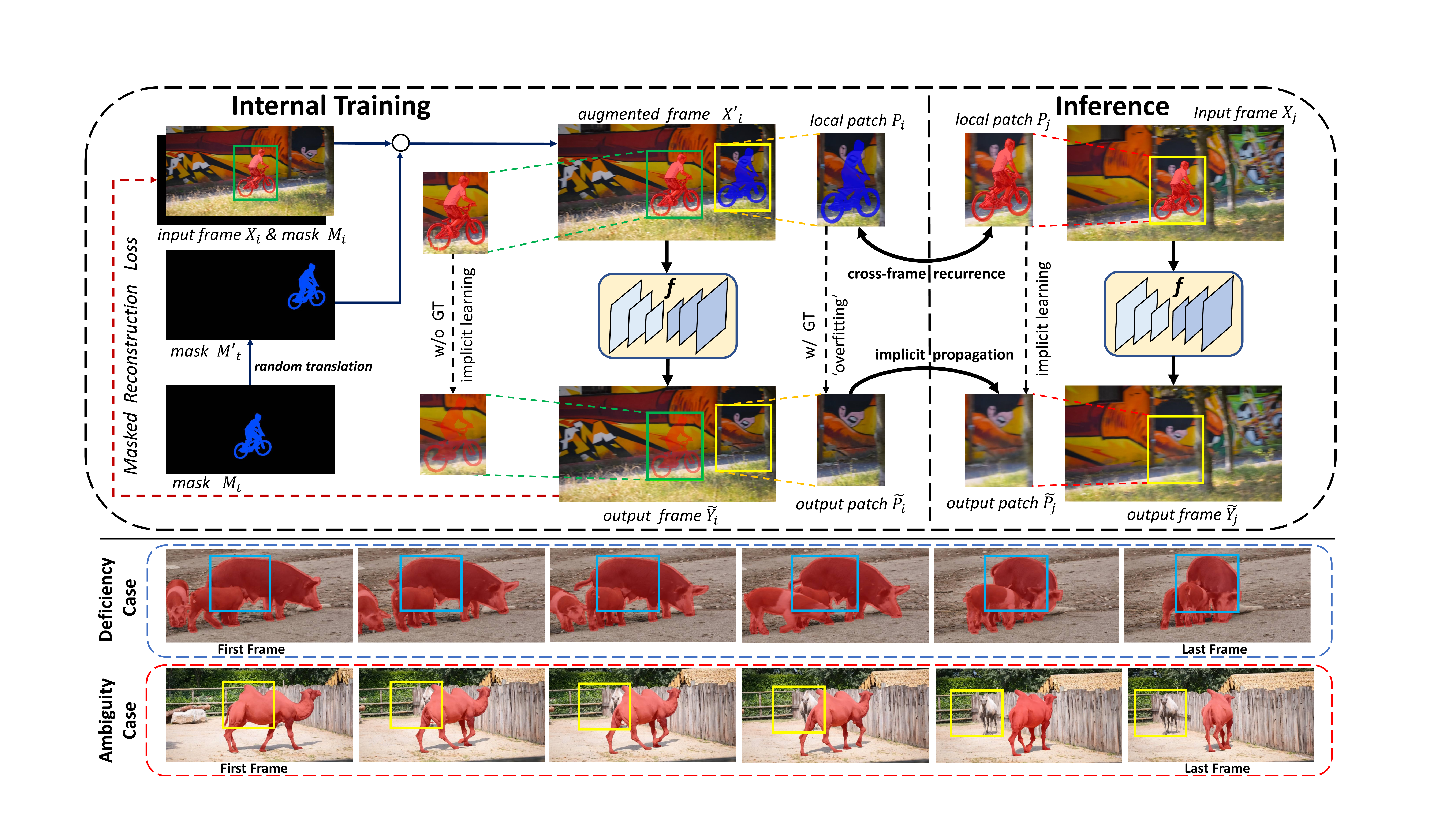}
\caption{\small{Overview of our internal video inpainting method. Without optical flow estimation and training on large datasets, we learn the implicit propagation via intrinsic properties of natural videos and neural network. By learning internally on augmented frames, the network $f$  serves as a neural memory function for long-range information. When inference, cross-frame contextual information is implicitly propagated to complete masked regions. For non-ideal cases of deficiency and ambiguity where cross-frame information is unavailable or ambiguous, we design two regularization terms for perceptually-realistic and temporally-consistent reconstruction.}}
\label{fig:formulation}
\vspace{-1.5mm}
\end{figure*}

\noindent\textbf{Video inpainting.}
Besides the spatial consistency, video inpainting involves another challenge, which is to ensure temporal consistency. Traditional methods usually complete regions by patch matching, including directly using 3D patches~\cite{wexler2004space} or 2D patches with explicit homography-based or optical-flow-based constraints ~\cite{newson2014video,huang2016temporally, granados2012background}. Most recent works  utilize deep convolutional neural networks trained on a large video dataset to learn how to collect information from the reference frames to generate the missing contents. Several works use 3D CNN structures~\cite{chang2019free, wang2019video} for feature extraction and content reconstruction but are extremely memory-consuming. Recent flow-based methods~\cite{gao2020flow, xu2019deep} propagate temporal information to fill in the regions by optical flow or flow field and fill the remaining pixels with pre-trained image inpainting models.  Other works utilize 2D CNN to fuse information of target frames and reference frames with different context aggregation modules~\cite{kim2019deep,lee2019copy,oh2019onion,xu2019deep,li2020short,zeng2020learning}. Instead of learning on a large dataset, our approach explores another direction of internal learning in a video sequence, which is flexible and can be applied to different video categories. 

\noindent\textbf{Deep internal learning.}
Deep internal learning is a recent trending topic that has demonstrated great potential in image manipulation and generation. Ulyanov et al.~\cite{ulyanov2018deep} are the first to utilize deep model as prior on several image restoration tasks, including image inpainting. Internal learning with deep models has then been applied in various tasks such as image super-resolution~\cite{shocher2018zero}, image generation~\cite{shocher2019ingan,shaham2019singan,bau2020semantic}, image inpainting~\cite{wang2021external},  video motion transfer~\cite{chan2019everybody} and video segmentation~\cite{gandelsman2019double}.  The most related paper to our work is the deep internal learning in video inpainting ~\cite{zhang2019internal} by jointing optimizing the generation of image and optical flow. However, their method relies heavily on the optical flow quality, which fails in large masks, slight motions, or incorrect frame selection. Our method adopts a totally different strategy by propagating the information from the known regions to the unknown regions implicitly by the neural network without optical flow and achieves more stable performance in the above-mentioned challenging cases.

\section{Method} \label{method}
\subsection{Problem formulation}
Video inpainting takes a sequence of video frames $X: \{X_1, X_2, ..., X_N\}$ with the corresponding masks $M: \{M_1, M_2, ..., M_N\}$ of the corrupted region or the undesired objects, where $N$ denotes the total number of the video frames. Our target is to generate the inpainted video $\Tilde{Y}:\{\Tilde{Y}_1, \Tilde{Y}_2, ..., \Tilde{Y}_N \}$ which is both spatially and temporally consistent. The known information can be represented as $\Bar{X}:\{X_1 \odot (1-M_1), X_2 \odot (1-M_2), ..., X_N \odot (1-M_N)\}$ and the quality of the inpainting depends profoundly on how we exploit the known information $\Bar{X}$. We observe three interesting and inspiring properties:

\begin{enumerate}[noitemsep,topsep=0pt]
  \item \textbf{Cross-frame recurrence} \label{itm:second}  Similar contents tend to appear multiple times in different frames in a video sequence. In Fig.~\ref{fig:formulation}, we use patch $P_i$ in frame  $X_i$ and $P_j$ in frame  $X_j$ to demonstrate this property. Note that $P_i$ and $P_j$ are not actually extracted in our pipeline but only to elaborate the properties. We denote the pixels of $P_i$ excluding the masked region (blue region) as $\Bar{P_i}$, the pixels of $P_j$ excluding the masked region (red region) as $\Bar{P_j}$ . The upper part of Fig.~\ref{fig:formulation} demonstrates an ideal case where the unmasked region $\Bar{P_j}$ in inference frame $X_j$ also occurs in frame $X_i$ where $\Bar{P_j} \approx \Bar{P_i}$. 
    
  \item \textbf{Neural network as a universal function} \label{itm:first} We can train a CNN to overfit to a certain amount of image data. With properly designed architecture, training an inpainting network $f$ that generates high-quality inpainting results on $\Bar{X}$ is achievable. As in Fig.~\ref{fig:formulation}, the  network $f$ generates nearly perfect content in the blue masked region in $\Tilde{P_i}$ by overfitting the ground-truth. 
  
  \item \textbf{Translational equivalent convolution} \label{itm:third}  Because of the weight sharing of convolution operation, if the input image of CNN is translated, the output of the network in each layer will translate in the same way~\cite{kondor2018generalization}.   This property indicates that if we regard $\Bar{P_j}$ as a translated $\Bar{P_i}$, the CNN $f$ is supposed to generate high-quality contents for $\Bar{P_j}$ as it for $\Bar{P_i}$. 
 
\end{enumerate}
 
Based on these properties, we propose a novel view on internal video inpainting: by learning how to inpaint the known region, the model implicitly learns how to inpaint the unknown region. Instead of explicitly find correspondence using optical flow, the propagation process is implicitly learned by using an augmented mask. The detailed training strategy is as follows: we train a neural network $f$, which takes three inputs $X_i$, $M_i$, and $M'_t$ in each training step, where $i$ and $t$ are randomly selected frame-index in the sequence. $M'_t$ is augmented by $M_t$ by random translation, and the input $X_i$ is masked by the binary union of $M_i$ and $M'_t$. Since the ground-truth pixels are not available in the unknown region masked by $M_i$, the reconstruction loss $\mathcal{L}_{rec}$ is only calculated in the known regions $\Bar{X}_i$:
\begin{align}
 \Tilde{Y}_i &=  f \left(  X_i\odot(1 - M_t'\cup M_i) \right),\\
 \mathcal{L}_{rec} &= \| (\Tilde{Y}_i - X_i) \odot (1 - M_i)\|_{1}.
\end{align}

 As suggested by Yu et al.~\cite{yu2018generative}, we replace vanilla convolution layers with gated convolution layers. The detailed architecture of network $f$ can be found in the \textbf{Appendix}.

\subsection{Ambiguity and deficiency analysis}
The solution with only the reconstruction loss  works well for ideal cases where all three properties are satisfied (e.g. `BMX-TREES' in DAVIS). However, in real applications, there are many exceptions. In Sec. \ref{ablation study}, we further explore Property \ref{itm:first} on the relationship between the model capacity and the complexity of video sequence, and Property  \ref{itm:third} on how the mask generation affects the inpainting quality. In this section, we focus on handling the violation cases on Property \ref{itm:second}: the cross-frame recurrence.

We first investigate the ideal cross-frame recurrence in a video sequence. As shown in Fig.~\ref{fig:formulation} upper parts, in the ideal case, there are consistent matches in the synthesized patches ($P_i$) and the inference patches ($P_j$). Two common non-ideal hard cases exist in the real video sequences: (a). There exist multiple unaligned matches and we call it ambiguity case. (b). There are no good matches and we call it deficiency case.  We propose two regularization terms to improve the performance in these difficult situations.

\subsection{Loss for ambiguity}
 
Ambiguity exists when conflict matching patches occur in multiple frames.  It usually appears when there are moving background objects or highly random textured regions in $\Bar{X}$.  Fig.~\ref{fig:formulation} demonstrates a sample ambiguous sequence `CAMEL' in  DAVIS. The background camel is raising his head while the foreground camel passes by. For the local patch $P_j$ in the first frame, there exist a set of patches $\{P_{i_1}, P_{i_2},...P_{i_n}\}$ with different contents (the position of camel face in this sequence). As expected, the predicted result of the moving region is blurry since it takes the average of all the ground truth, and we generate a blurry camel face shown in Fig.~\ref{fig:ambiguity_loss}.

To address the ambiguity issue, We propose an ambiguity regularization term that brings the details back in the later training stage. This term is inspired by recent work \cite{mechrez2018contextual}, which matches blurry source areas with nearest neighbors in target regions. Let $\{ s_p \}_{p\in P}$  and $\{ t_q \}_{q\in Q}$ denote feature points collections of source image $S$ and target image $T$ extracted by a shared encoder.  For each source point $s_p$, this loss tries to search for the most relevant target point under certain distance metric $\mathbb{D}_{(s_p,t_q)}$, and calculate $\delta_p(S, T)= \min _{q}\left(\mathbb{D}_{s_{p}, t_{q}}\right)$ \cite{mechrez2018contextual}. In our case, we focus on how to utilize the cross-frame correlation within a video. For a output frame $\Tilde{Y}_{i}$, we randomly select a frame $X_{t}$ as the target, and improve the high-frequency details by imposing the anti-ambiguity regularization:

\begin{equation}
\mathcal{L}_{\mathrm{ambiguity}} = \frac{1}{P} \sum_{p=1}^{P} \delta_p ( \Tilde{Y}_{i}\odot M_{t}, X_{t}\odot M_{t} ).
\end{equation}

\subsection{Loss for deficiency}
Deficiency situations exist when reliable cross-frame recurrence is not available.  It usually appears when the foreground objects only have slight movement, and a large part of the video is always occluded in all frames. We also show a sample sequence `PIGS' from DAVIS in Fig.~\ref{fig:formulation}, where a large region (e.g., the region in the blue mask) is constantly blocked. In this case, the network $f$ cannot fill in these regions by directly propagating similar cross-frame information. However, as the model $f$ is essentially a generative model learned on the internal video data, it can synthesize plausible new contents instead. The generated region follows the distribution of the known regions, which is acceptable in most cases. The quality depends on the complexity of the information (full results and the empirical analysis of extensive video sequences with deficient frames are presented in \textbf{Appendix}). As CNN $f$ becomes a pure single frame generation model on the constantly blocked regions, severe flickering artifacts appear when there are continuous deficient frames.

The main issue is how to improve temporal consistency when encountering the deficiency cases.  The recent works on improving the temporal consistency mostly depend on the pixel-to-pixel correspondence~\cite{lai2018learning, lei2020dvp, eilertsen2019single}, while in the inpainting task, there is no such correspondence. We observe that even though the changes in the input surrounding region are trivial (such as small rotation), the network $f$ can still generate a inconsistent result, which inspires us to apply constraints to the gradient of network $f$ \textit{w.r.t.} the input.  Following settings in ~\cite{eilertsen2019single}, suppose that the input $X_i, M_t', M_i$, go through a slight modification $g$, where $g$ is a random combination of homography transformation and image manipulation filters such as brightness change and blur kernel (when $g$ is applied to mask, the manipulation filters are ignored.). The new inpainted result $\Tilde{Y}_i'$ can be calculated as $f \{  g(X_i)\odot (1 - g(M_t')\cup g(M_i)) \}$. We expect the change of outputs is consistent with the change of inputs~\cite{eilertsen2019single}, and calculate the gradient difference as:
\begin{align}
 \Delta_{s} &= (\Tilde{Y}_i' - \Tilde{Y}_i) - (g(X_i) - X_i).
\end{align}
To minimize this term, we also need to exclude the pixels from unknown regions $M_i$ and $g(M_i)$:
\begin{align}
\mathcal{L}_{stabilize} &=\| \Delta_{s} \odot  (1 - M_i \cup g(M_i) )\|_{1} .
\end{align}

 The detailed ranges of each parameter used in $g$ is attached in the \textbf{Appendix}. The overall training loss is the weighted sum of $\mathcal{L}_{rec}$, $\mathcal{L}_{ambiguity}$, and $\mathcal{L}_{stablize}$.

\begin{figure}[t]
\centering
\hspace*{0.1mm}
\includegraphics[width=1.06\linewidth]{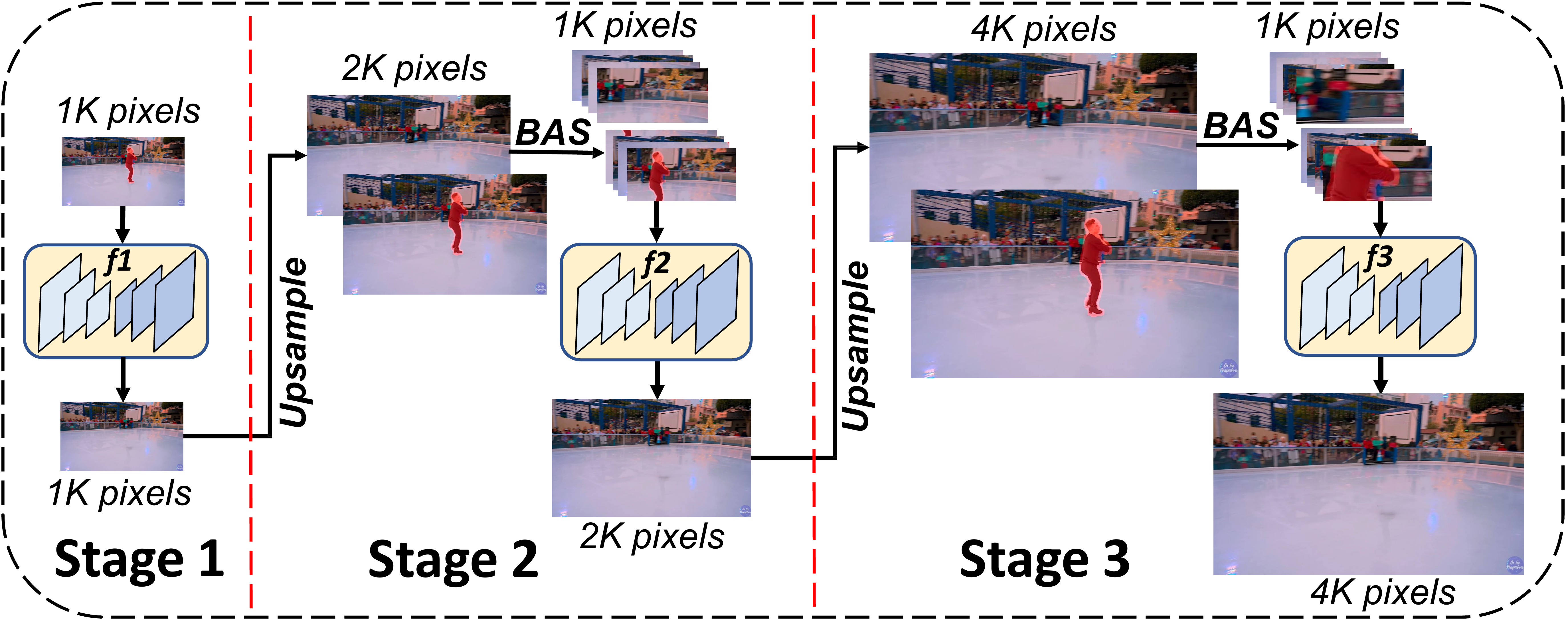}
\caption{ \small{The pipeline of our progressive learning strategy for high-resolution video inpainting.}}
\label{fig:multi-scale inpainting}
\end{figure}

\begin{figure}[t]
\centering
\includegraphics[width=0.98\linewidth]{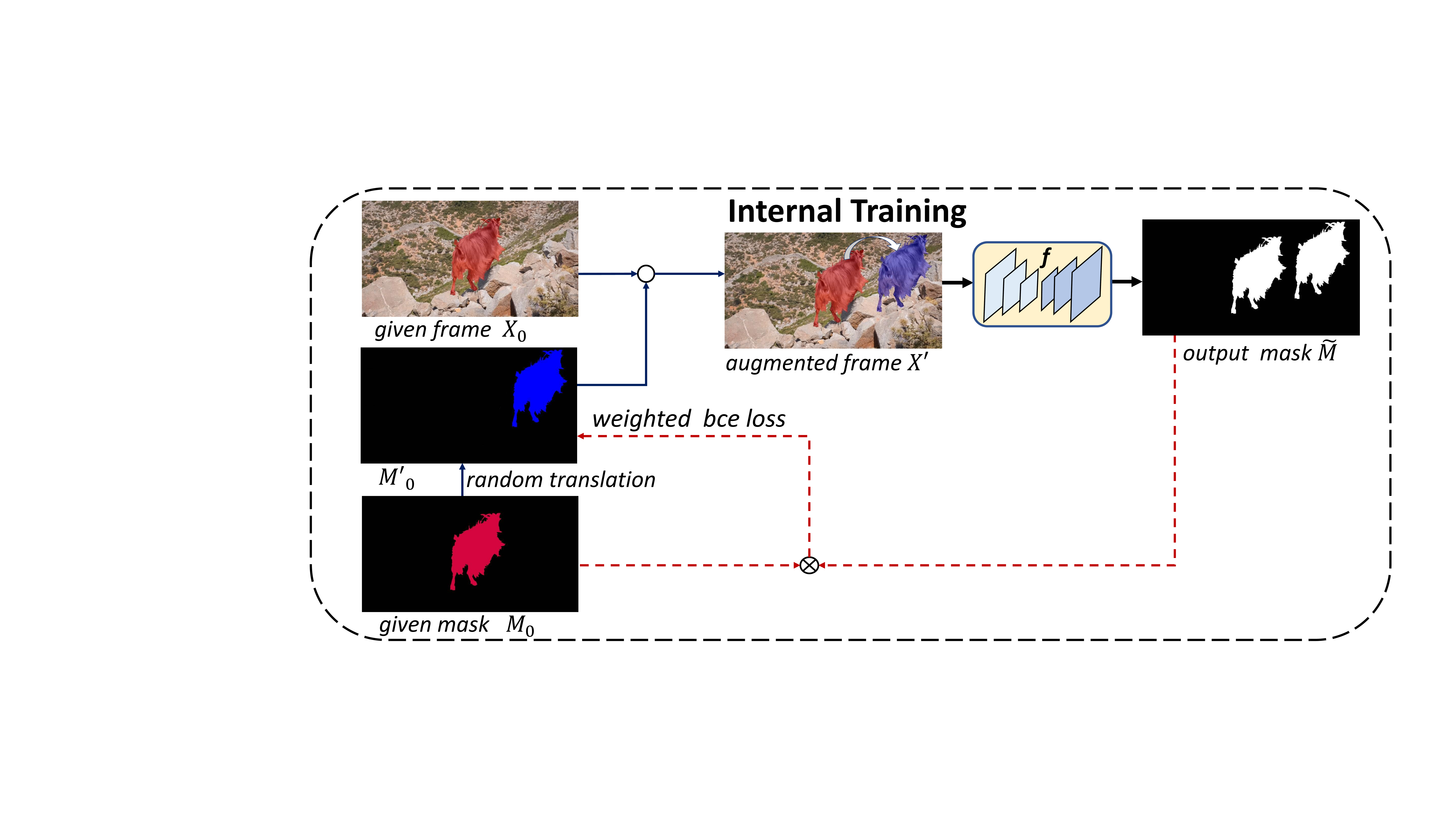}
\caption{\small The training process of our mask propagation. Given only one  mask, we can predict masks for the whole sequence.}
\label{fig:extension}
\end{figure}

\section{Extensions}  \label{extensions}

\subsection{Progressive high-resolution inpainting }
 While the original formulation can complete videos up to 1K resolution, we further extend it to a progressive scheme for inpainting high-resolution videos such as 2K or 4K videos, as illustrated in Fig.~\ref{fig:multi-scale inpainting}. Instead of overfitting the network to the full-resolution frames, we now overfit it using sampled patches. Two problems appear as we increase the resolution.  The surrounding pixels may provide limited information as the mask in one specific patch can be especially large. We thus use the upsampled inpainting result from the previous stage as an additional prior. Another issue is the low sampling efficiency, which may result in low training speed. We utilize a boundary-aware sampling (BAS) strategy, which samples more patches from regions around the object boundary. It is based on the fact that in most real-world object removal cases, the pixels around the boundary of the mask contain more valuable information and is of higher importance. More details, including the mask generation, grid-based inference, and BAS, are included in the \textbf{Appendix}.

\subsection{Video inpainting with a single frame mask}
Removing objects in video sequences using only a single frame mask is highly preferred to decrease the involved human labor. We show that by exchanging the input and the output in the above formulation,  the proposed scheme can also propagate a given object mask to other frames.
 
As shown in Fig.~\ref{fig:extension}, the input $X'$ is augmented using contextual information by randomly translating the undesired object on a single given frame $X_0$.  We calculate loss between the prediction mask $\Tilde{M}$ and the augmented mask $M'_0$. Based on the above-mentioned properties, the network $f$  learns to segment similar objects in other frames. Unlike the traditional reference-guided video mask propagation~\cite{oh2019videoseg, wug2018fast}, our final goal is not to detect the accurate mask but to remove the desired object. We thus enforce less penalty when we categorize the non-object pixel as object pixel than in the opposite direction as the incorrectly included background pixels can usually be filled in the following inpainting process. We adopt a weighted binary cross-entropy loss $\mathcal{L}_{we}$  which is calculated as $\mathcal{L}_{we}(y, \Tilde{y}) = \sum_{i}y_{i} \log \Tilde{y_{i}}+\alpha\left(1-y_{i}\right) \log \left(1-\Tilde{y_{i}}\right)$. Thus the reconstruction loss is formulated by
\begin{align}
 \mathcal{L}_{rec} =& -\mathcal{L}_{we}(M_0' \odot (1 -M_0), \Tilde{M} \odot (1 -M_0)),
\end{align}
where $M_0'$ and $X_0'$ are randomly translated by $M_0$ and $X_0$, and $\alpha$ is set to 0.8.

\section{Experiments} \label{experiments}
\subsection{Training details}
 For each video sequence in DAVIS~\cite{perazzi2016benchmark},  we first train the model with only reconstruction loss for about 60,000 iterations and then combine with the regularization loss for another 20,000 iterations for the hard sequences. We use Adam Optimizer with a learning rate of $2e-4$. The training process takes about 4 hours on a single NVIDIA RTX 2080 Ti GPU for a 80-frame video. The training settings of the extension tasks are reported in the  \textbf{Appendix}. 
\subsection{Qualitative results}
\begin{figure*}[t]
    \centering 
    \small
    \begin{tabular}{@{}c@{\hspace{0.3mm}}c@{\hspace{0.3mm}}c@{\hspace{0.3mm}}c@{\hspace{0.3mm}}c@{\hspace{0.3mm}}c@{}}

     \includegraphics[width=0.162\linewidth]{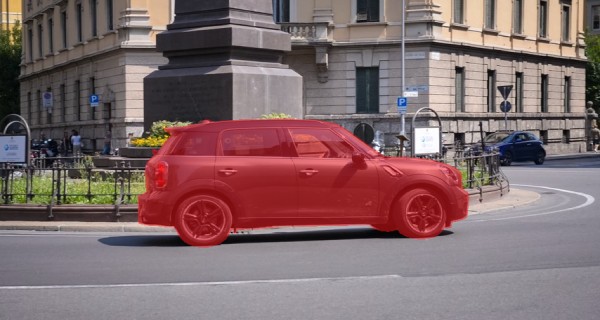}       &   
     \includegraphics[width=0.162\linewidth]{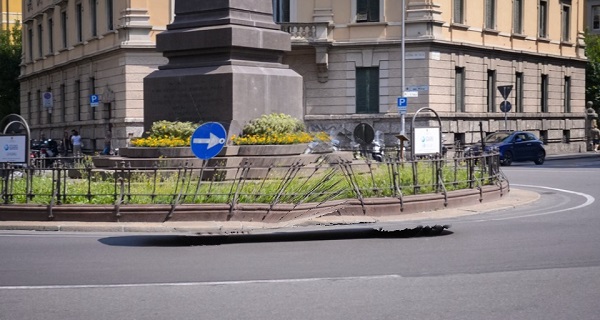}       &   
     \includegraphics[width=0.162\linewidth]{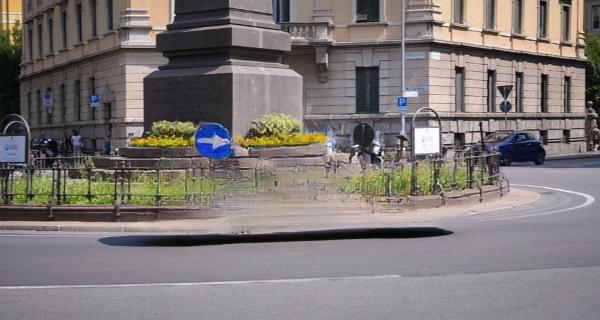}       &   
     \includegraphics[width=0.162\linewidth]{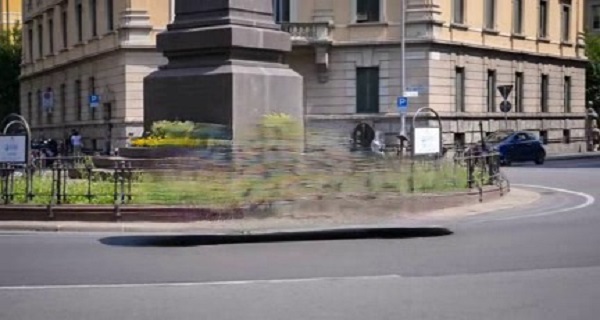}       &   
     \includegraphics[width=0.162\linewidth]{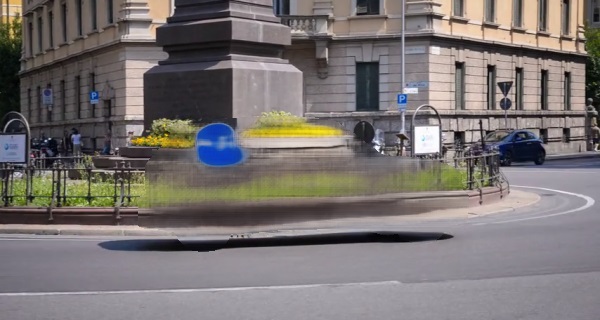}       &  
     \includegraphics[width=0.162\linewidth]{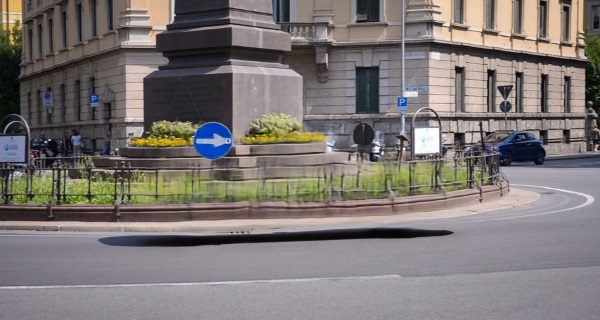}       \\

     \includegraphics[width=0.162\linewidth]{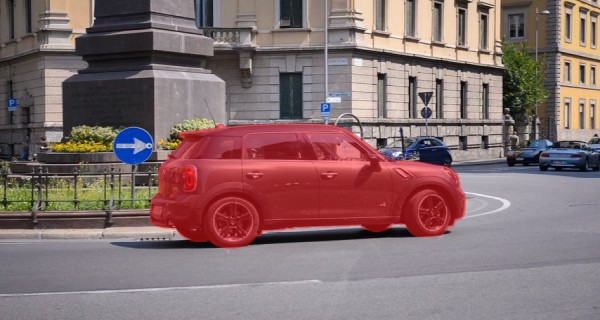}  &
     \includegraphics[width=0.162\linewidth]{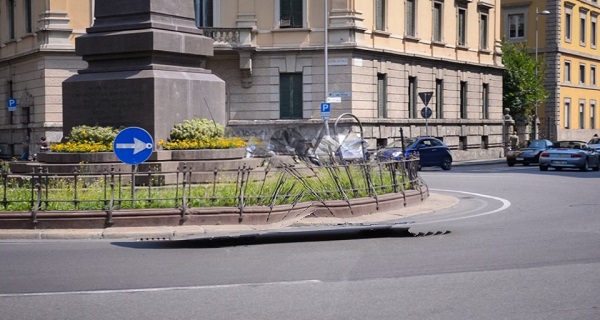}  &
     \includegraphics[width=0.162\linewidth]{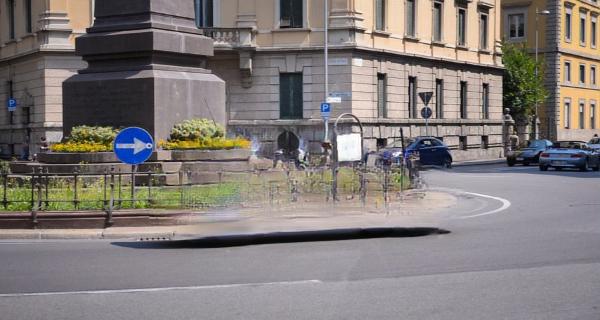}  &
     \includegraphics[width=0.162\linewidth]{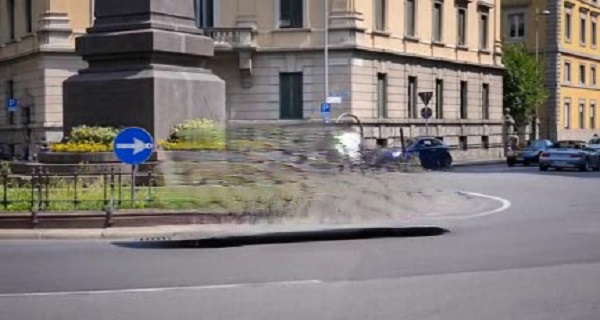}  &
     \includegraphics[width=0.162\linewidth]{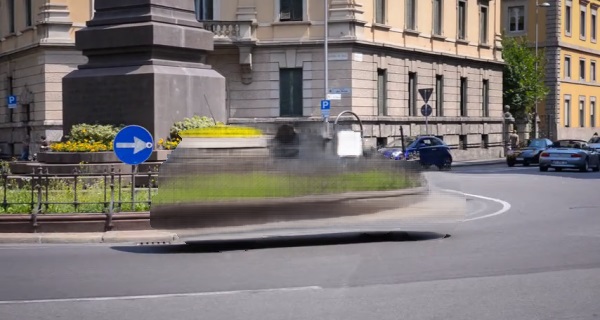}  &
     \includegraphics[width=0.162\linewidth]{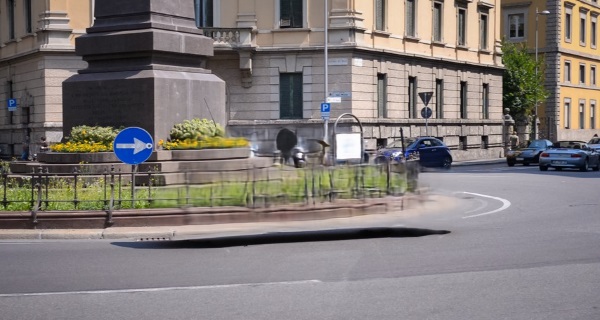} \\

     \includegraphics[width=0.162\linewidth]{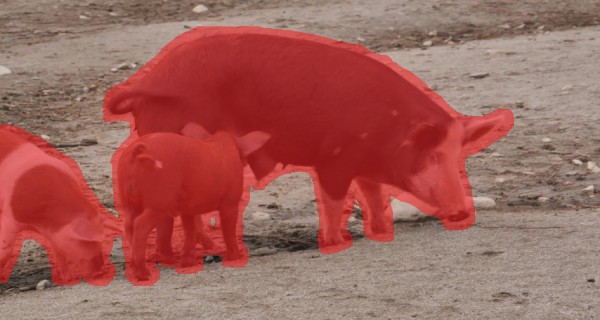} &      
     \includegraphics[width=0.162\linewidth]{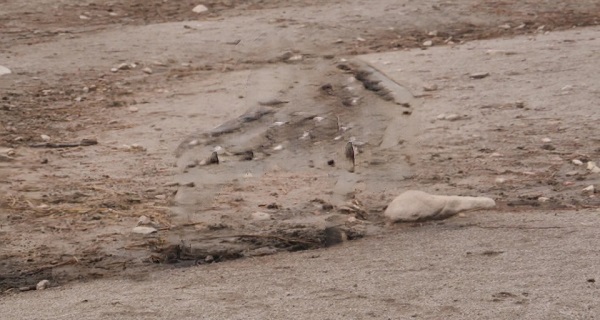} &    
     \includegraphics[width=0.162\linewidth]{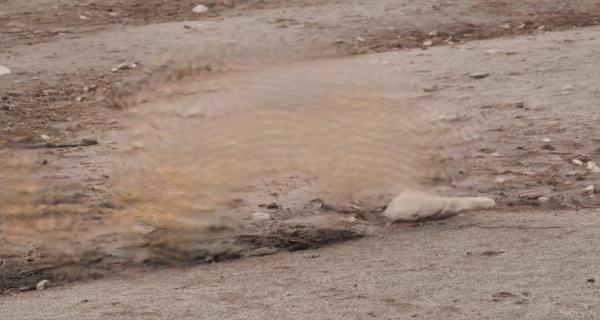} &       
     \includegraphics[width=0.162\linewidth]{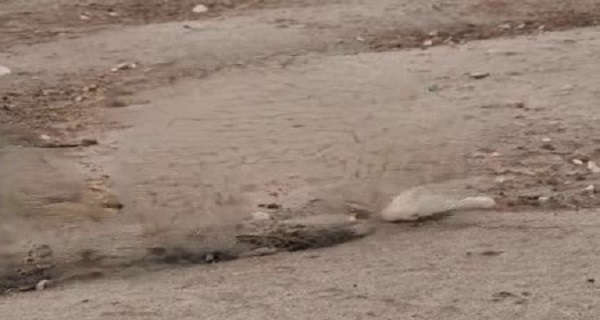} &     
     \includegraphics[width=0.162\linewidth]{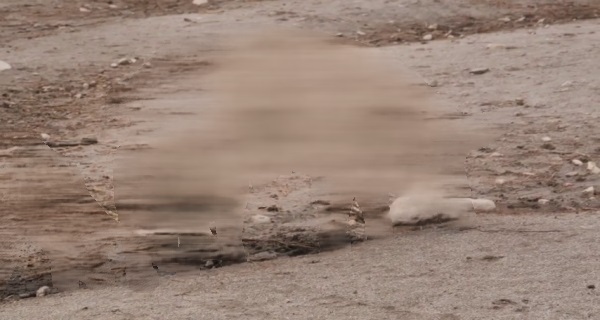}&      
     \includegraphics[width=0.162\linewidth]{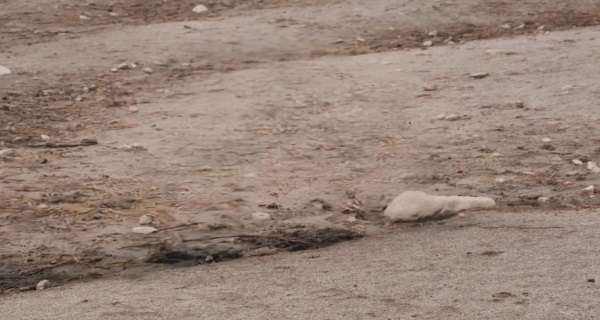} \\        
     
     \includegraphics[width=0.162\linewidth]{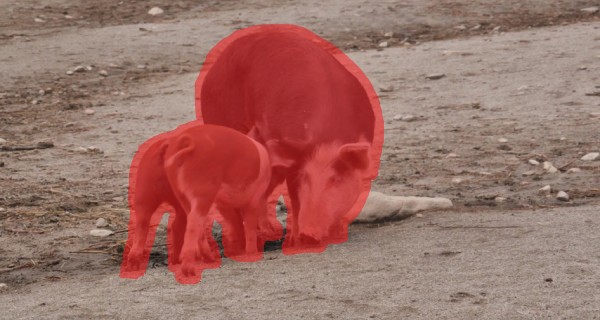} &      
     \includegraphics[width=0.162\linewidth]{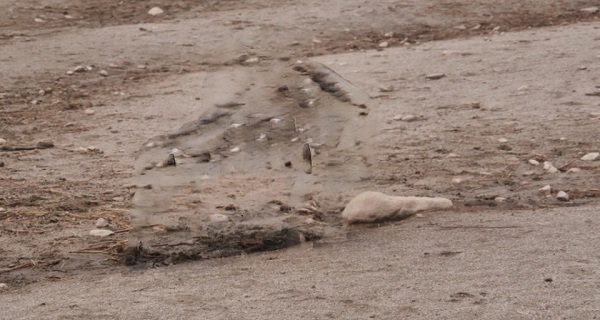} &    
     \includegraphics[width=0.162\linewidth]{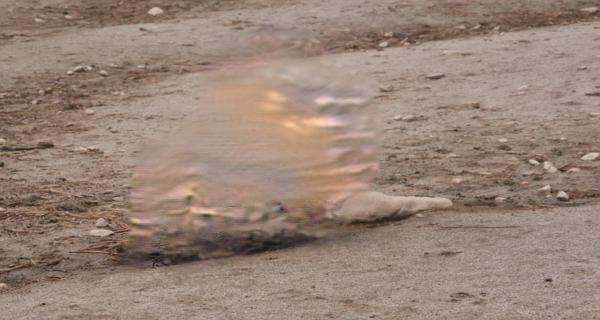} &       
     \includegraphics[width=0.162\linewidth]{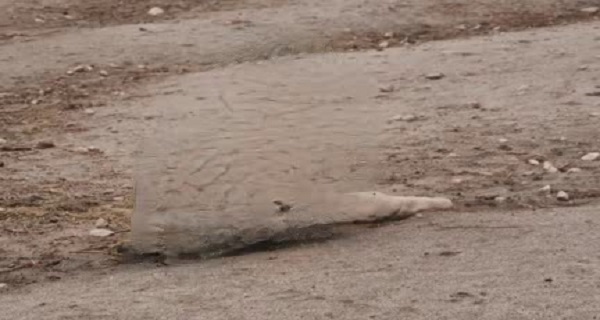} &     
     \includegraphics[width=0.162\linewidth]{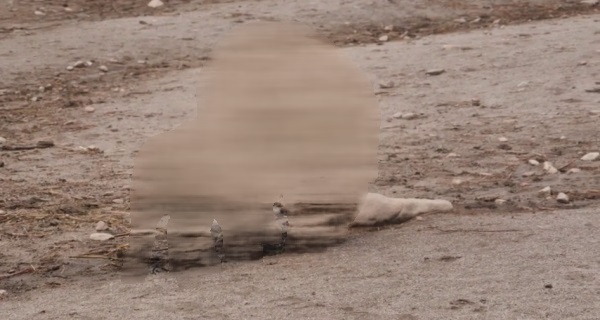}&      
     \includegraphics[width=0.162\linewidth]{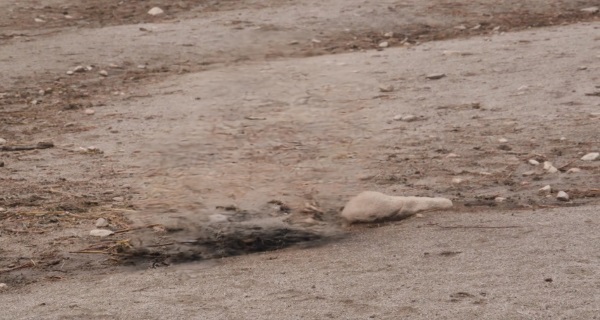} \\     

     \includegraphics[width=0.162\linewidth]{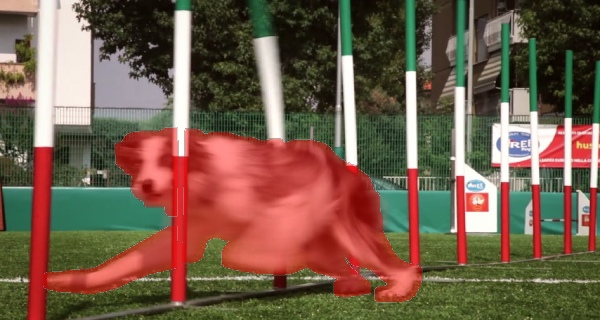} &      
     \includegraphics[width=0.162\linewidth]{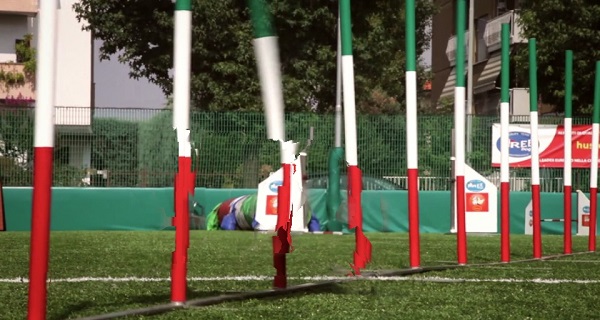} &    
     \includegraphics[width=0.162\linewidth]{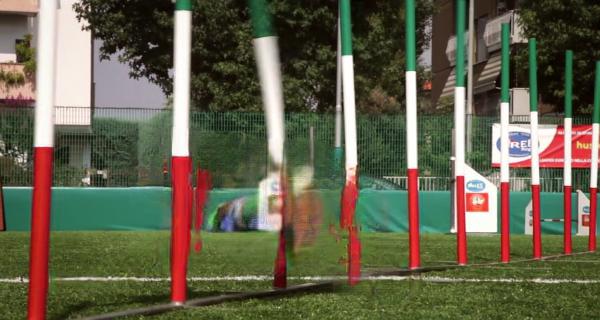} &       
     \includegraphics[width=0.162\linewidth]{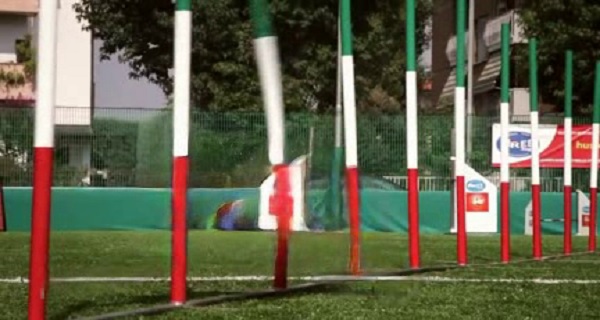} &     
     \includegraphics[width=0.162\linewidth]{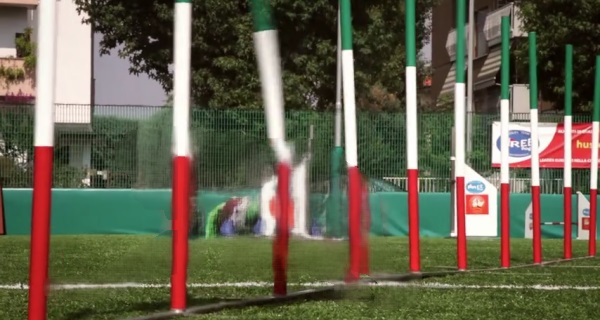}&      
     \includegraphics[width=0.162\linewidth]{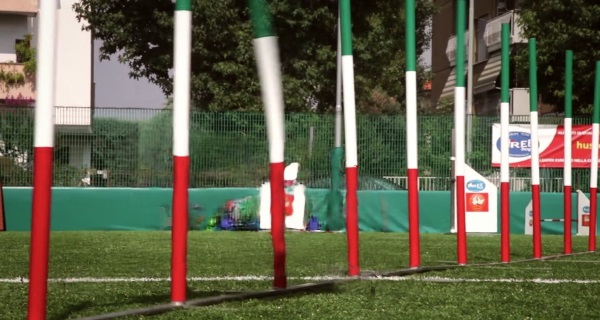} \\    
 
     \includegraphics[width=0.162\linewidth]{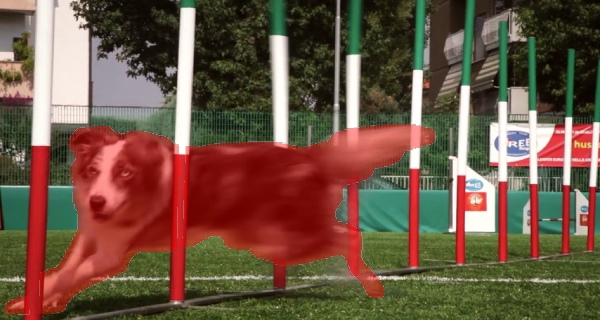} &      
     \includegraphics[width=0.162\linewidth]{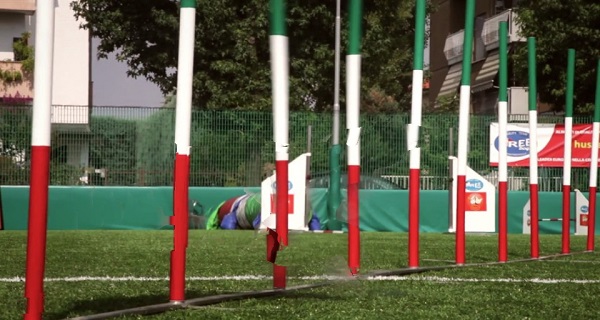} &    
     \includegraphics[width=0.162\linewidth]{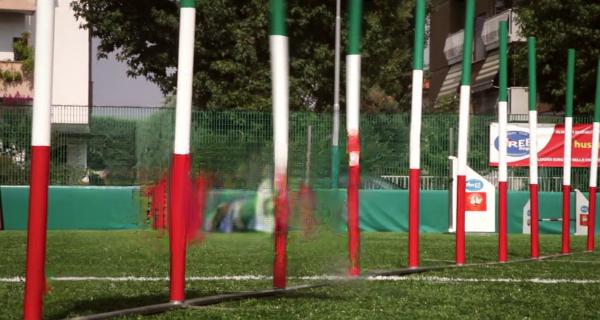} &       
     \includegraphics[width=0.162\linewidth]{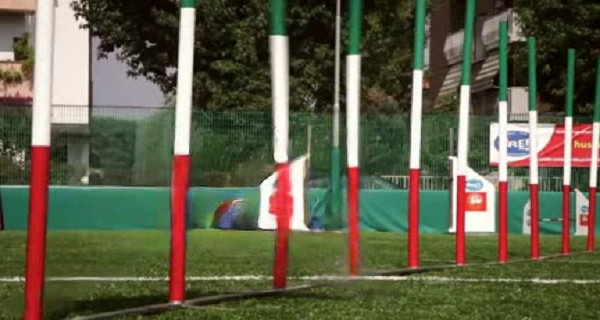} &     
     \includegraphics[width=0.162\linewidth]{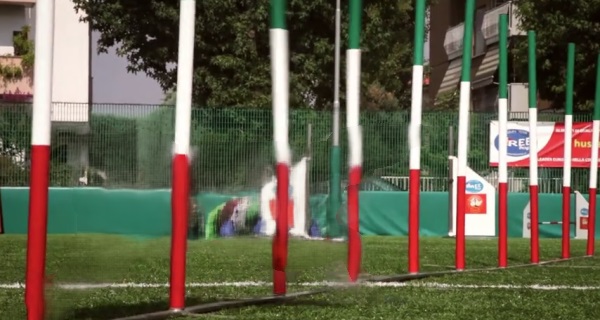}&      
     \includegraphics[width=0.162\linewidth]{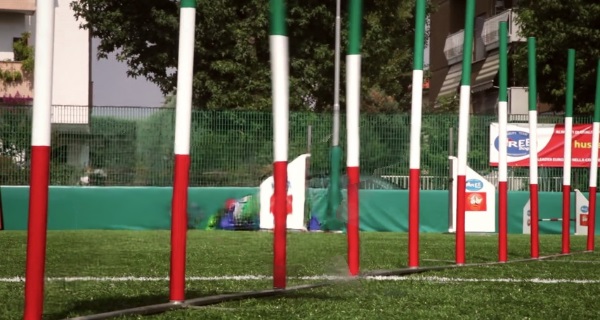} \\       
     
     \includegraphics[width=0.162\linewidth]{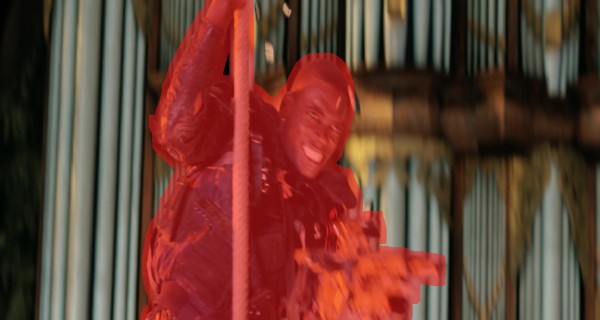} &      
     \includegraphics[width=0.162\linewidth]{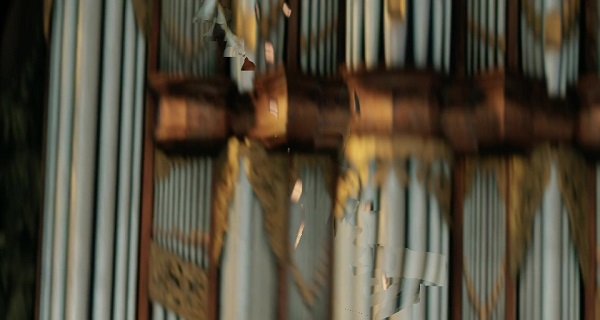} &    
     \includegraphics[width=0.162\linewidth]{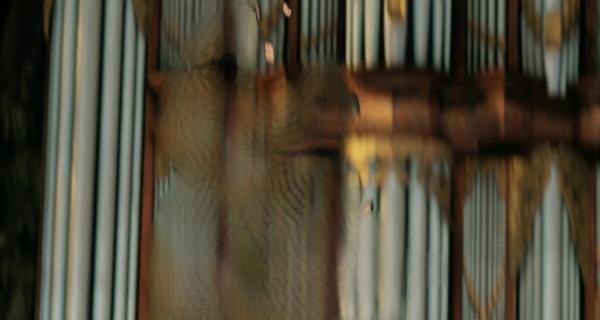} &       
     \includegraphics[width=0.162\linewidth]{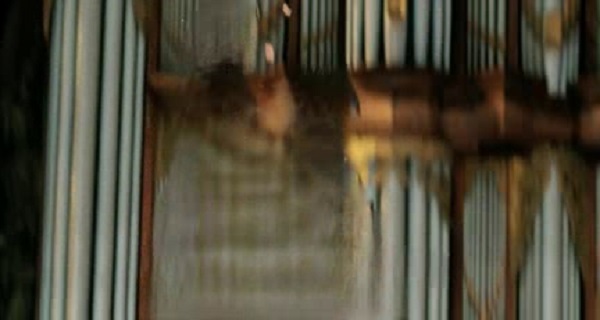} &     
     \includegraphics[width=0.162\linewidth]{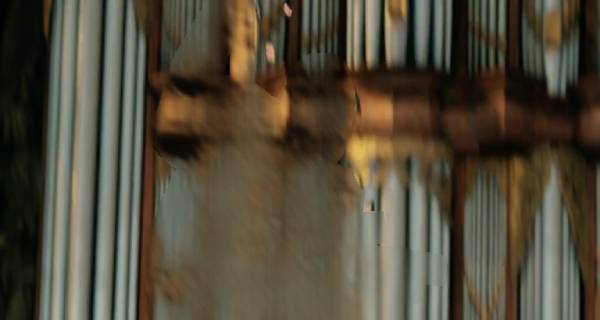}&      
     \includegraphics[width=0.162\linewidth]{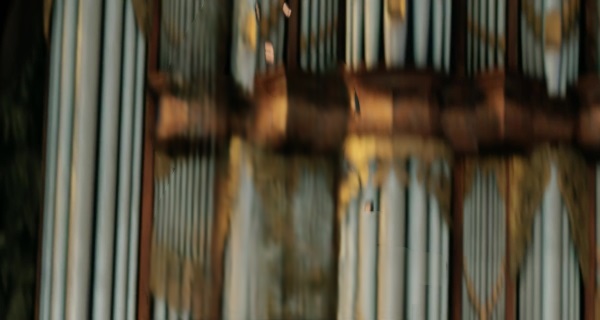} \\ 

     \includegraphics[width=0.162\linewidth]{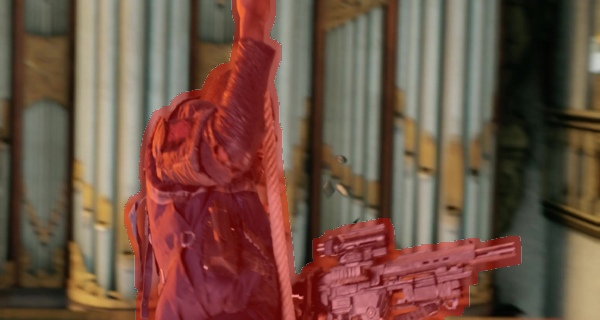} &      
     \includegraphics[width=0.162\linewidth]{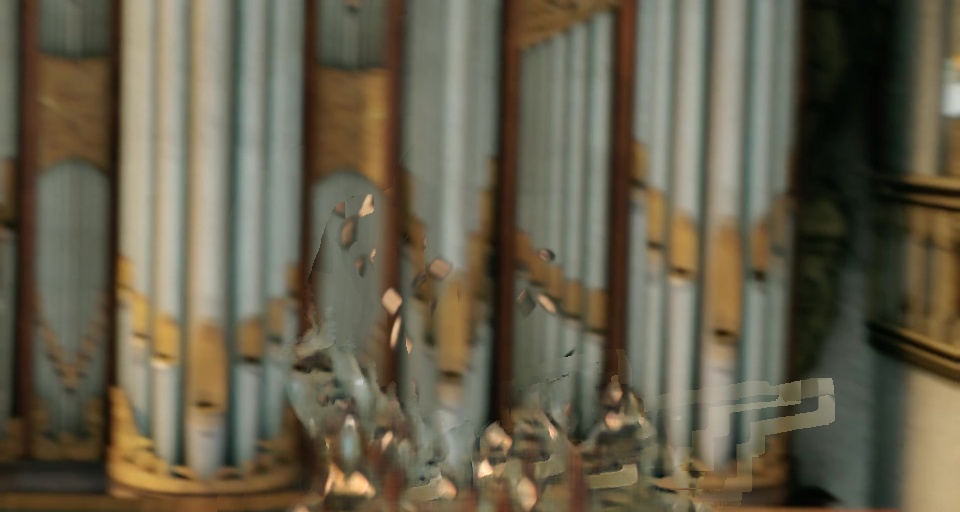} &    
     \includegraphics[width=0.162\linewidth]{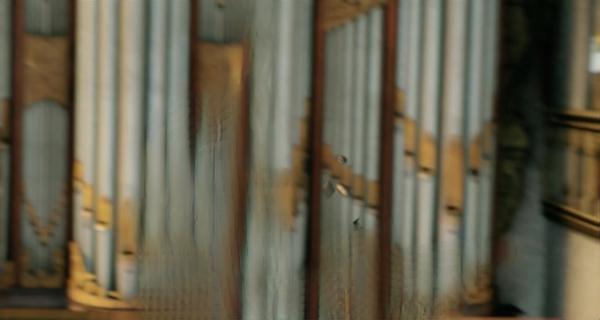} &       
     \includegraphics[width=0.162\linewidth]{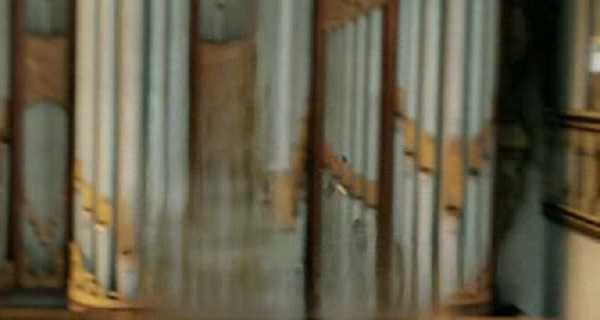} &     
     \includegraphics[width=0.162\linewidth]{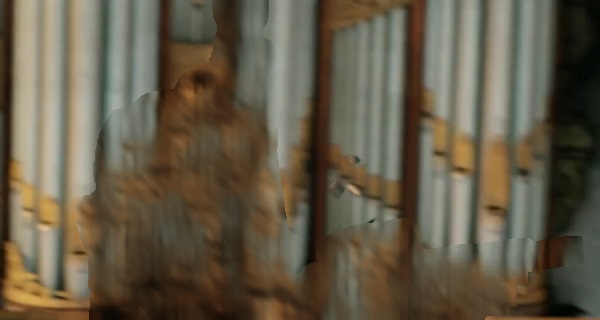}&      
     \includegraphics[width=0.162\linewidth]{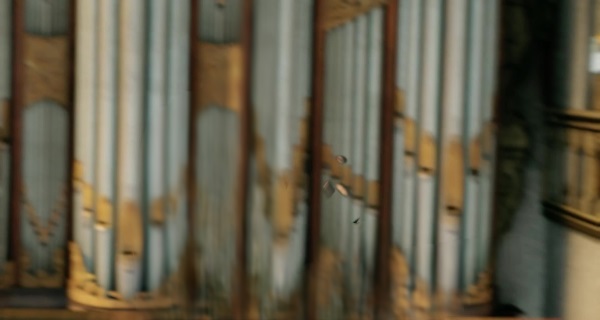} \\

    \includegraphics[width=0.162\linewidth]{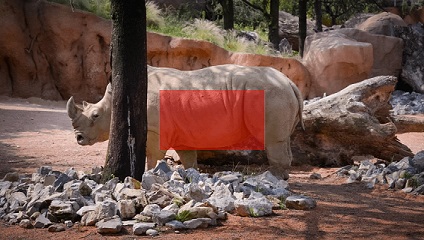} & 
     \includegraphics[width=0.162\linewidth]{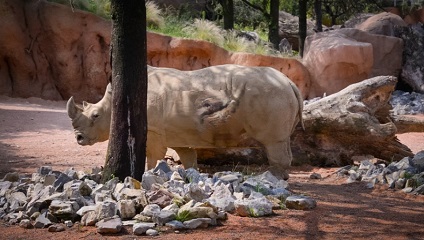} & 
     \includegraphics[width=0.162\linewidth]{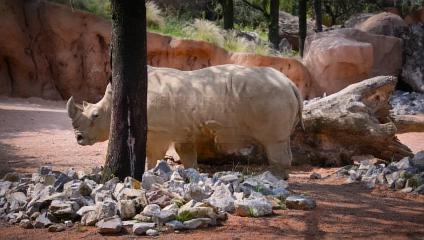} & 
     \includegraphics[width=0.162\linewidth]{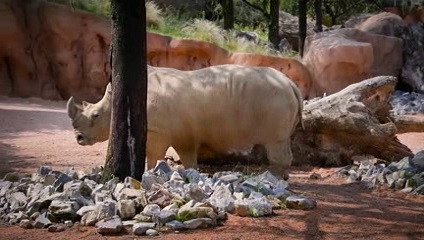} & 
     \includegraphics[width=0.162\linewidth]{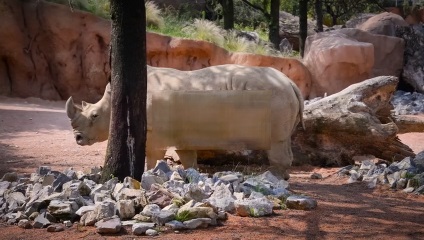} & 
     \includegraphics[width=0.162\linewidth]{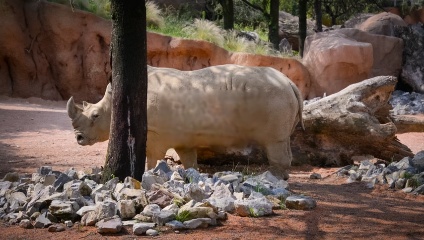} \\
     \includegraphics[width=0.162\linewidth]{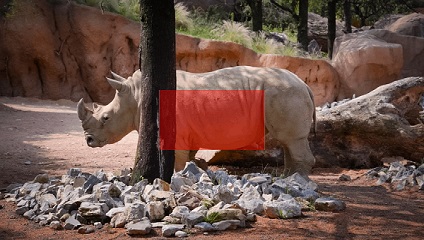} &      
     \includegraphics[width=0.162\linewidth]{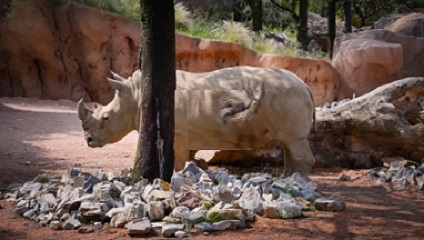} &    
     \includegraphics[width=0.162\linewidth]{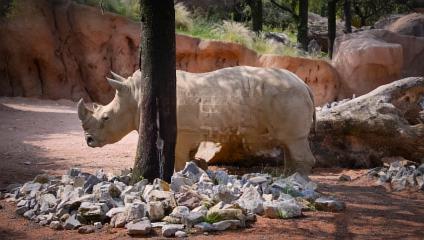} &       
     \includegraphics[width=0.162\linewidth]{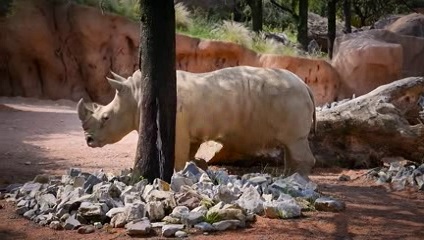} &     
     \includegraphics[width=0.162\linewidth]{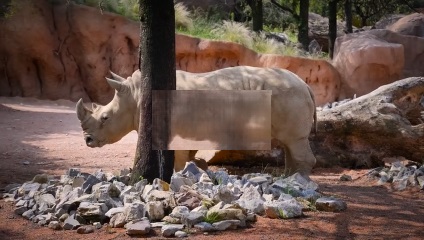}&      
     \includegraphics[width=0.162\linewidth]{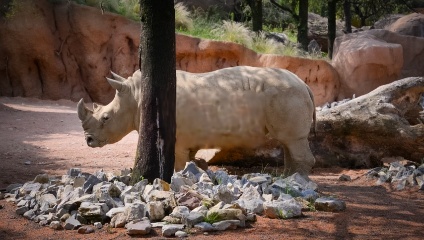} \\

      \includegraphics[width=0.162\linewidth]{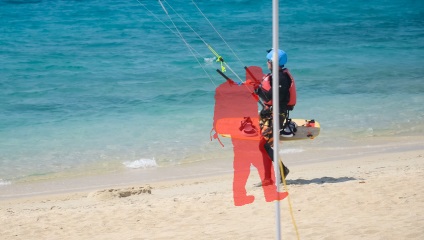} & 
     \includegraphics[width=0.162\linewidth]{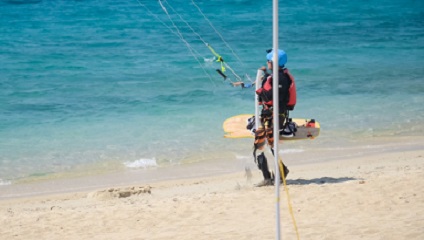} & 
     \includegraphics[width=0.162\linewidth]{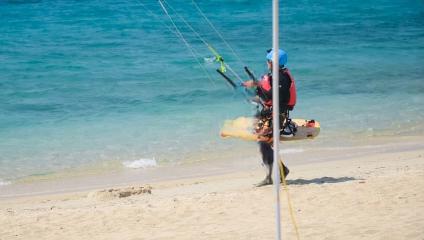} & 
     \includegraphics[width=0.162\linewidth]{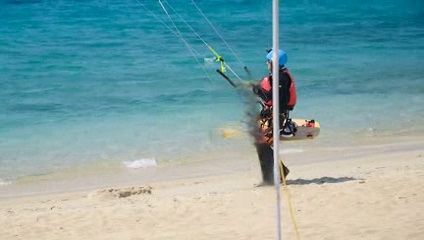} & 
     \includegraphics[width=0.162\linewidth]{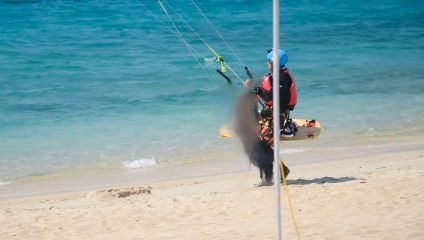} & 
     \includegraphics[width=0.162\linewidth]{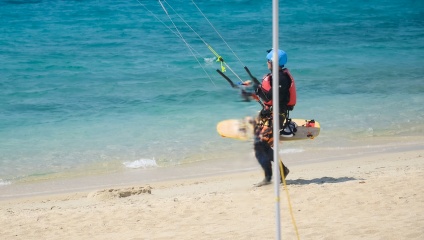} \\
     \includegraphics[width=0.162\linewidth]{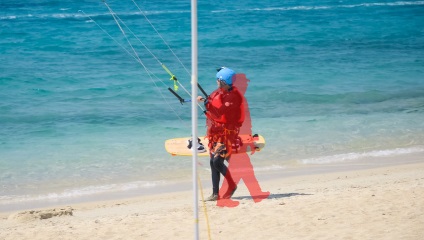} &      
     \includegraphics[width=0.162\linewidth]{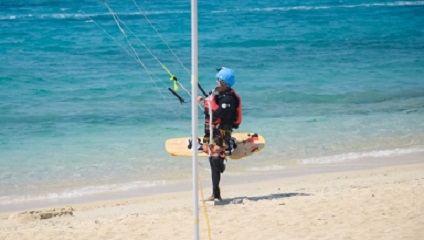} &    
     \includegraphics[width=0.162\linewidth]{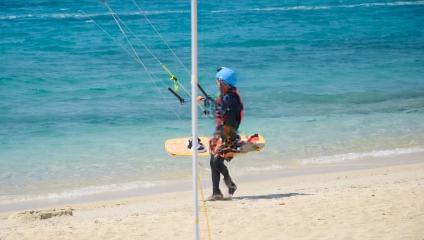} &       
     \includegraphics[width=0.162\linewidth]{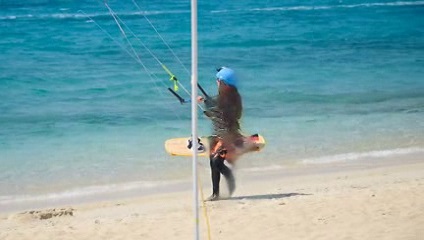} &     
     \includegraphics[width=0.162\linewidth]{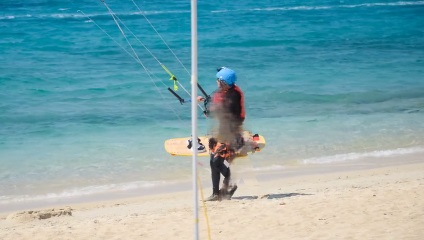}&      
     \includegraphics[width=0.162\linewidth]{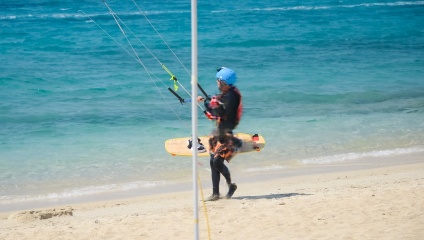} \\  

     Input&FGVC~\cite{gao2020flow}&OPN~\cite{oh2019onion}&STTN~\cite{zeng2020learning}&InternalVI~\cite{zhang2019internal}&Ours
    \end{tabular}
   \caption{Visual comparisons on the DAVIS dataset.  Zoom in for details.}
    \label{fig:qualitative}
\end{figure*}

We select five competitive baselines from three categories for fair comparison. CAP~\cite{lee2019copy}, OPN~\cite{oh2019onion}, and STTN~\cite{zeng2020learning} are the most recent external approaches based on deep neural network, FGVC~\cite{gao2020flow} is the newest flow-based optimization method and InterVI~\cite{zhang2019internal} is based on deep internal-learning. In Fig.~\ref{fig:qualitative}, we show the qualitative results on object removal, fixed masks and random object masks. Even if highly-similar content is available in other frames, previous approaches fail to propagate the long-range contextual information and produce blurry and distorted details (e.g., 1st, 3rd examples). They also fail to complete contantly blocked large missing regions and show abrupt artifacts (e.g., 2nd and 4th examples). In contrast, our method can generate realistic textures and reconstruct sharper and clearer structures. Extensive results on other sequences are attached in the \textbf{Appendix}. 

\begin{figure*}[t]
    \centering 
    \scriptsize
    \begin{tabular}{@{}c@{\hspace{0.3mm}}c@{\hspace{0.3mm}}c@{\hspace{0.3mm}}c@{\hspace{0.3mm}}c@{\hspace{0.3mm}}c@{}}
 
     \includegraphics[width=0.16\linewidth]{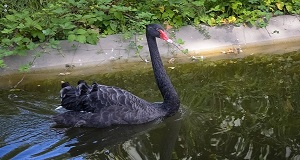} &      
     \includegraphics[width=0.16\linewidth]{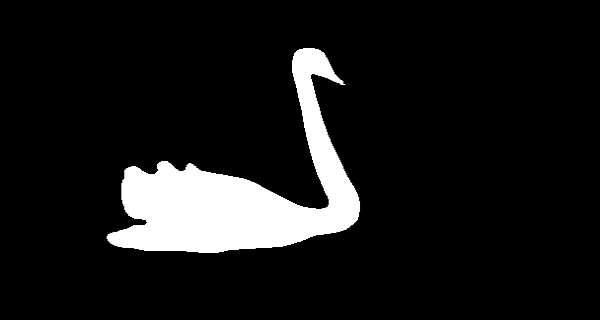} &    
     \includegraphics[width=0.16\linewidth]{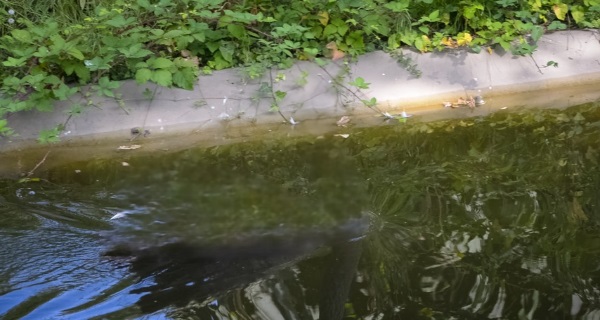} &       
     \includegraphics[width=0.16\linewidth]{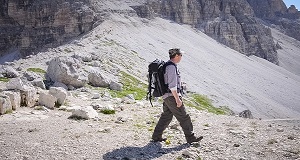} &     
     \includegraphics[width=0.16\linewidth]{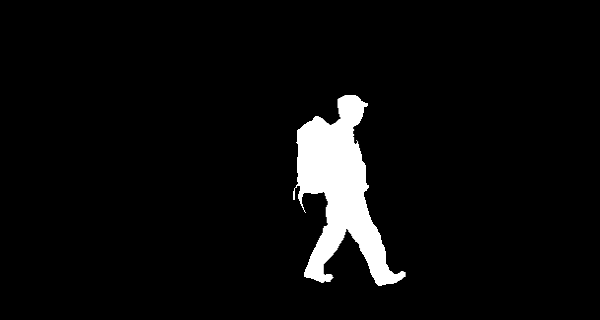}&      
     \includegraphics[width=0.16\linewidth]{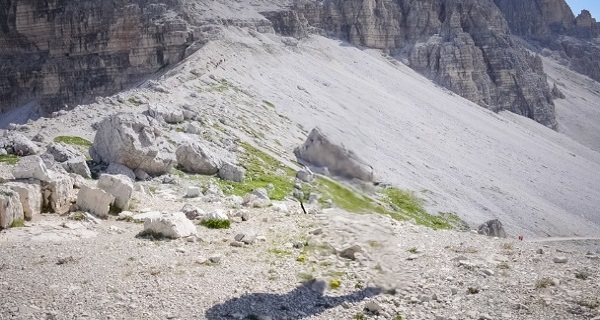} \\

      \includegraphics[width=0.16\linewidth]{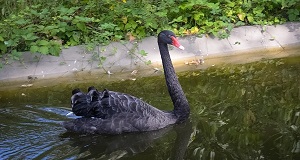} &      
     \includegraphics[width=0.16\linewidth]{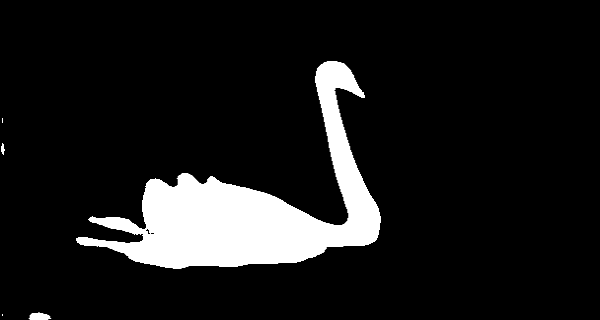} &    
     \includegraphics[width=0.16\linewidth]{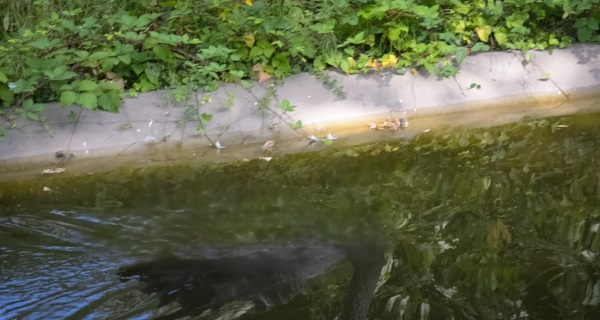} &       
     \includegraphics[width=0.16\linewidth]{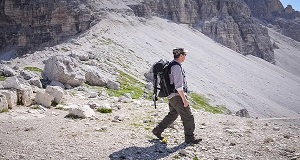} &     
     \includegraphics[width=0.16\linewidth]{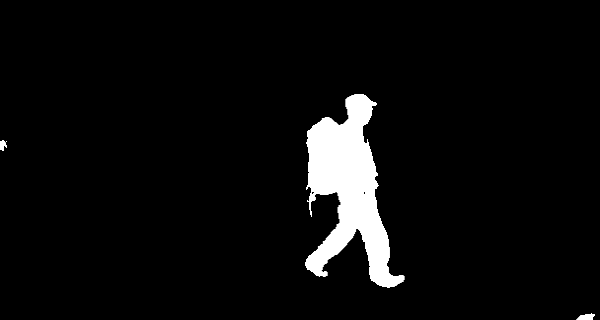}&      
     \includegraphics[width=0.16\linewidth]{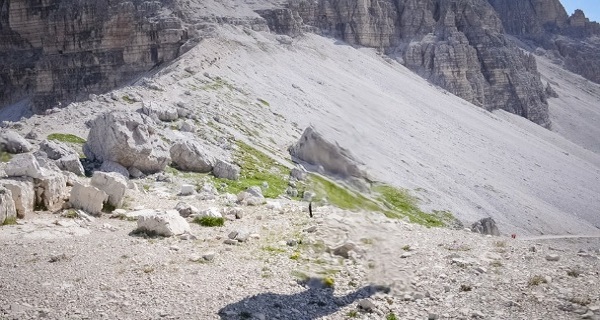} \\       
 
     Input Frame&Predicted Mask&Removal Result&Input Frame&Predicted Mask&Removal Result
    \end{tabular}
   \caption{\small{Mask prediction and object removal results given the mask of only the first frame.  }}
    \label{fig:single frame}
\end{figure*}

We also show the mask prediction and inpainting results in Fig.~\ref{fig:single frame}. Given a single mask of only one frame, our method can propagate it to other frames of the whole sequence automatically. Note that our internal method usually includes slightly more pixels to ensure that all objects pixels are correctly classified. It performs stably on non-rigid deformation masks and obtains promising results.

\begin{table} [t]
\scriptsize
\centering 
\begin{tabular}{@{}l @{\hspace{3mm}}  c @{\hspace{4mm}}  c @{\hspace{3mm}} c @{\hspace{3mm}} c @{\hspace{0.1mm}}c@{\hspace{6mm}}  c @{\hspace{3mm}} c @{\hspace{3mm}} c @{\hspace{0.1mm}}} 
\toprule 
&& \multicolumn{3}{c}{Fix Masks}&& \multicolumn{3}{c}{Object Masks} \\ 
\cmidrule{3-5} \cmidrule{7-9} 
Method& Type&  PSNR & SSIM & LPIPS  && PSNR & SSIM & LPIPS \\ 
\midrule 
CAP & \textit{external}& 28.04&	0.906 &0.1041 && 29.37&	0.910 &0.0483 \\
OPN & \textit{external}& 28.72 &	0.915 &0.0872 && 28.40	&0.904  &0.0596\\
STTN & \textit{external}&29.05	&0.927 &0.0637 && 29.45&	0.918 &0.0279 \\
\midrule
FGVC & \textit{flow-based}& 29.68	&0.942 &0.0564 && \textbf{33.98}& \textbf{0.951} &0.0195\\
\midrule
InterVI & \textit{internal}& 26.89	& 0.868 &0.1126 && 27.96&	0.875 &0.0545 \\
Ours & \textit{internal}& \textbf{29.90}&	\textbf{0.944} & \textbf{0.0414} && 31.09&	0.948 & \textbf{0.0182}\\
\bottomrule 
\end{tabular}
\caption{\small{Quantitative comparison on  DAVIS.}  } 
\label{tab:metric} 
\end{table}

\begin{figure}[t]
    \centering
    \includegraphics[width= \linewidth]{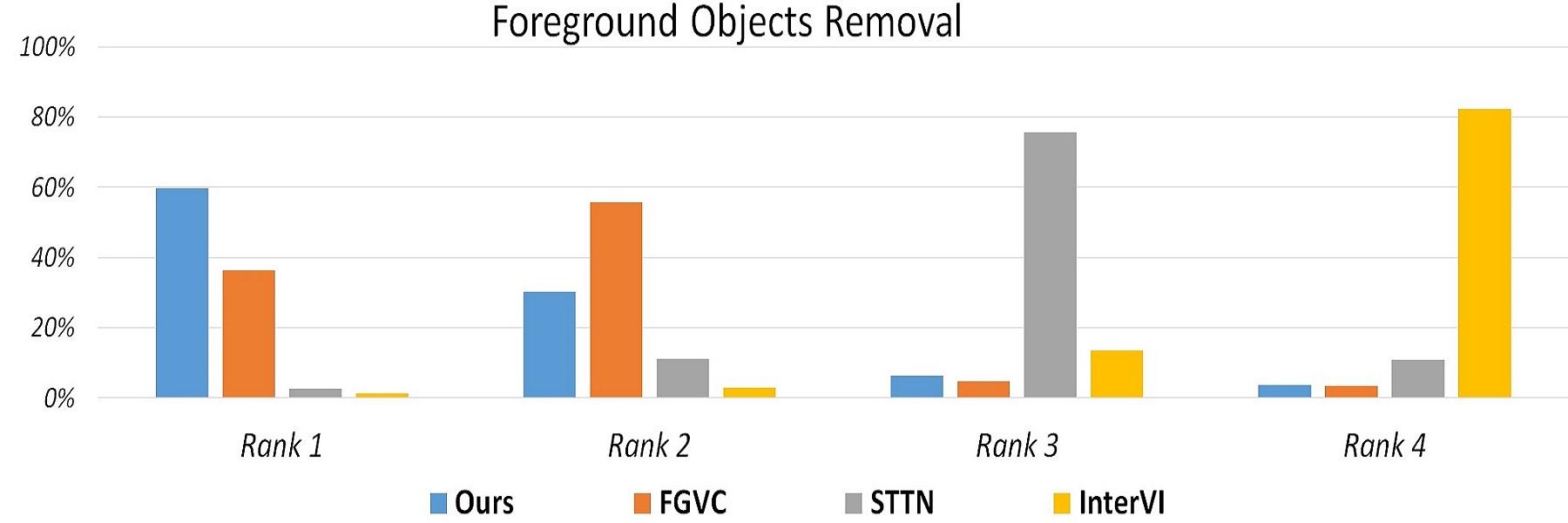}
    \caption{\small{Results of the user study. “Rank $x$” means the percentage of results from each method being chosen as the $x$-th best.}}
    \label{fig:user study}
\end{figure}

\subsection{Quantitative results}
There are no suitable quantitative metrics for video inpainting due to the ambiguity of ground truth. Nevertheless, we report comparison results of PSNR and SSIM on DAVIS in Table~\ref{tab:metric}, with both fixed and object masks.  For the fixed masks, we follow the settings in~\cite{xu2019deep} to simulate a fixed rectangular mask in the center of each frame. The width and height of rectangular masks are both $1/4$ of the original frame. For object masks, we follow the settings in~\cite{kim2019deep, oh2019onion,lee2019copy} to add imaginary objects on original videos by shuffling DAVIS sequences and masks. In this way, we can simulate test videos with known ground-truth. We show more detailed settings in the \textbf{Appendix}. As in Table~\ref{tab:metric}, our method shows substantially comparable performance with state-of-the-art approaches. The flow-based method achieving higher performance on random object mask is understandable since the synthesized object motion is simple which is the ideal case for using the flow-based method as mentioned in~\cite{gao2020flow}. The next section shows the user study on real-world object removal task which demonstrate the advantage that our method is more robust to complex motions.

\subsection{User study}
Quantitative metrics on synthetic dataset are not sufficient to evaluate the quality of real-world video inpainting. We, therefore, conduct a user study to compare our method with state-of-the-art approaches. Specifically, we randomly select 45 sequences from DAVIS dataset, and slow down the results to 10 FPS for better comparison.  We invite 18 participants to rank the  results of four methods for each video in terms of visual quality and temporal consistency, and received 810 effective votes totally. Fig.~\ref{fig:user study} shows that our approach outperforms other methods by a large margin.

\section{Ablation study} \label{ablation study}
\subsection{Regularization terms}

\begin{figure}[t]
    \centering 
    \scriptsize
    \begin{tabular}{@{}c@{\hspace{0.35mm}}c@{\hspace{0.35mm}}c@{}}
 
     \includegraphics[width=0.321\linewidth]{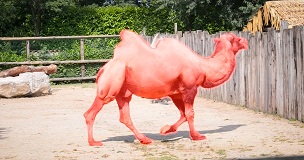} &      
     \includegraphics[width=0.321\linewidth]{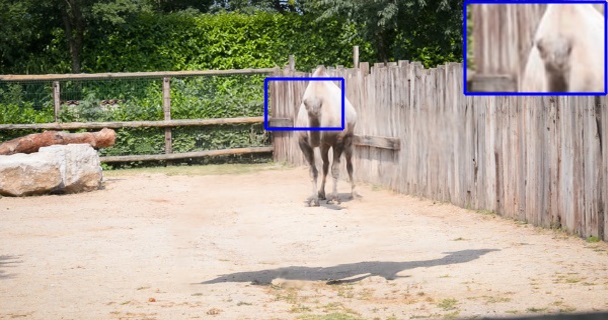} &    
     \includegraphics[width=0.321\linewidth]{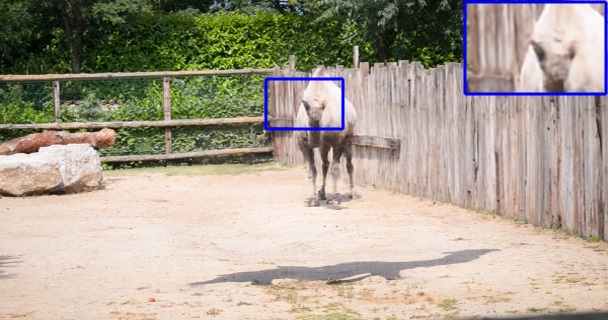}  \\

      \includegraphics[width=0.321\linewidth]{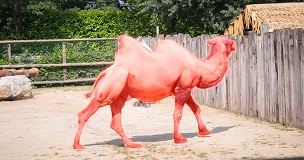} &      
     \includegraphics[width=0.321\linewidth]{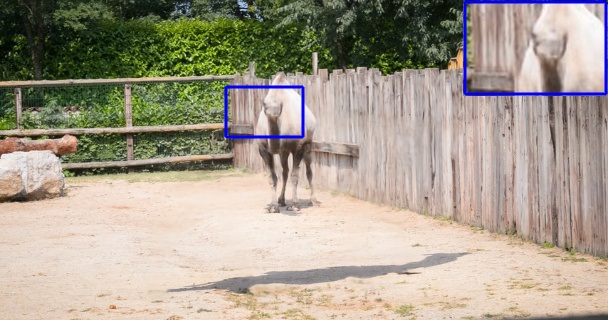} &    
     \includegraphics[width=0.321\linewidth]{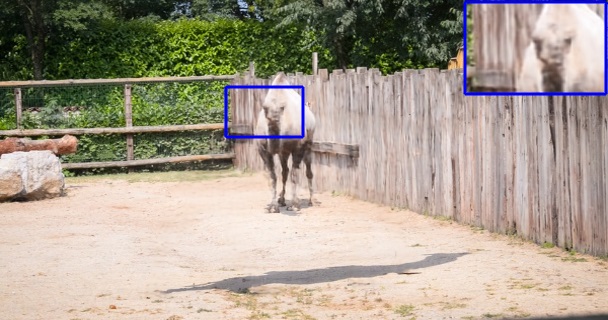}  \\    
     
    \includegraphics[width=0.321\linewidth]{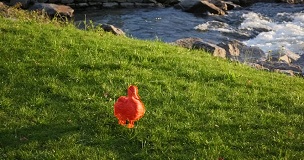} &      
     \includegraphics[width=0.321\linewidth]{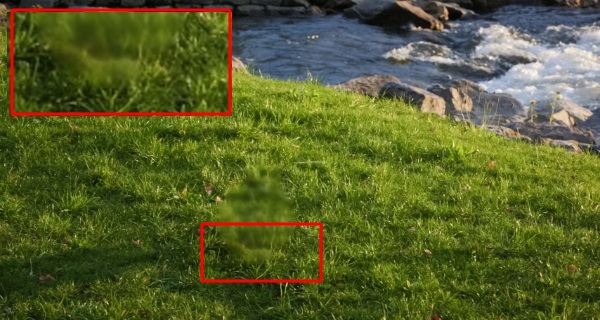} &    
     \includegraphics[width=0.321\linewidth]{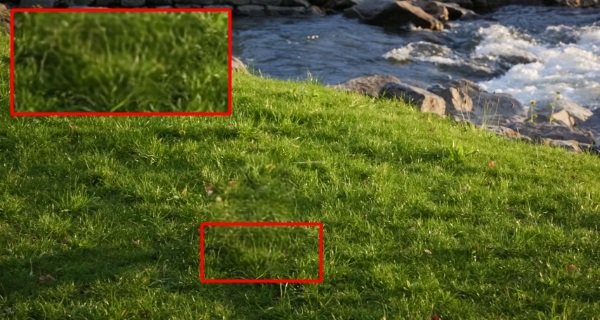}  \\   
 
      \includegraphics[width=0.321\linewidth]{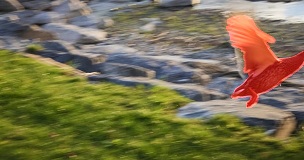} &      
     \includegraphics[width=0.321\linewidth]{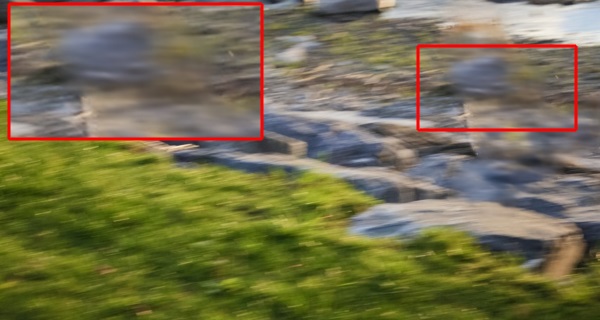} &    
     \includegraphics[width=0.321\linewidth]{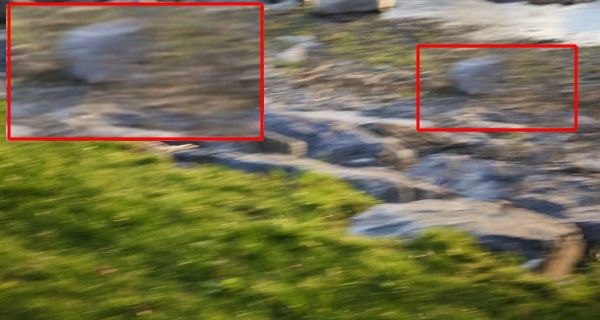}  \\  
     
    \includegraphics[width=0.321\linewidth]{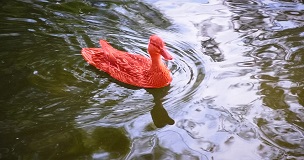} &      
     \includegraphics[width=0.321\linewidth]{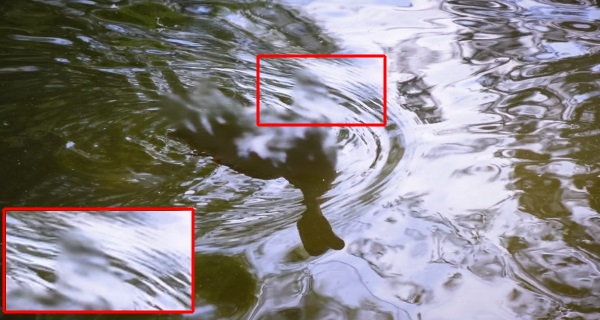} &    
     \includegraphics[width=0.321\linewidth]{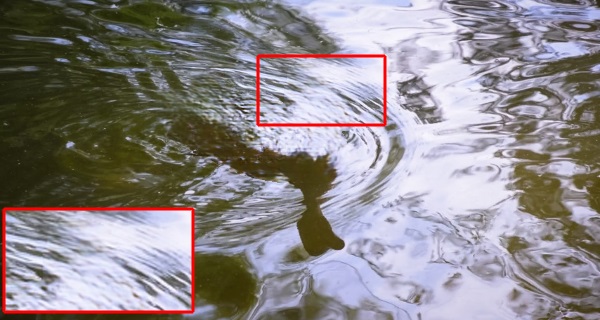}  \\

      \includegraphics[width=0.321\linewidth]{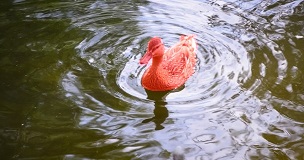} &      
     \includegraphics[width=0.321\linewidth]{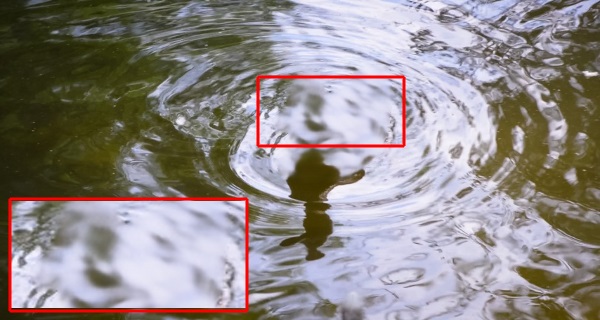} &    
     \includegraphics[width=0.321\linewidth]{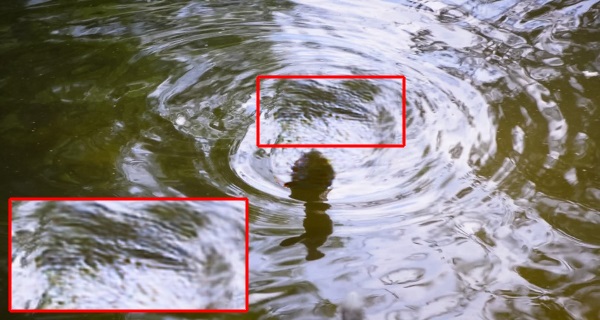}  \\  
   Input Frame & w/o anti-ambiguity loss & with anti-ambiguity loss
    \end{tabular}
  
   \caption{\small{Ablation study on the anti-ambiguity regularization. Zoom in for details.}}
    \label{fig:ambiguity_loss}
\end{figure}

As introduced in Sec. ~\ref{method}, the anti-ambiguity loss aims at recovering the blurry details for the ambiguous cases.  Fig.~\ref{fig:ambiguity_loss}  demonstrates examples of moving background, motion blur or complex textures. With the proposed anti-ambiguity term, the background camel's face, the textures of the grass, and the water ripples contain more realistic details. 

\begin{figure}[t]
\centering
\includegraphics[width=\linewidth]{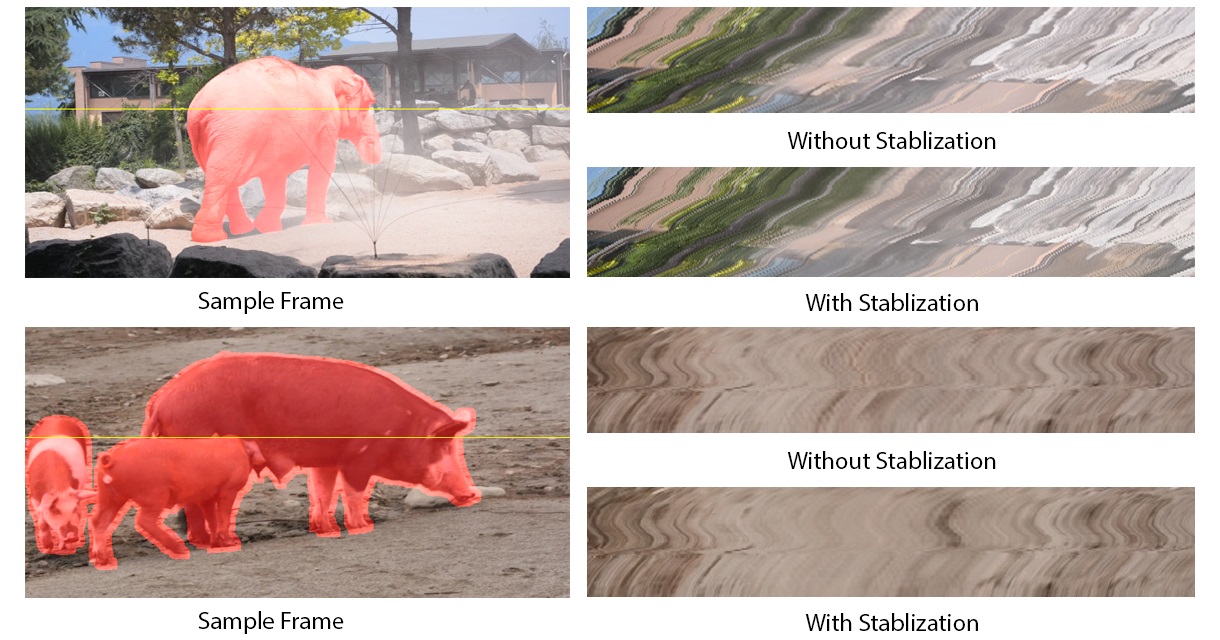}
\caption{\small{Ablation study on the stabilization regularization. }}
\label{fig:stablization}
\end{figure}

In Fig.~\ref{fig:stablization}, we compare the temporal consistency of the model with and without the stabilization term. The pixels indicated by the yellow line are stacked from all the frames. Without the proposed stabilization procedure, the temporal graph contains a considerable amount of noise.

\subsection{Properties analysis}
As the first step of our method is to overfit a CNN to memorize all the information in a video sequence, an important aspect is how many parameters are necessary for a specific video sequence which is to analyze Property \ref{itm:first}. We use three networks with an increasing number of parameters for testing. With detailed analysis and examples attached in the \textbf{Appendix}, we find that using a simple CNN with fewer parameters will suffer a performance drop as the video complexity increase. By involving more parameters to CNN models, the inpainting result on videos with dramatic change or complicated background becomes more realistic. 

As mentioned in Sec. \ref{method}, we utilize augmented object masks for the internal training as Property \ref{itm:third} requires all the inputs, including the mask shape, to be ideally the same. Another interesting aspect is the performance change if we replace the mask generation strategy with random free-form mask generation. In the new settings, although the training loss diverges, the testing results suffer degradation (in the \textbf{Appendix}). This observation shows that using the augmented ``object masks", the model can better propagate the information of known regions  to unknown regions.

\section{Conclusion} \label{discussions}
In this work, we propose to formulate video inpainting in a novel way: implicitly propagating long-range information using an overfitted CNN without explicit guidance like optical flow. It succeeds in challenging cases such as large masks, complex motion, and long-term occlusion with the proposed anti-ambiguity and stabilization regularization. We also extend the method to more challenging settings, such as using only a single mask or high-resolution 4K videos. However, there are still two main limitations. Our method is not real-time in training each sequence, like all the other internal learning methods. The other issue is that the details of inpainted areas for the deficiency cases can sometimes be semantically incorrect (See the \textbf{Appendix} for failure cases). It is possible to incorporate external semantic information to generate such regions. In the future work, we may explore more on producing better details for deficient frames with inpainted region's confidence.

{\small
\bibliographystyle{ieee_fullname}
\bibliography{egbib}
}
\clearpage

\appendix 
\section*{\LARGE Appendix}
\section{Implementation details}
\subsection{Network architecture}
Following the network design in~\cite{yu2018generative},  we use three basic blocks in our network:

\textbf{ConvBlock} contains three layers in a sequence: a Conv2d layer with output channel number $out\_c$, kernel size $k\_size$, stride $s$ and dilation $d$, an activation layer using ELU activation, and the splitting and gating layer in~\cite{yu2018generative}.  The padding method is reflection padding. We denote it as $C[out\_c, k\_size, s, d]$.

\textbf{DeconvBlock} is composed of a bilinear upsampling layer followed by the ConvBlock. We denote it as $D[out\_c, k\_size, s, d]$.

\textbf{OutputBlock}  is designed for generating the output. It is composed of one ConvBlock $C[3, 1, 1, 1]$ followed by a $tanh$ activation.  We denote it as $O$. 

\textbf{Base network} comprises of 18 convolution layers. It can be represented as:\\ C(48,5,1,1)-C(96,3,2,1)-C(96,3,1,1)-C(192,3,2,1)-C(192,3,1,1)-C(192,3,1,1)-C(192,3,1,2)-C(192,3,1,4)-C(192,3,1,8)-C(192,3,1,16)-C(192,3,1,1)-C(192,3,1,1)-D(96,3,1,1)-C(96,3,1,1)-D(48,3,1,1)-C(24,3,1,1)-O. 

In our experiments, we use two additional models for analyzing the relationship between the model capacity and the complexity of the video sequence. 

\textbf{Stacked network} We stack our base network two times and train the model with additional intermediate supervision on the first network. The parameters in this network are doubled in comparison to the base network.  

\textbf{Coarse-to-fine network} We add an additional branch in the second base network in the stacked network and adapt it to a coarse-to-fine version. The new branch is represented as C(48,5,1,1)-C(96,3,2,1)-C(96,3,1,1)-C(192,3,2,1)-C(192,3,1,1)-C(192,3,1,1)-C(192,3,1,2)-C(192,3,1,1)-C(192,3,1,1). We concatenate the output of it to the features of the original branch and go through the output branch for the final output.

We use the coarse-to-fine network as the default settings for most of our experiments.

\subsection{Training details}
\textbf{Object removal on DAVIS.}
By default, we first train the model with only reconstruction loss for about 60,000 iterations and then combine with the regularization loss for another 20,000 iterations for the hard sequences. We use Adam Optimizer with a learning rate of $2e-4$. The training process takes about 4 hours on a single NVIDIA RTX 2080 Ti GPU for 80-frame videos. 
 
 \textbf{Fixed masks and random object masks on DAVIS.}
 For the fixed masks, we follow the settings in~\cite{xu2019deep} to simulate a fixed rectangular mask in the center of each frame. The width and height of rectangular masks are both $1/4$ of the original frame. For object masks, we follow the settings in~\cite{kim2019deep, oh2019onion,lee2019copy} to add imaginary objects on original videos by shuffling DAVIS sequences and masks. In this way, we can simulate test videos with known ground-truth for quantitative metrics calculation. Note that for inpainting of both fixed masks and random object masks, the ground-truth images are only used for PSNR and SSIM calculation, and the ground-truth pixels of masked regions are never seen when training.\\
 
\textbf{Object removal with a single mask.}
We first train the network for object mask propagation for 6,000 iterations. We then follow the same setting as that on Davis  but use the predicted masks for object removal.\\

\textbf{4K video sequences.}
We first train a coarse inpainting net at 1K resolution with a learning rate of $1e-4$ using batch size 2 for 30,000 iterations. The model is then progressively trained at 2K and 4K resolution with the same learning rate for 45,000 and 60,000 iterations, respectively.

\subsection{Objective functions}
We include more details of our objective functions in this section.  The total loss is summarized as:
\begin{equation}
\mathcal{L} = \mathcal{L}_{rec} + \lambda_{a}\mathcal{L}_{ambiguity} + \lambda_{s}\mathcal{L}_{stabilize}, 
\end{equation}
where $\lambda_{a}$ and $\lambda_{s}$ are set to 0.1 and 0.2 as default. \\

\textbf{Ambiguity regularization}
We further describe more details of the anti-ambiguity regularization with  the distance measurement $\delta$, as follows: 
\begin{equation}
\mathbb{D}^{s\times t}=\Theta\left(\frac{\Phi(X_{s}\odot M_{t} )}{\left\|\Phi(X_{s}\odot M_{t} )\right\|}, \frac{\Phi(X_{t}\odot M_{t})}{\left\|\Phi(X_{t}\odot M_{t})\right\|}\right),
\end{equation}
where $\Theta(S,T)$ calculates the all-pair distance matrix between two tensors, $\mathbb{D}_{i, j}$ denotes the distance between $S_i$ and $ T_j$.

\begin{equation}
\mathcal{L}_{\mathrm{ambiguity}} = \frac{1}{N}   \sum_{j} \min_{i} \mathbb{D}_{i, j}
\end{equation}

This term can be seen as a  perceptual loss for unaligned images~\cite{mechrez2018contextual}. The source image is each output frame $\Tilde{Y}_i$, and the target image is another random frame $X_t$  in the same input video.  The shared encoder is a pre-trained VGG-19 network, and we use it to extract the feature maps (\textit{relu3-4}) for loss calculation.  The  distance metric $\mathbb{D}$ is cosine distance between `pixels'   in the source   and  target feature map.  We minimize the distance   between each source feature point and its nearest  target feature point.

We do not impose   constrains for the random frame selection in $\mathcal{L}_{ambiguity}$. As the training involves thousands of iterations, in some iterations the selected frames are possibly also blocked, while in other iterations the selected frames can be helpful. In the case  we select a ``unblocked frame", the ambiguity loss can be regarded as a perceptual loss for unaligned images to refine the details . In the case we select a ``blocked frame", since the loss at the blocked region is always set to 0, it will neither improve nor damage the generation quality. A better selection strategy can be explored to improve the convergence speed in the future work.  \\

\textbf{Stabilization regularization}
We give more details of the augmentation parameters in the stabilization loss. As mentioned in our paper, we simulate slight changes in the video sequences by random transformation and image manipulation.
For the transformation, following~\cite{eilertsen2019single}, we utilize four types of transformation types that are translation, rotation, scaling, and shearing. The range of these parameters are respectively from $[-3 pixel, 3 pixel], [-2^{\circ}, 2^{\circ}], [0.95, 1.05]$ and $[-1^{\circ}, 1^{\circ}]$. For image manipulations, we use two types of image manipulation operations. One is the linear intensity changes that range from $[0.95, 1.05]$ and the other one is the standard Gaussian kernel with $3\times3$ or $5\times5$ kernel size.

\subsection{Progressive inpainting}
As mentioned in our paper, our approach can be extended to complete ultra high-resolution videos with a progressive inpainting scheme. In this section, we describe the specified procedure of progressive inpainting.\\

\textbf{Progressive learning} 
The target of our progressive learning is to generate high-resolution video sequences. In the first stage, the input size and the output size of the image are set to $960\times540$ to generate the coarse inpainting results as an additional prior for the later stages. In the second stage, we randomly sample a $960\times540$ patch from the $1920\times1080$ input with the up-sampled test result from the first stage. The sampled patch is then augmented with a random mask from the grid mask set (introduced in the next subsection). For the last stage, we adopt the same training strategy as the second stage and replace the prior with the test result from the second stage.\\

\textbf{Grid-based inference}
In the second stage and the last stage, we respectively utilize $2\times2$ and $4\times4$ grid of patches (with resolution $960\times540$) for inference. We split the corresponding original masks accordingly to the grid mask set. To translate the mask, we exclude the masks from the grid mask set that have less than 1,000 object pixels. We sample an additional testing patch around the edge of the grid for robustly blending the edge. \\

\textbf{Boundary-aware sampling}
 In our original formulation, for every training iteration, the mask is augmented by random translation. As the boundary regions usually contain more related information, we expect to sample more patches around the boundary regions for efficient training. For each frame in the sequence, we calculate the bounding box of the mask. We adopt a simple strategy: calculate a  translation range that moves the center of the augmented mask to around the boundary (45 pixels for height and 80 pixels for width) of the input image mask and set the possibility of this range five times of others.    We call it ``boundary-aware” because we sample more patches around the object boundary by setting higher possibility near the boundary.

\begin{table} [t]
\small
\centering 
\begin{tabular}{@{\hspace{2mm}}l @{\hspace{2mm}}  c @{\hspace{3mm}}  c @{\hspace{3mm}}  c @{\hspace{1mm}}c@{\hspace{5mm}}  c @{\hspace{3mm}} c @{\hspace{2mm}}} 
\toprule 
&& \multicolumn{2}{c}{Simple}&& \multicolumn{2}{c}{Complex} \\ 
\cmidrule{3-4} \cmidrule{6-7} 
Model&  &  PSNR & SSIM  && PSNR & SSIM \\ 
\midrule

Base model & &  30.05  & 0.948	&& 25.46 &	0.908 \\
Stacked model && 32.58 & 0.951	&&  27.27 &  0.909 \\
Coarse-to-fine model & &  32.70 & 0.953 &&  28.54  & 0.920	 \\
\bottomrule 
\end{tabular}
\caption{\small{Quantitative results of models with different capacities.}  } 
\label{tab:capacity} 
\end{table}

\begin{table} [t]
\small
\centering 
\begin{tabular}{@{}l @{\hspace{5mm}}   c @{\hspace{4mm}}  c @{\hspace{4mm}}c@{\hspace{4mm}}  c @{\hspace{4mm}} c @{\hspace{1mm}}} 
\toprule 
Method& $\mathcal{L}_{rec}$& $\mathcal{L}_{ambiguity}$ & $\mathcal{L}_{stable}$  & PSNR & SSIM \\ 
\midrule 
Ours & \checkmark&-&-& 29.36&	0.932 \\
Ours & \checkmark&\checkmark&-& 29.21&	0.931 \\
Ours & \checkmark&-&\checkmark& 29.08&	0.929\\
\bottomrule 
\end{tabular}
\caption{Quantitative evaluation of different regularization losses. Note that results with higher scores of PSNR and SSIM are not necessarily visually better.  } 
\label{tab:metric-supp} 
\end{table}

\section{Analysis}
\textbf{Model capacity and video complexity}
We further explore the relationship between the model capacity and video complexity. The definition of the video complexity follows the design in ~\cite{zhang2019internal}, and we conduct experiments with the fixed rectangular masks. As introduced in section 1.1, the model parameters decrease from the coarse-to-fine model, the stacked model, to the base model. In Tab.~\ref{tab:capacity}, we observe that all three models achieve good performance for the simple sequences while the coarse-to-fine model achieves the most stable performance on the complex sequences.  \\

\textbf{``Overfitting'' ability and video length}
In property 2, we assume that the neural network can be regarded as a universal function to fit to a certain amount of image data. However, it remains unknown how the amount of data affects the overfitting ability. We thus evaluate the performance of fitting unmasked regions in regards to different number of video frames. We use subsets of the \textit{BMX-TREES} sequence (80 frames in total)  for testing. For video lengths of 5/20/80, the PSNR of the over-fitting results are respectively 48.78/45.32/41.20. The current model can fit to DAVIS sequence quite well but for longer sequences we may need larger models. \\

\begin{figure}[t]
    \centering 
    \scriptsize
    \begin{tabular}{@{}c@{\hspace{0.3mm}}c@{\hspace{0.3mm}}c@{}}
 
     \includegraphics[width=0.33\linewidth]{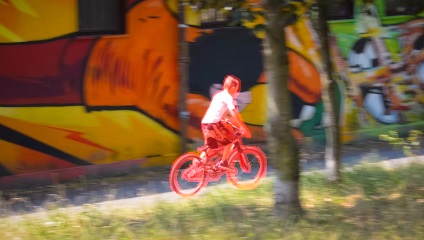} &      
     \includegraphics[width=0.33\linewidth]{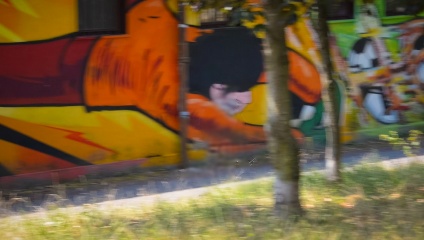} &    
     \includegraphics[width=0.33\linewidth]{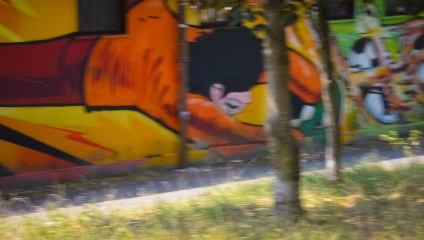}  \\   
     
          \includegraphics[width=0.33\linewidth]{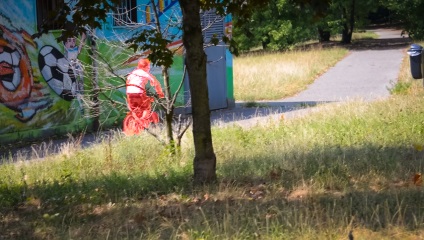} &      
     \includegraphics[width=0.33\linewidth]{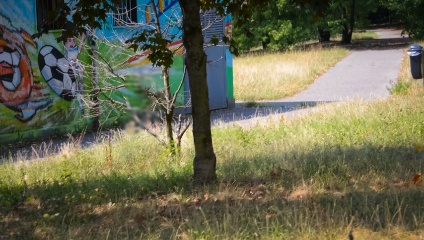} &    
     \includegraphics[width=0.33\linewidth]{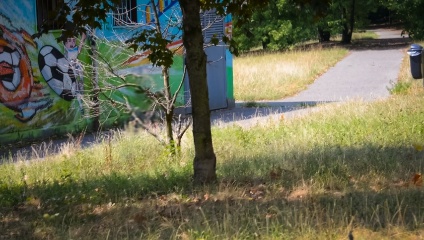}  \\   
     
          \includegraphics[width=0.33\linewidth]{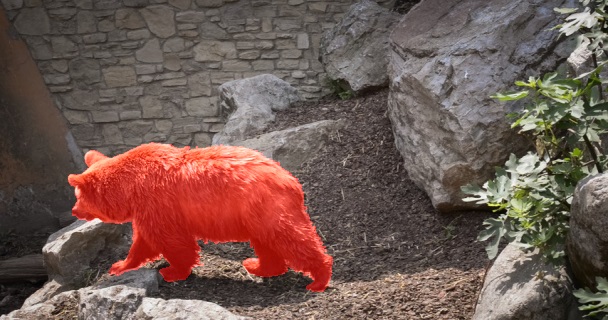} &      
     \includegraphics[width=0.33\linewidth]{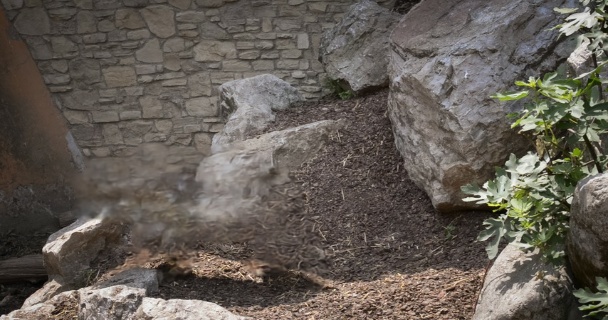} &    
     \includegraphics[width=0.33\linewidth]{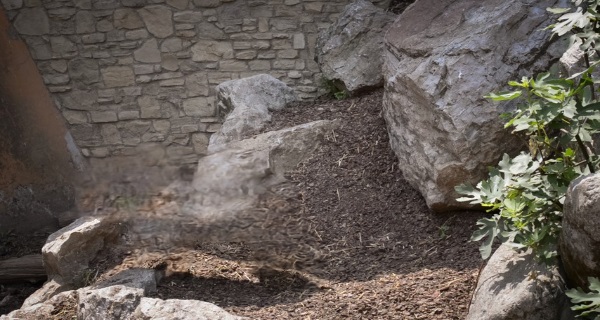}  \\   
     
     \includegraphics[width=0.33\linewidth]{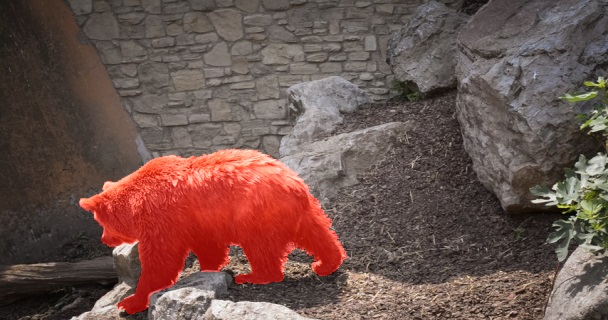} &      
     \includegraphics[width=0.33\linewidth]{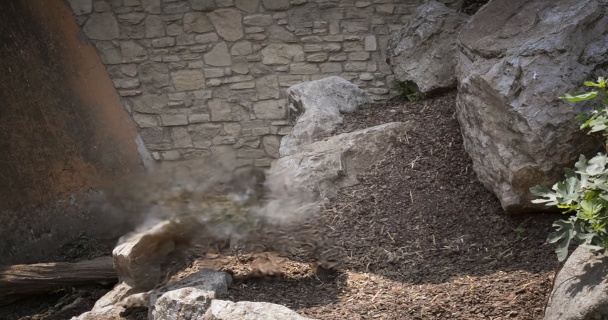} &    
     \includegraphics[width=0.33\linewidth]{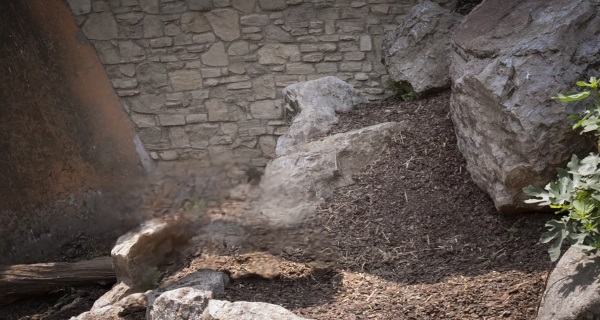}  \\

   Input Frame & Random Mask & Augmented Object Mask
    \end{tabular}
  
   \caption{\small{Effect of different mask generation schemes for our internal video inpainting approach. Zoom in for details.}}
    \label{fig:mask_generation}
\end{figure}

\textbf{Mask generation scheme}
We also explore how the mask generation affects the performance of the inpainting. Specifically, we change the mask generation scheme from our augmented object masks to the random mask generation scheme proposed in~\cite{yu2019free}.  We observe a huge performance drop when we adopt this alternative mask generation strategy. As shown in Fig.~\ref{fig:mask_generation}, the model trained with random masks tends to generate blurry details.  \\

\textbf{Regularization loss}
We observe significant perceptual improvements in terms of the video quality brought by the proposed regularization terms in our paper. However,  as shown in many previous works, models trained with only reconstruction loss usually tend to achieve better quantitative performance with blurry results, while visually better results may suffer performance drop in terms of PSNR and SSIM. Still, we report the quantitative metrics in Table~\ref{tab:metric-supp}.

\section{More results}

\begin{figure*}
    \centering 
    \small
    \begin{tabular}{@{}c@{\hspace{0.3mm}}c@{\hspace{0.3mm}}c@{\hspace{0.3mm}}c@{\hspace{0.3mm}}c@{\hspace{0.3mm}}c@{}}

     \includegraphics[width=0.162\linewidth]{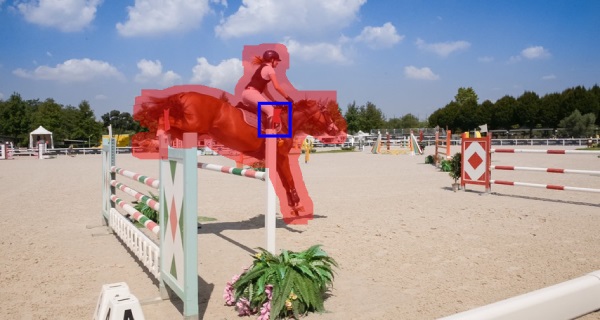} & 
     \includegraphics[width=0.162\linewidth]{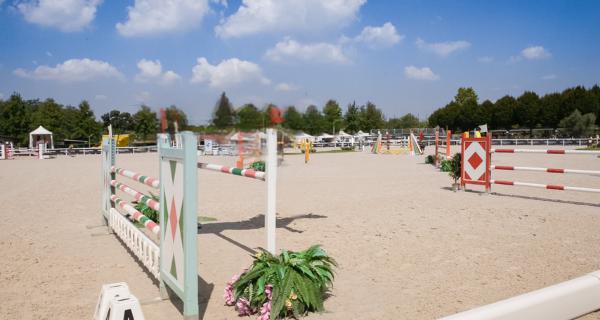} & 
     \includegraphics[width=0.162\linewidth]{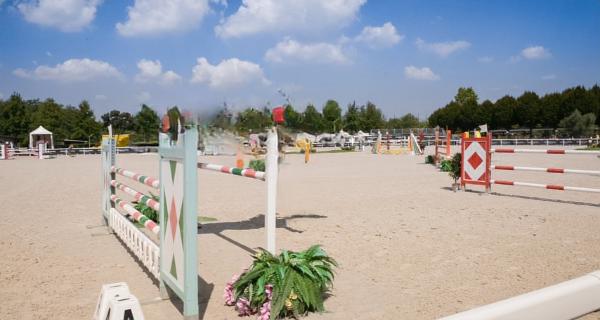} & 
     \includegraphics[width=0.162\linewidth]{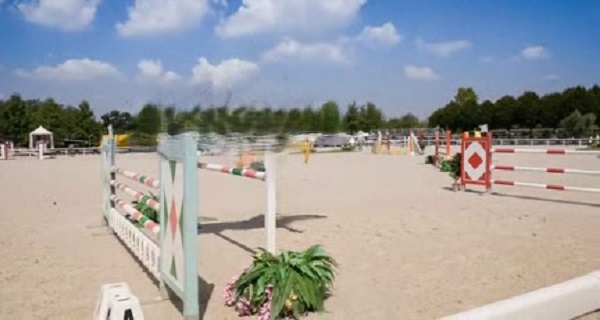} & 
     \includegraphics[width=0.162\linewidth]{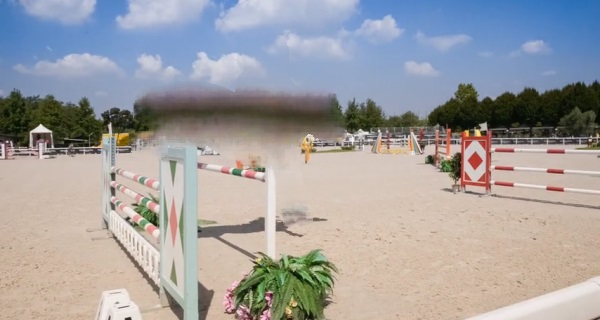} & 
     \includegraphics[width=0.162\linewidth]{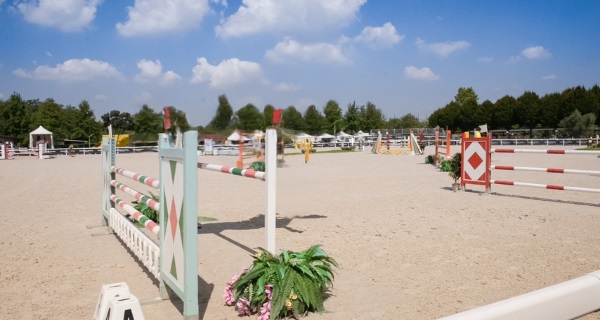} \\
     \includegraphics[width=0.162\linewidth]{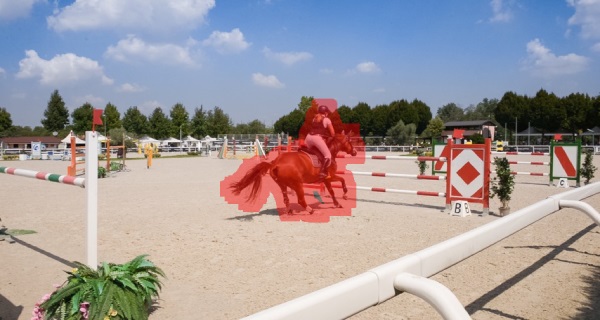} &      
     \includegraphics[width=0.162\linewidth]{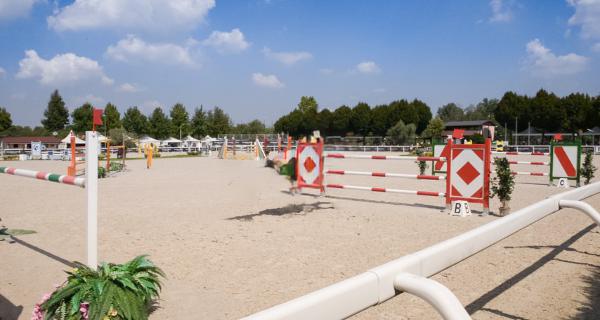} &    
     \includegraphics[width=0.162\linewidth]{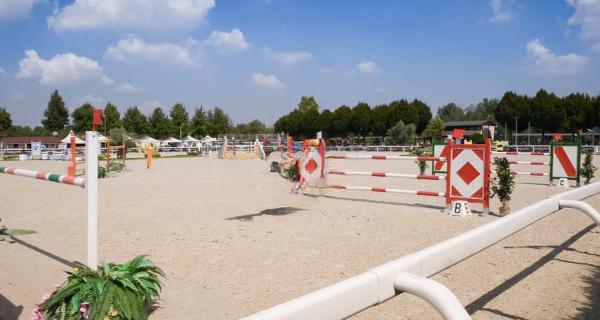} &       
     \includegraphics[width=0.162\linewidth]{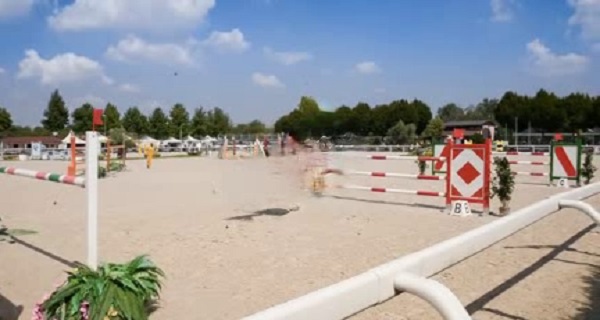} &     
     \includegraphics[width=0.162\linewidth]{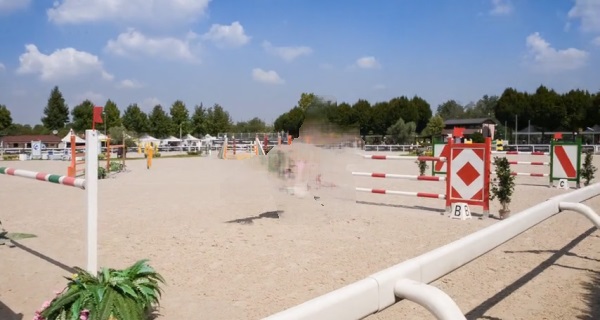}&      
     \includegraphics[width=0.162\linewidth]{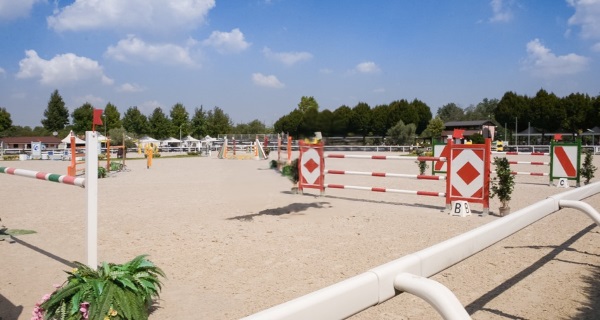} \\

     \includegraphics[width=0.162\linewidth]{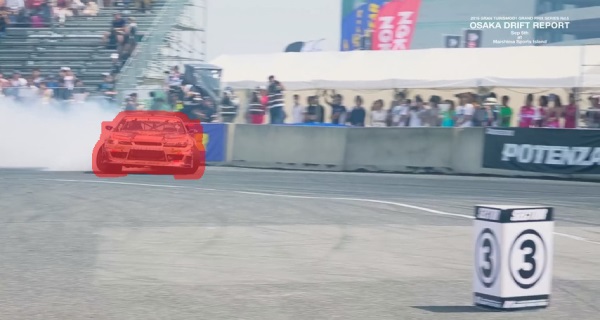} & 
     \includegraphics[width=0.162\linewidth]{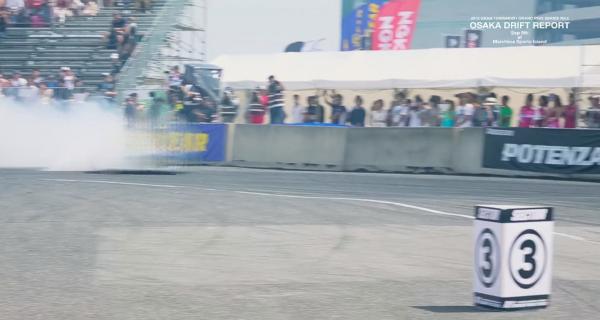} & 
     \includegraphics[width=0.162\linewidth]{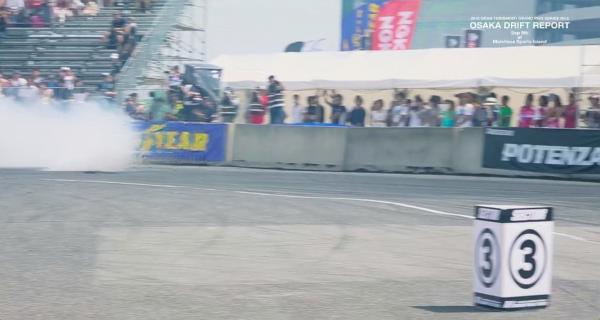} & 
     \includegraphics[width=0.162\linewidth]{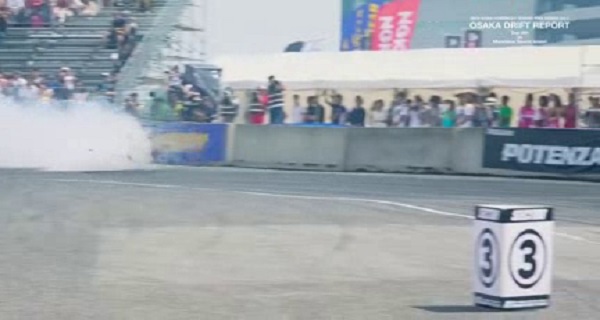} & 
     \includegraphics[width=0.162\linewidth]{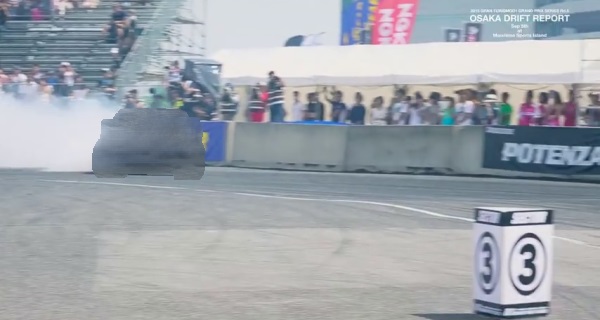} & 
     \includegraphics[width=0.162\linewidth]{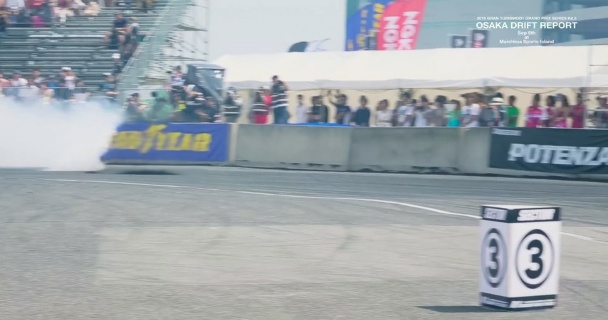} \\
     \includegraphics[width=0.162\linewidth]{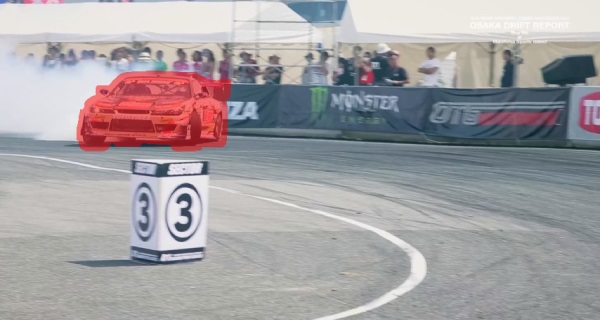} &      
     \includegraphics[width=0.162\linewidth]{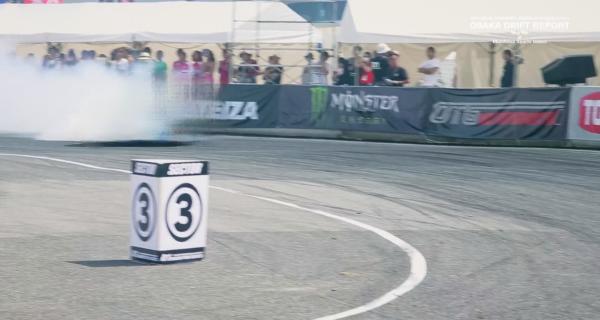} &    
     \includegraphics[width=0.162\linewidth]{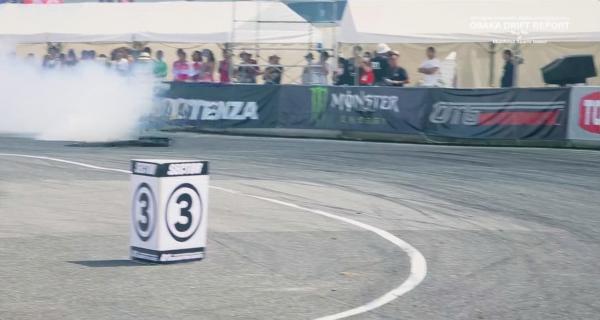} &       
     \includegraphics[width=0.162\linewidth]{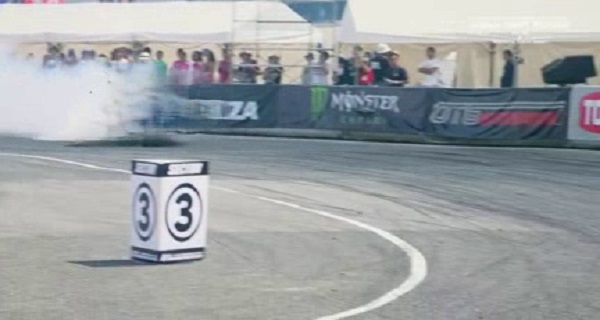} &     
     \includegraphics[width=0.162\linewidth]{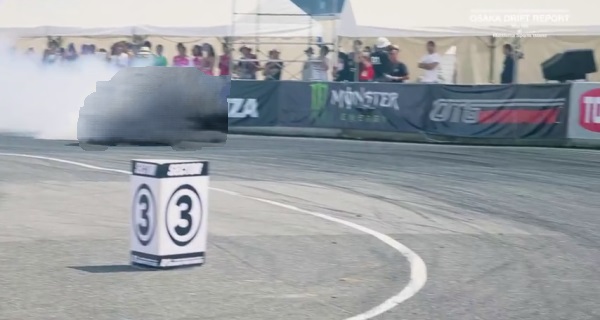}&      
     \includegraphics[width=0.162\linewidth]{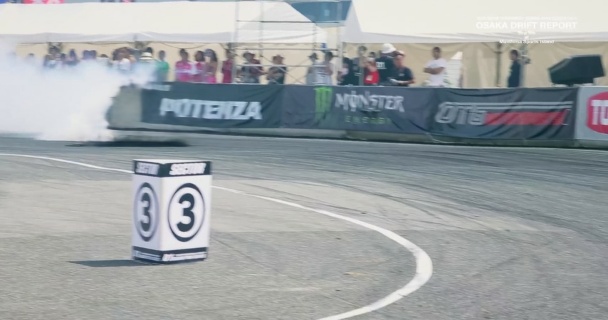} \\   
     
    \includegraphics[width=0.162\linewidth]{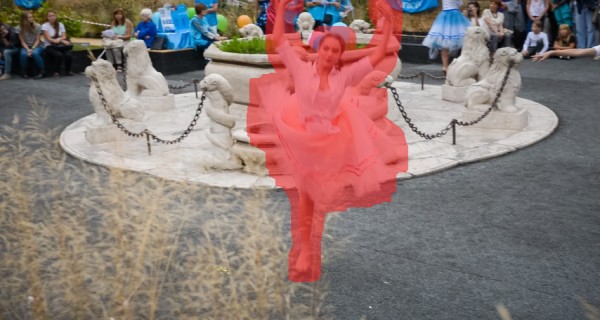} &      
     \includegraphics[width=0.162\linewidth]{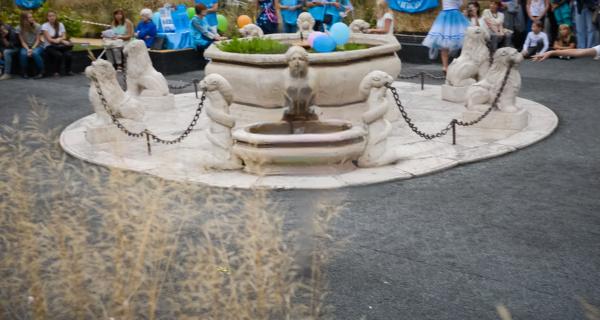} &    
     \includegraphics[width=0.162\linewidth]{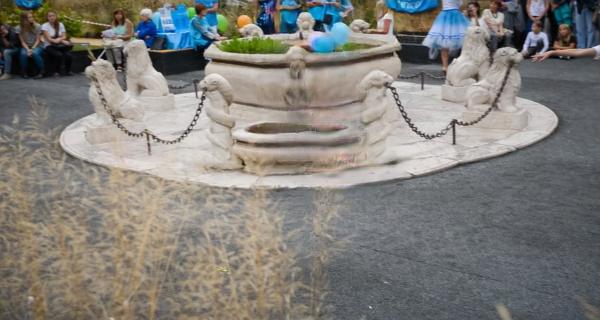} &       
     \includegraphics[width=0.162\linewidth]{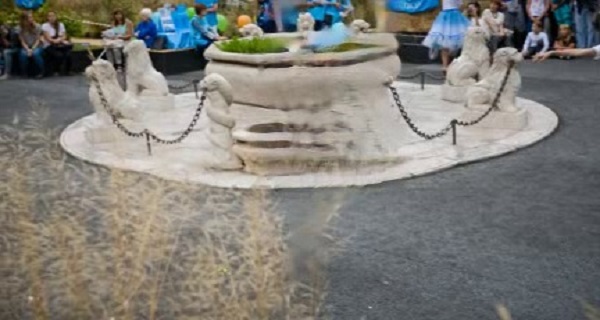} &     
     \includegraphics[width=0.162\linewidth]{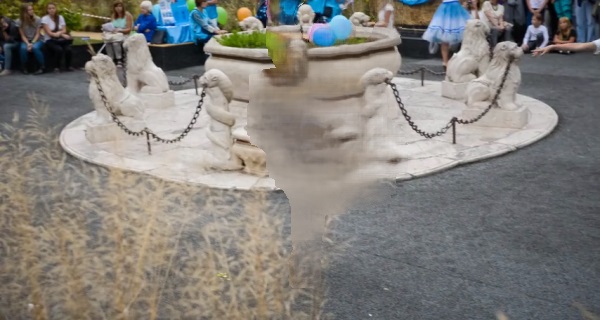}&      
     \includegraphics[width=0.162\linewidth]{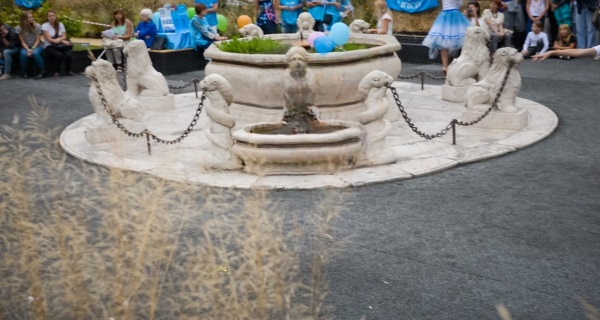} \\ 

     \includegraphics[width=0.162\linewidth]{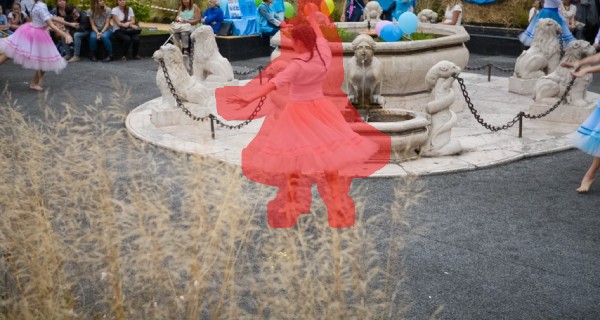} &      
     \includegraphics[width=0.162\linewidth]{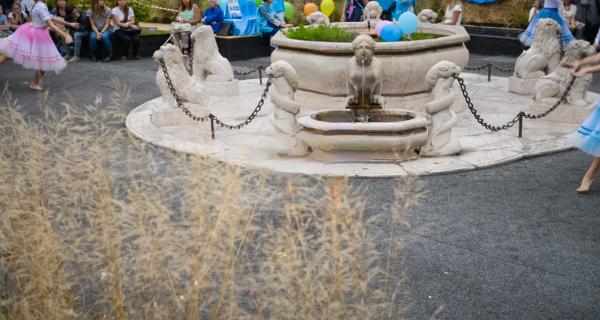} &    
     \includegraphics[width=0.162\linewidth]{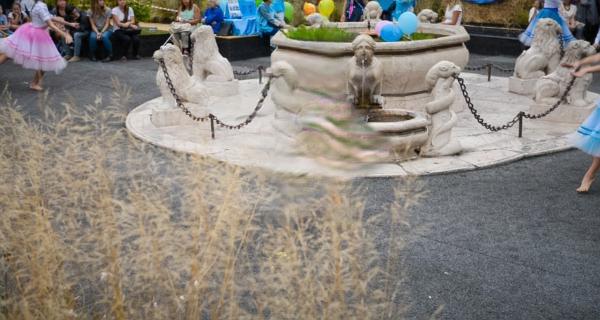} &       
     \includegraphics[width=0.162\linewidth]{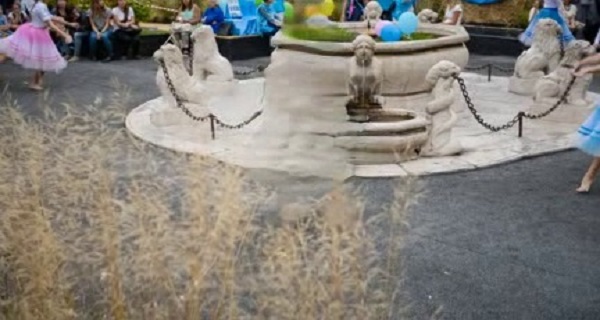} &     
     \includegraphics[width=0.162\linewidth]{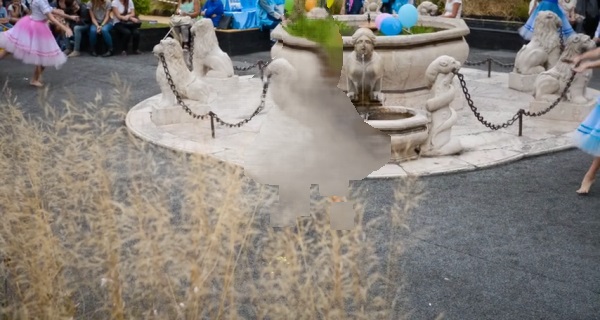}&      
     \includegraphics[width=0.162\linewidth]{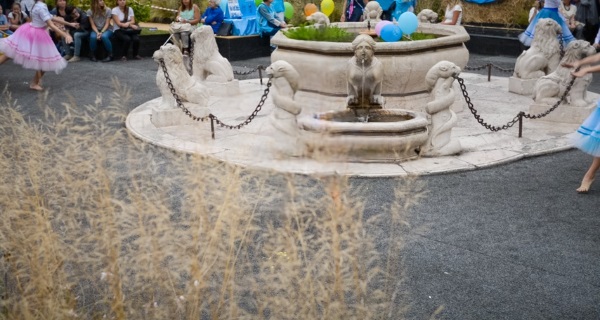} \\

     \includegraphics[width=0.162\linewidth]{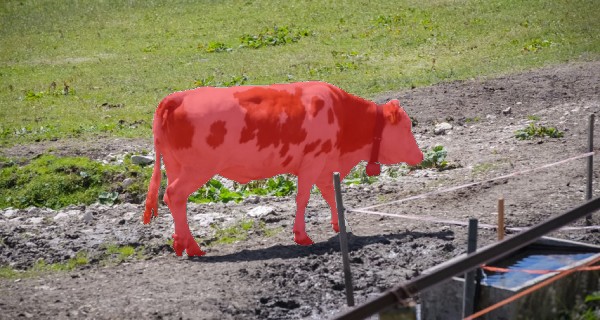} &      
     \includegraphics[width=0.162\linewidth]{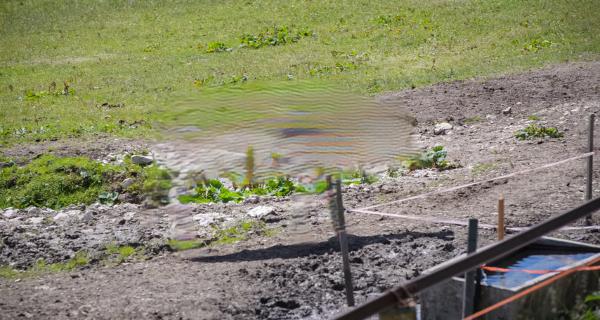} &    
     \includegraphics[width=0.162\linewidth]{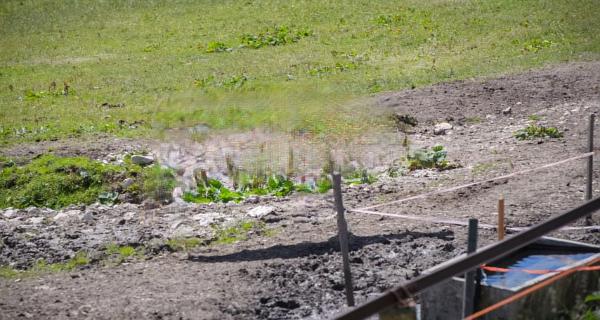} &       
     \includegraphics[width=0.162\linewidth]{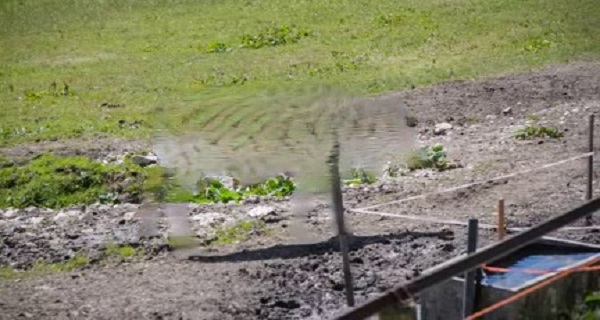} &     
     \includegraphics[width=0.162\linewidth]{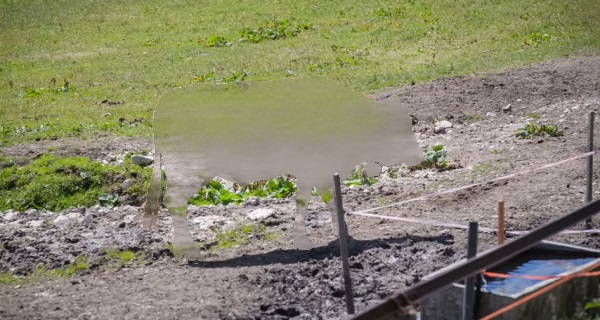}&      
     \includegraphics[width=0.162\linewidth]{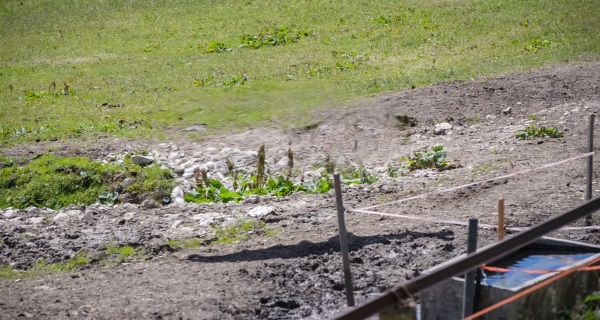} \\    
 
     \includegraphics[width=0.162\linewidth]{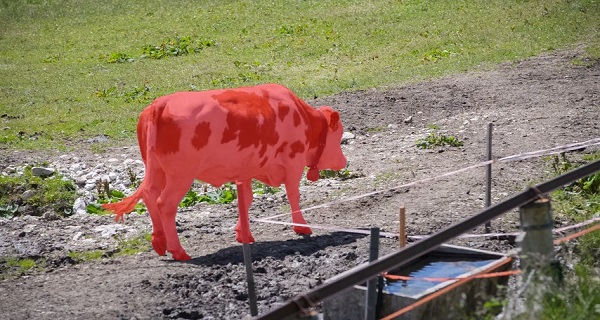} &      
     \includegraphics[width=0.162\linewidth]{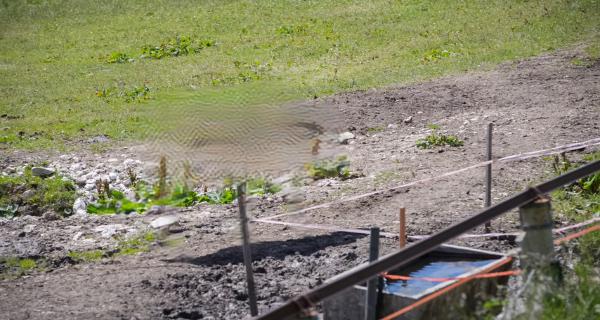} &    
     \includegraphics[width=0.162\linewidth]{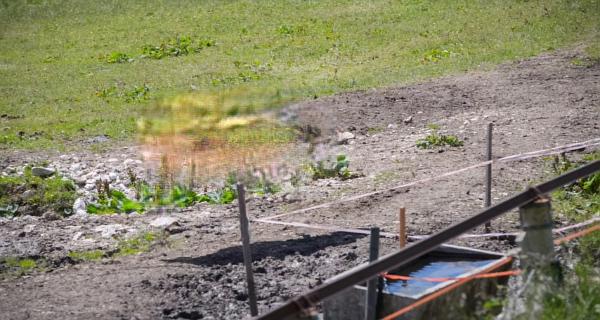} &       
     \includegraphics[width=0.162\linewidth]{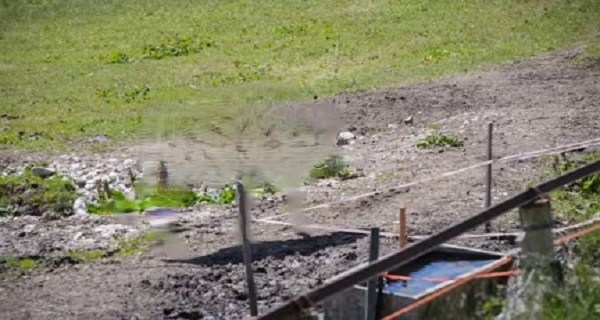} &     
     \includegraphics[width=0.162\linewidth]{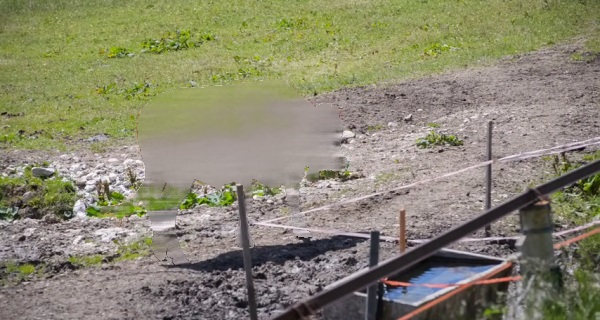}&      
     \includegraphics[width=0.162\linewidth]{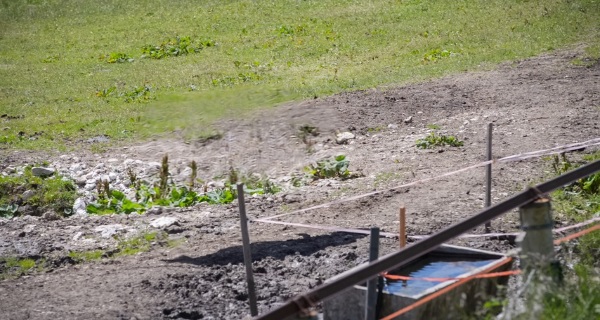} \\    
     \includegraphics[width=0.162\linewidth]{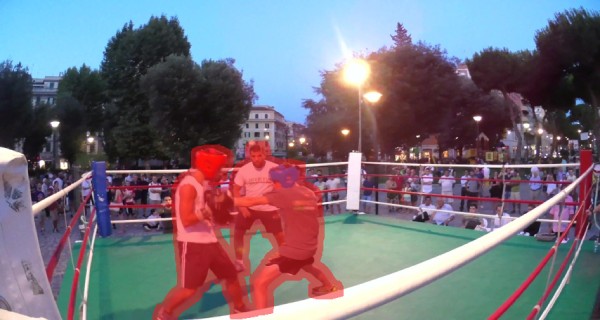}  &
     \includegraphics[width=0.162\linewidth]{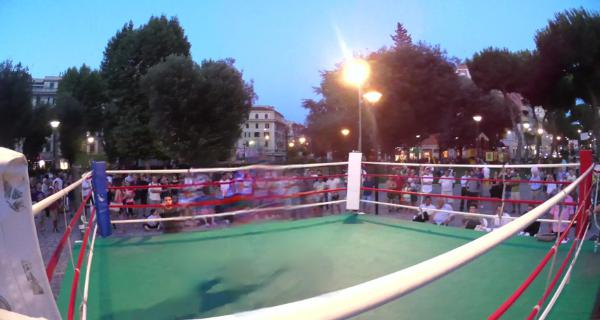}  &
     \includegraphics[width=0.162\linewidth]{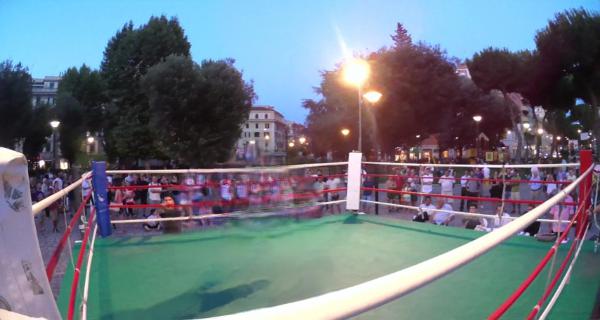}  &
     \includegraphics[width=0.162\linewidth]{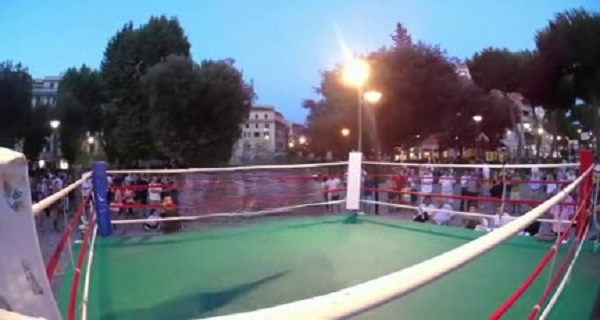}  &
     \includegraphics[width=0.162\linewidth]{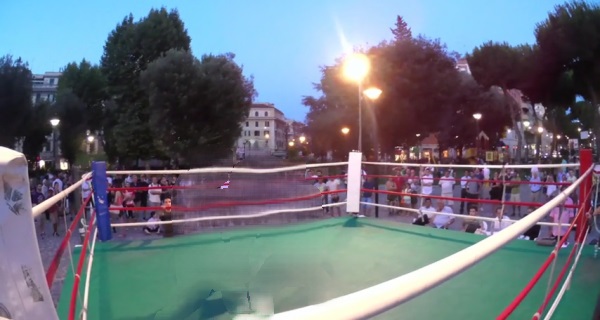}  &
     \includegraphics[width=0.162\linewidth]{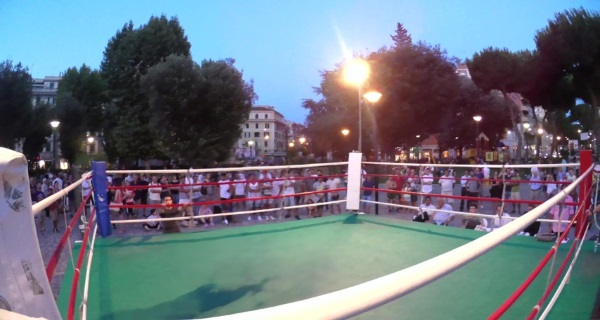}  \\
     
     \includegraphics[width=0.162\linewidth]{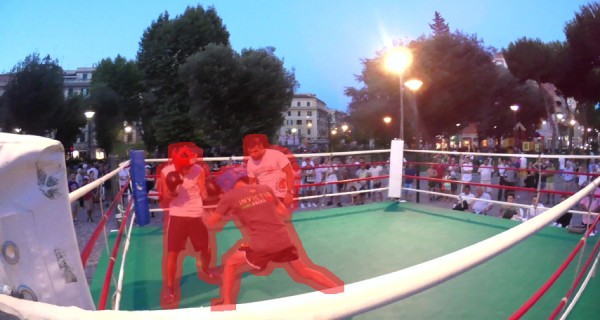} &   
     \includegraphics[width=0.162\linewidth]{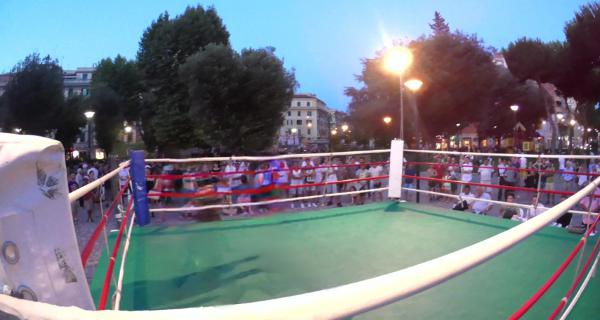} & 
     \includegraphics[width=0.162\linewidth]{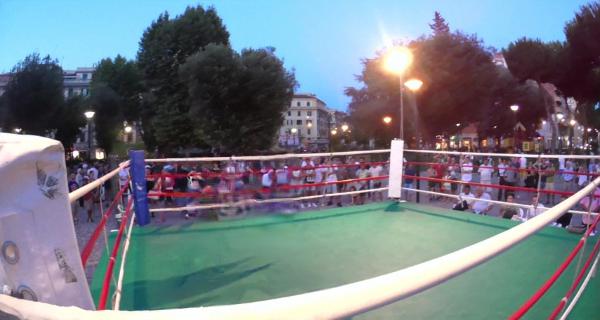} & 
     \includegraphics[width=0.162\linewidth]{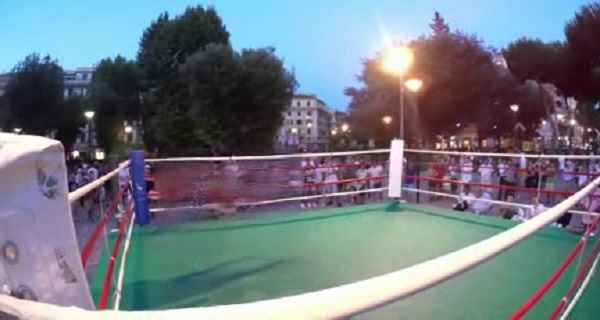} & 
     \includegraphics[width=0.162\linewidth]{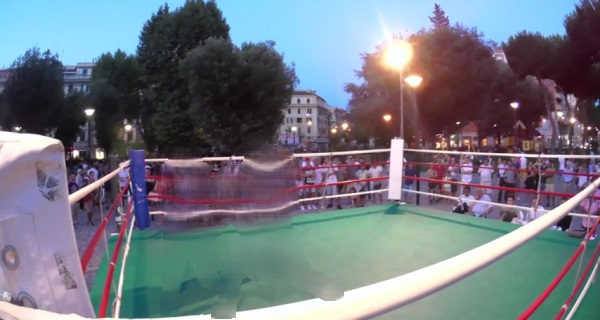} & 
     \includegraphics[width=0.162\linewidth]{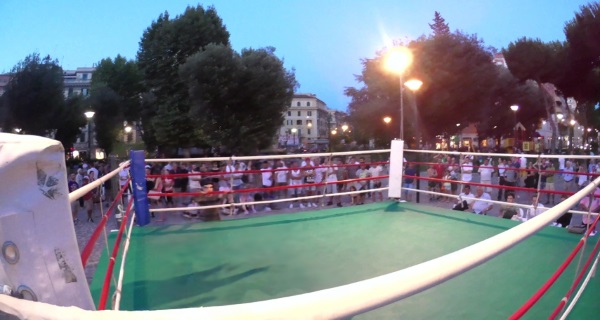}\\         
     \includegraphics[width=0.162\linewidth]{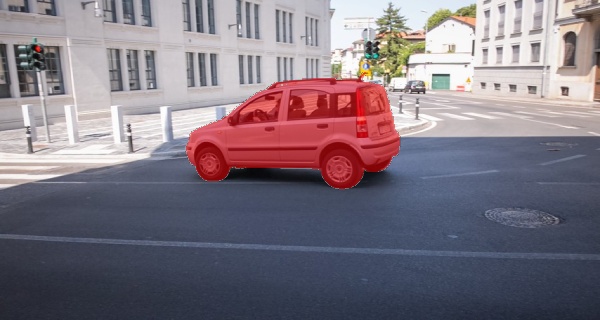} &      
     \includegraphics[width=0.162\linewidth]{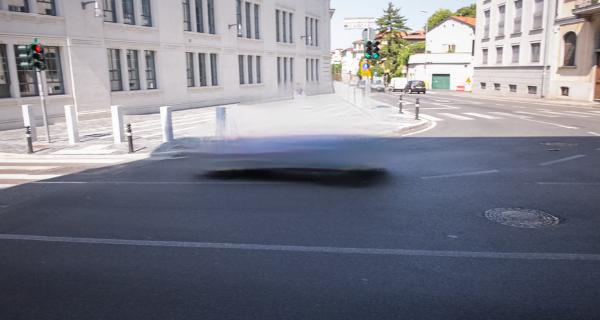} &    
     \includegraphics[width=0.162\linewidth]{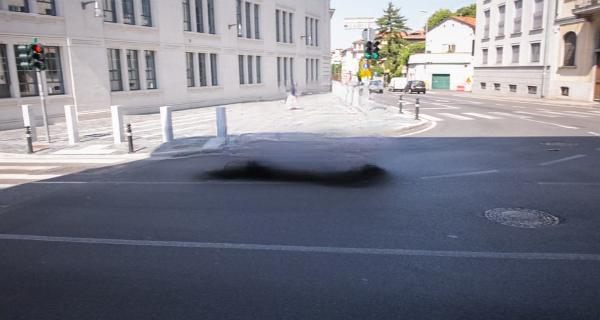} &       
     \includegraphics[width=0.162\linewidth]{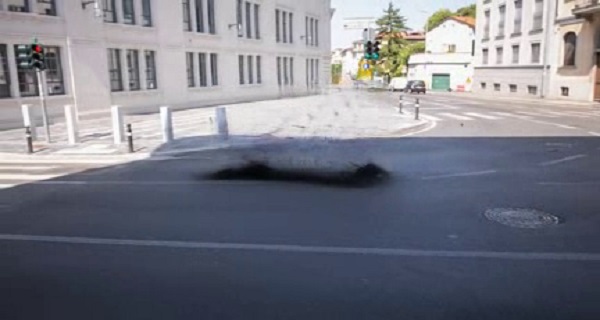} &     
     \includegraphics[width=0.162\linewidth]{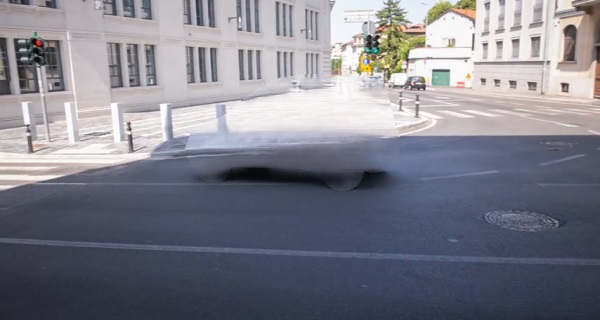}&      
     \includegraphics[width=0.162\linewidth]{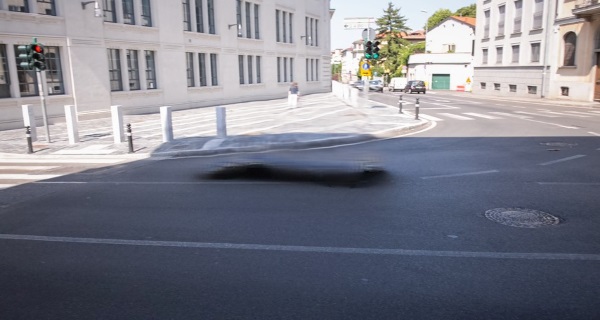} \\        
     \includegraphics[width=0.162\linewidth]{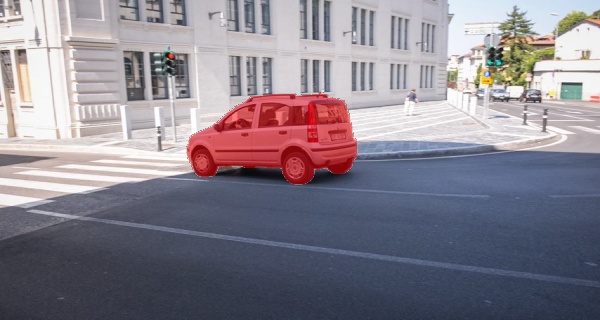} &      
     \includegraphics[width=0.162\linewidth]{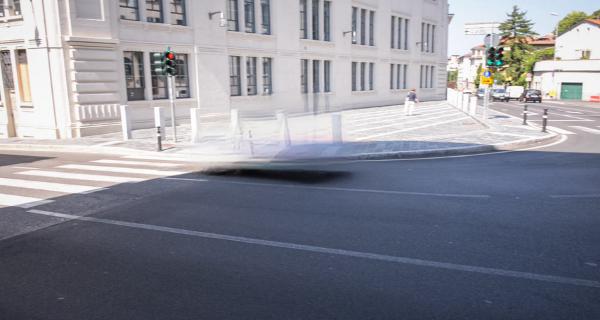} &    
     \includegraphics[width=0.162\linewidth]{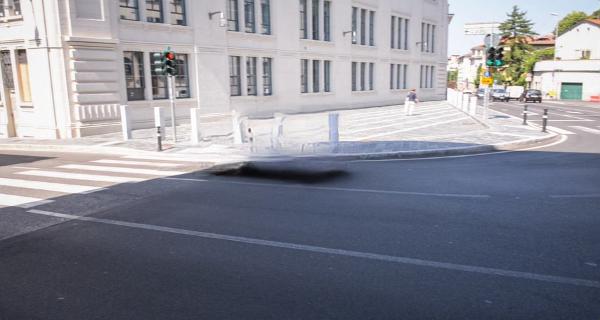} &       
     \includegraphics[width=0.162\linewidth]{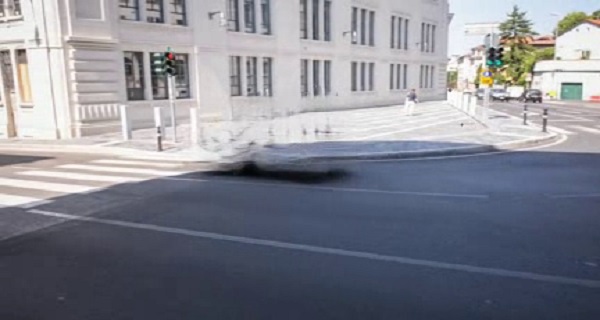} &     
     \includegraphics[width=0.162\linewidth]{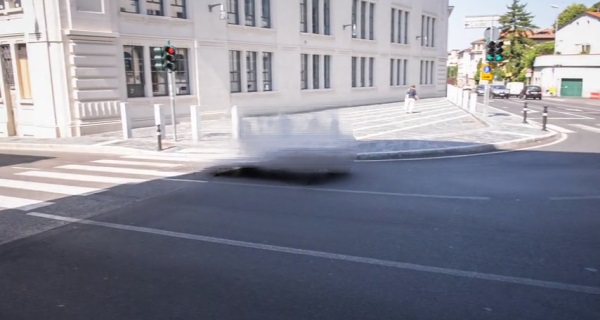}&      
     \includegraphics[width=0.162\linewidth]{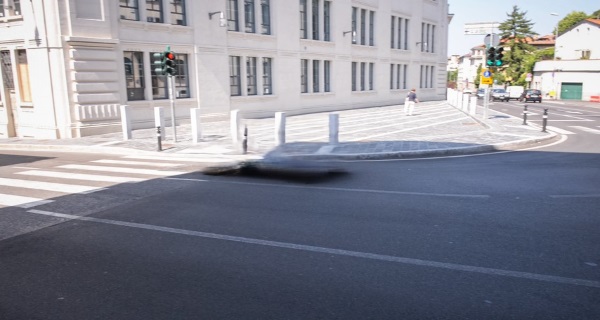} \\   
     
      \includegraphics[width=0.162\linewidth]{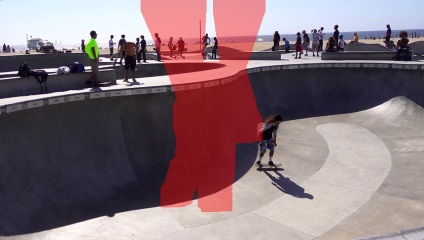} & 
     \includegraphics[width=0.162\linewidth]{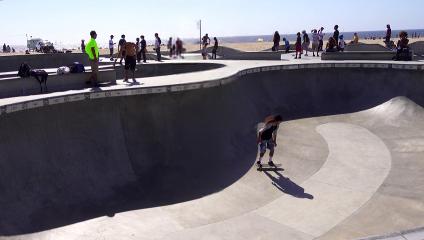} & 
     \includegraphics[width=0.162\linewidth]{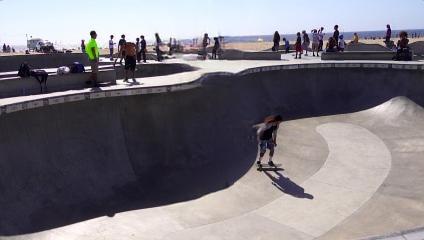} & 
     \includegraphics[width=0.162\linewidth]{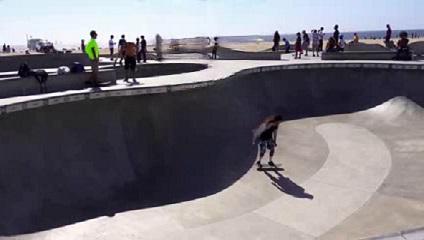} & 
     \includegraphics[width=0.162\linewidth]{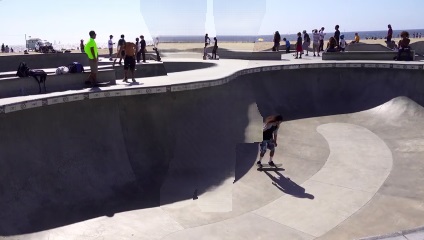} & 
     \includegraphics[width=0.162\linewidth]{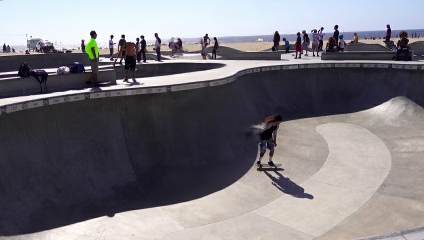} \\
     \includegraphics[width=0.162\linewidth]{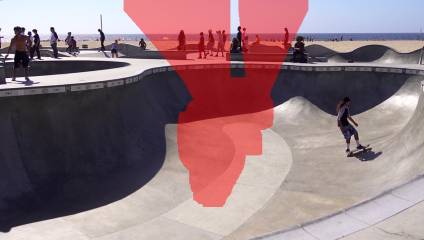} &      
     \includegraphics[width=0.162\linewidth]{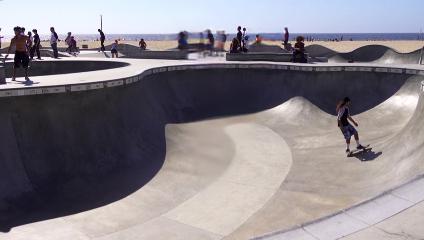} &    
     \includegraphics[width=0.162\linewidth]{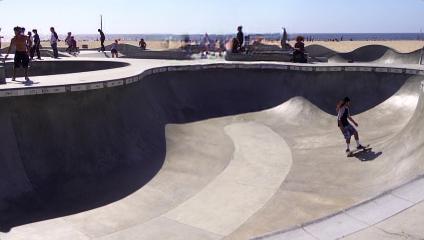} &       
     \includegraphics[width=0.162\linewidth]{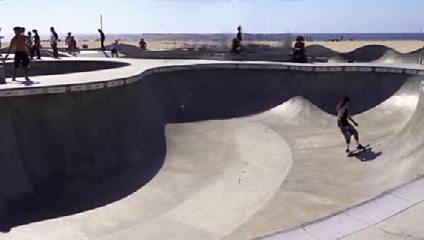} &     
     \includegraphics[width=0.162\linewidth]{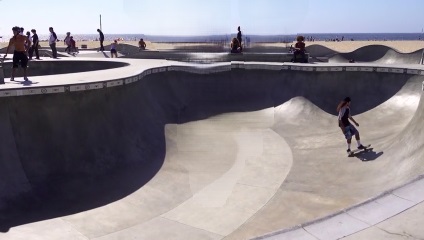}&      
     \includegraphics[width=0.162\linewidth]{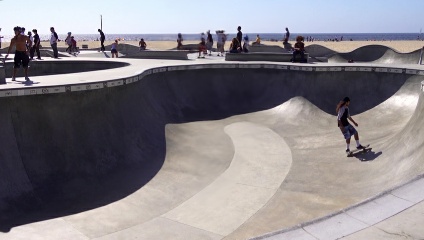} \\          
     
     Input&CAP~\cite{lee2019copy}&OPN~\cite{oh2019onion}&STTN~\cite{zeng2020learning}&InterVI~\cite{zhang2019internal}&Ours
    \end{tabular}
   \caption{Visual comparisons of object removal on the DAVIS dataset. Zoom in for details.}
    \label{fig:qualitative-supp}
\end{figure*}

\textbf{Results on DAVIS} More results on DAVIS  are shown in Fig.~\ref{fig:qualitative-supp}.  

\begin{figure*}
    \centering 
    \small
    \begin{tabular}{@{}c@{\hspace{0.3mm}}c@{\hspace{0.3mm}}c@{\hspace{0.3mm}}c@{\hspace{0.3mm}}c@{\hspace{0.3mm}}c@{}}
     
      \includegraphics[width=0.162\linewidth]{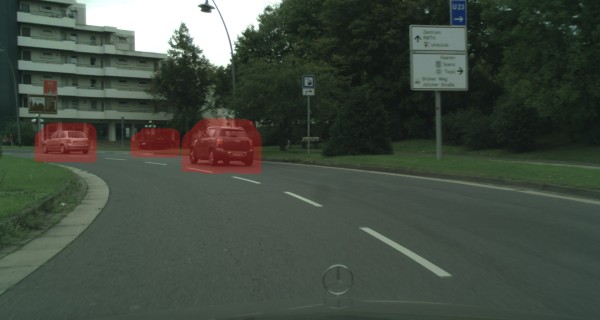} &      
     \includegraphics[width=0.162\linewidth]{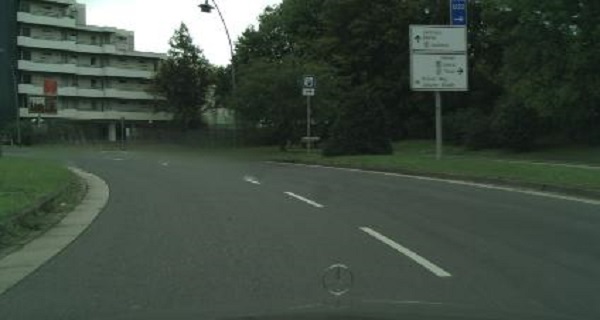} &    
     \includegraphics[width=0.162\linewidth]{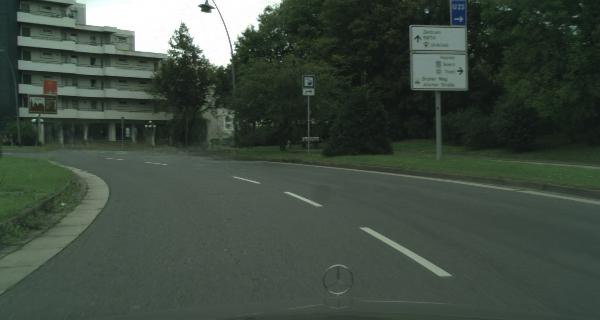} &       
     \includegraphics[width=0.162\linewidth]{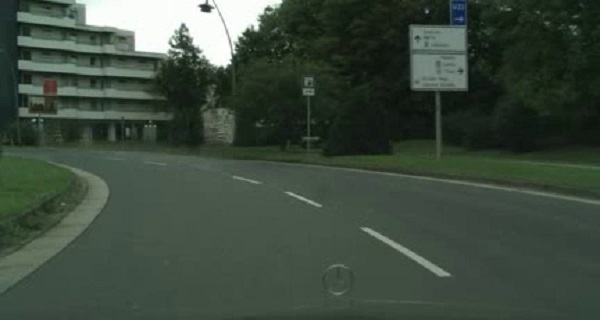} &     
     \includegraphics[width=0.162\linewidth]{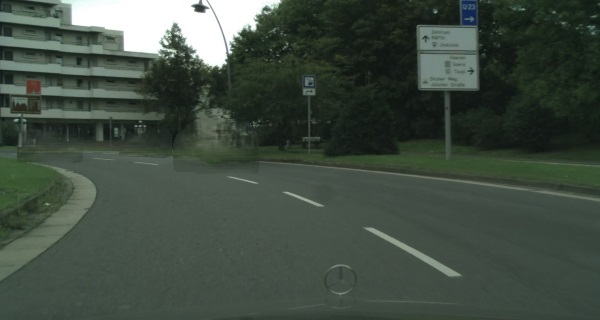}&      
     \includegraphics[width=0.162\linewidth]{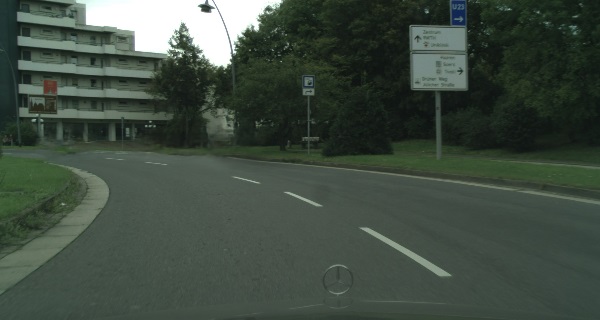} \\      
     \includegraphics[width=0.162\linewidth]{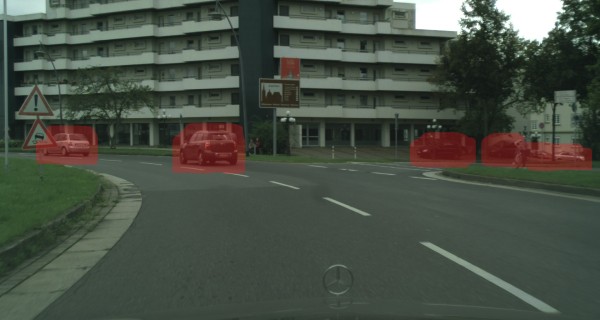} &      
     \includegraphics[width=0.162\linewidth]{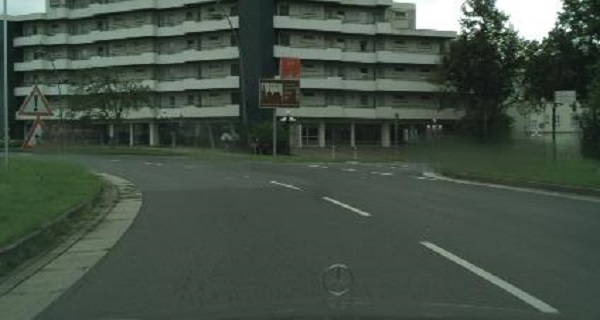} &    
     \includegraphics[width=0.162\linewidth]{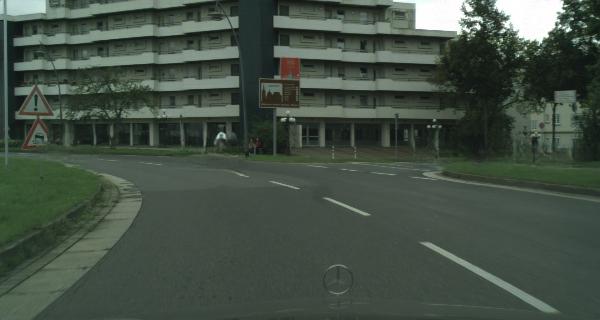} &       
     \includegraphics[width=0.162\linewidth]{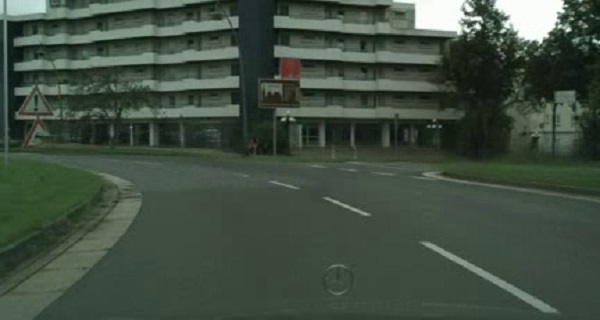} &     
     \includegraphics[width=0.162\linewidth]{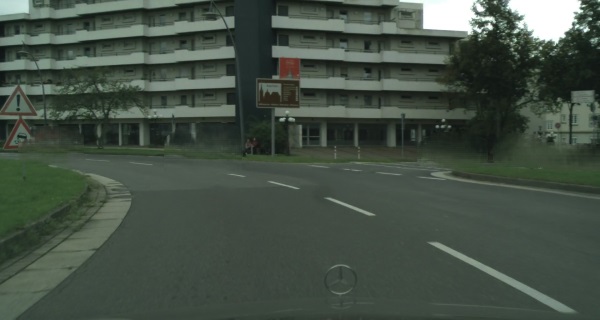}&      
     \includegraphics[width=0.162\linewidth]{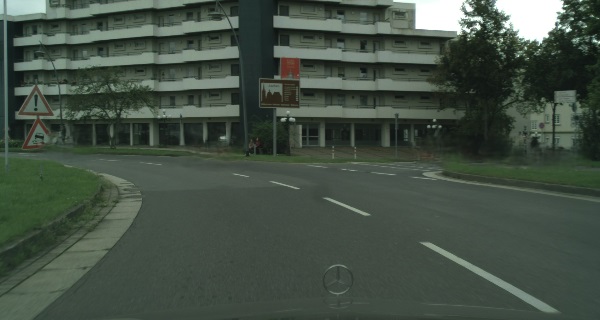} \\    
      
     \includegraphics[width=0.162\linewidth]{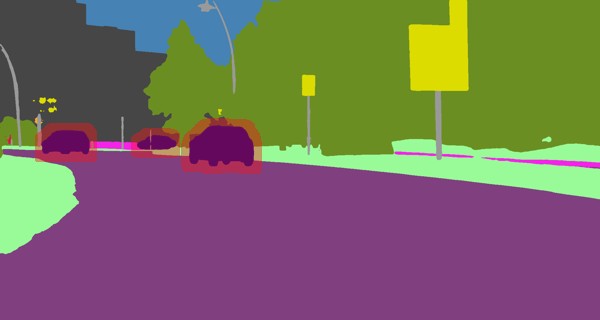} &      
     \includegraphics[width=0.162\linewidth]{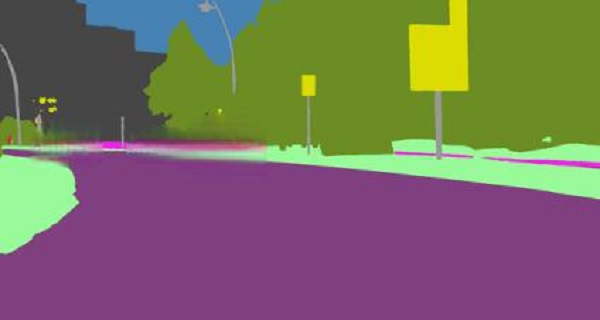} &    
     \includegraphics[width=0.162\linewidth]{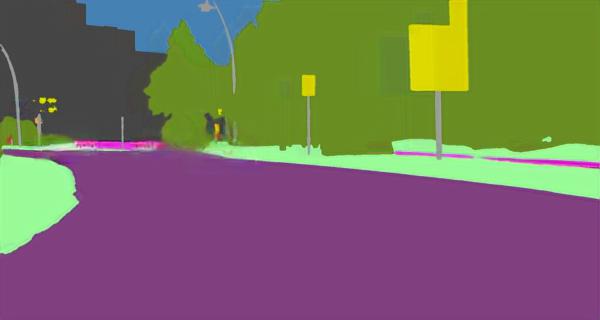} &       
     \includegraphics[width=0.162\linewidth]{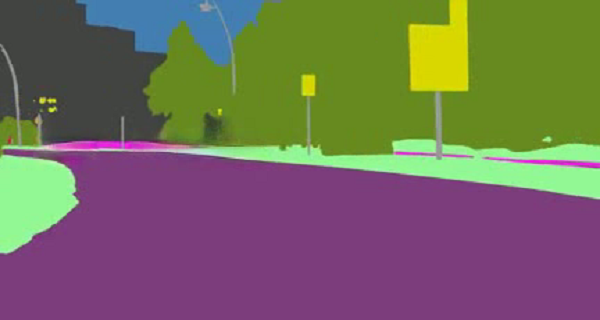} &     
     \includegraphics[width=0.162\linewidth]{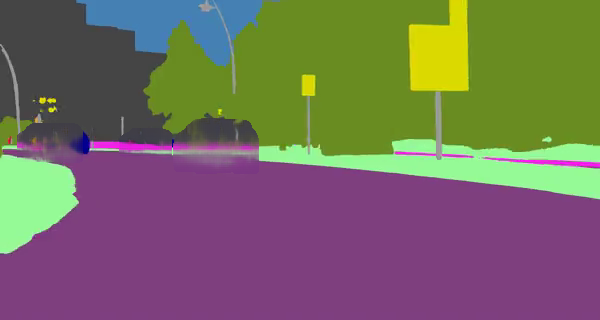}&      
     \includegraphics[width=0.162\linewidth]{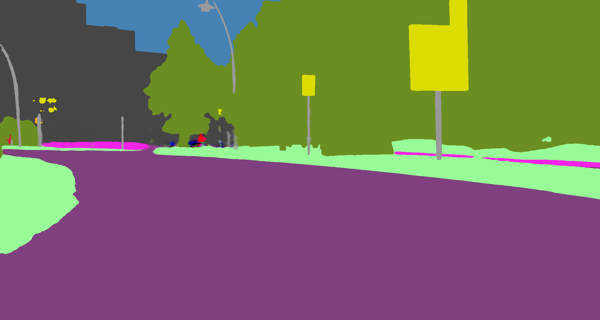} \\   
         \includegraphics[width=0.162\linewidth]{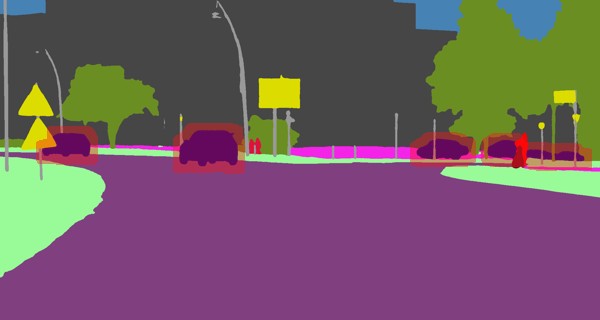} &      
     \includegraphics[width=0.162\linewidth]{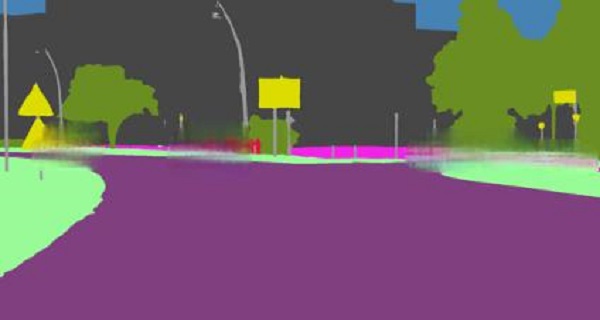} &    
     \includegraphics[width=0.162\linewidth]{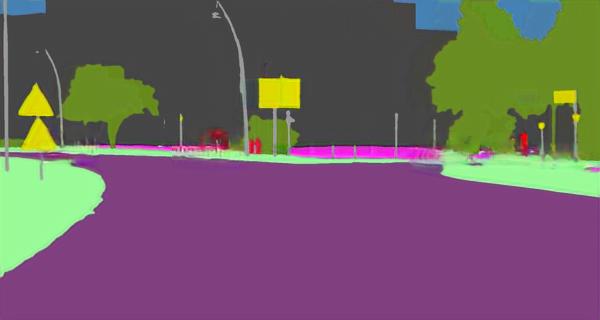} &       
     \includegraphics[width=0.162\linewidth]{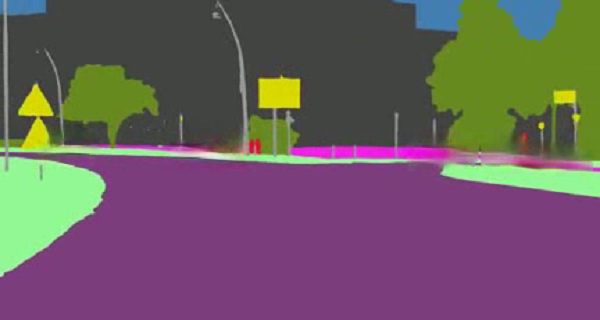} &     
     \includegraphics[width=0.162\linewidth]{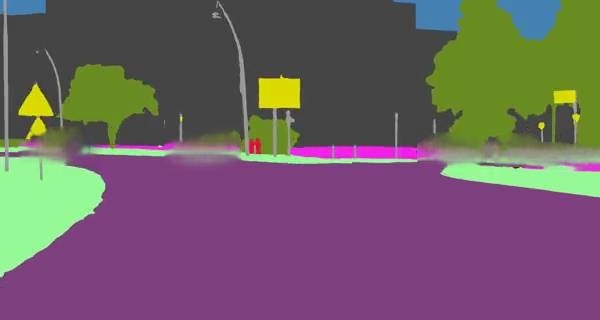}&      
     \includegraphics[width=0.162\linewidth]{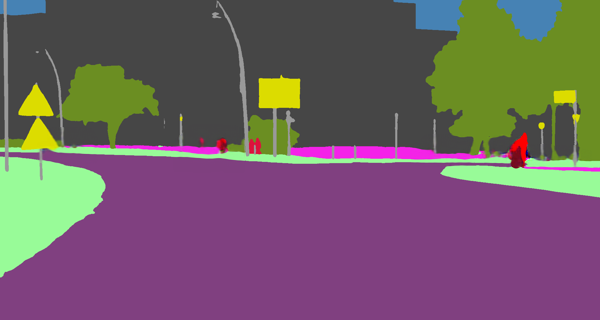} \\   
     
     \includegraphics[width=0.162\linewidth]{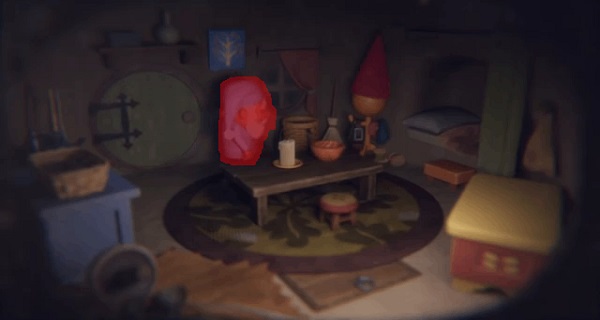} &      
     \includegraphics[width=0.162\linewidth]{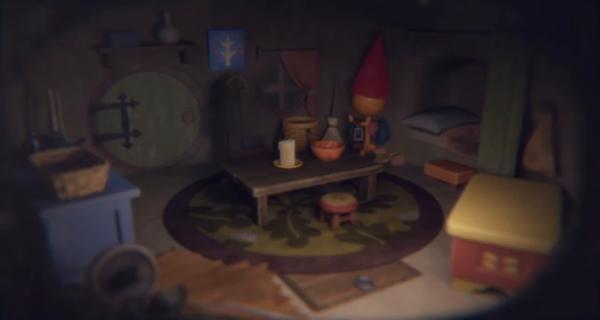} &    
     \includegraphics[width=0.162\linewidth]{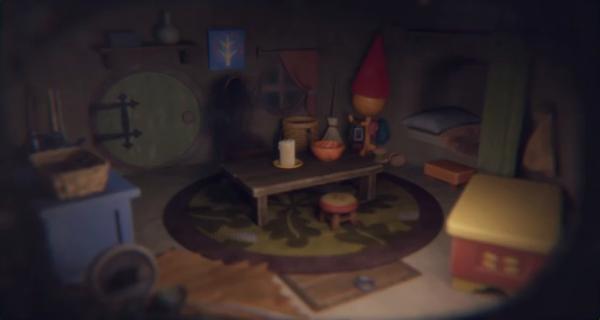} &       
     \includegraphics[width=0.162\linewidth]{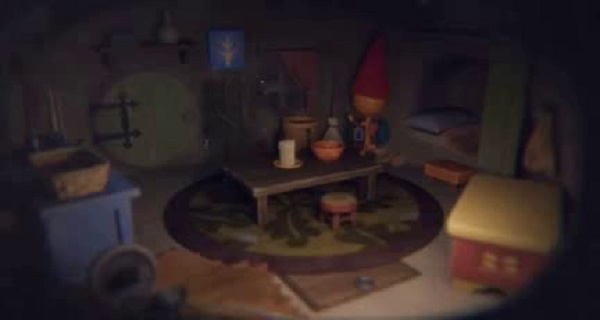} &     
     \includegraphics[width=0.162\linewidth]{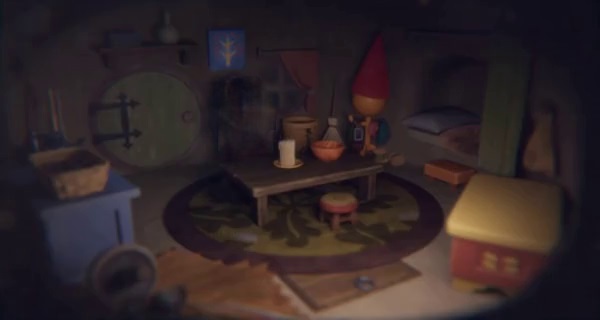}&      
     \includegraphics[width=0.162\linewidth]{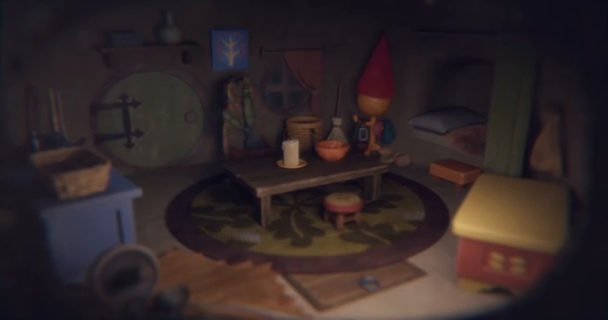} \\         
     \includegraphics[width=0.162\linewidth]{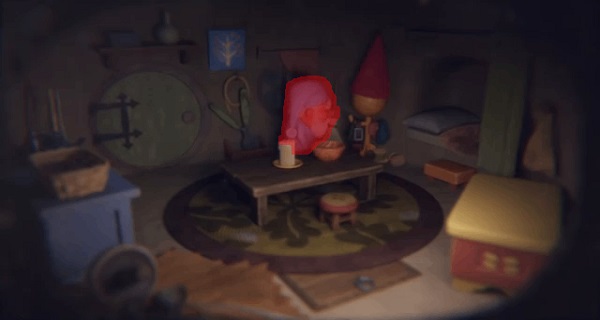} &      
     \includegraphics[width=0.162\linewidth]{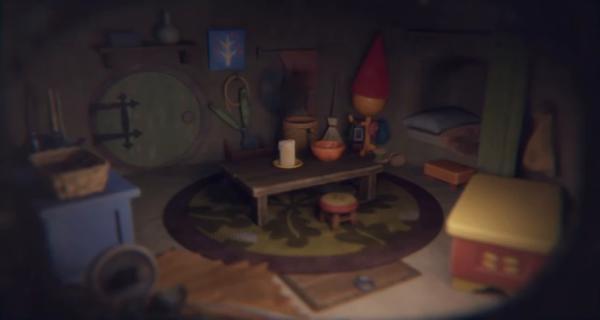} &    
     \includegraphics[width=0.162\linewidth]{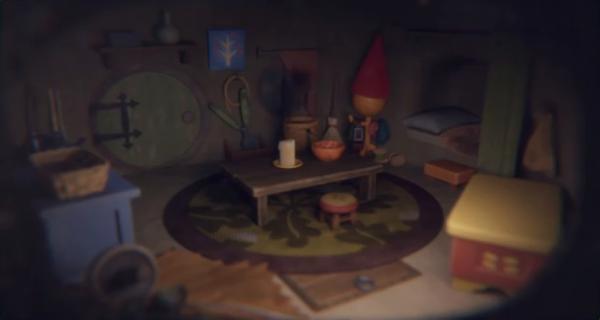} &       
     \includegraphics[width=0.162\linewidth]{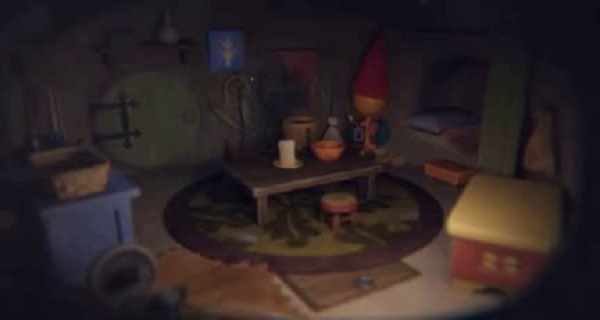} &     
     \includegraphics[width=0.162\linewidth]{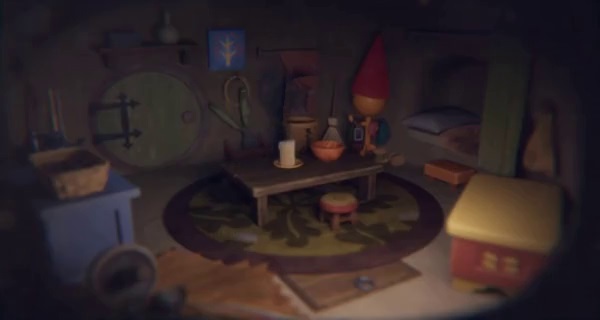}&      
     \includegraphics[width=0.162\linewidth]{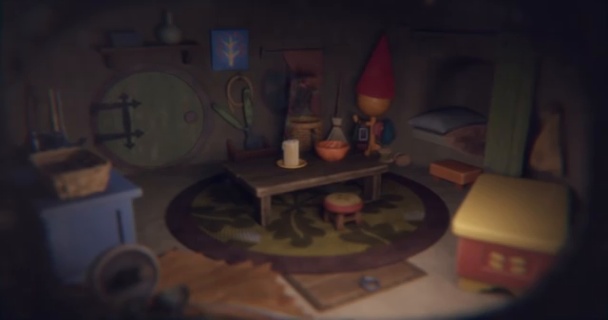} \\         
     Input&CAP &OPN  &STTN &InterVI &Ours
    \end{tabular}
   \caption{Visual comparisons on different domains.  We show inpainting results on autonomous driving videos, semantic map sequences, and anime videos. Zoom in for details.}
    \label{fig:domain}
\end{figure*}

\textbf{Results on multiple domains} As we mention in Section 1 in the main paper, the performance of the externally trained methods suffer  degradation when the domain gap is huge. In Fig.~\ref{fig:domain}, we show the comparison of the cross-domain test results. \\

\begin{figure*}
    \centering 
    \scriptsize
    \begin{tabular}{@{}c@{\hspace{0.3mm}}c@{\hspace{0.3mm}}c@{\hspace{0.3mm}}c@{\hspace{0.3mm}}c@{\hspace{0.3mm}}c@{\hspace{0.3mm}}c@{}}
    
     \rotatebox{90}{~Input frame}& \includegraphics[width=0.162\linewidth]{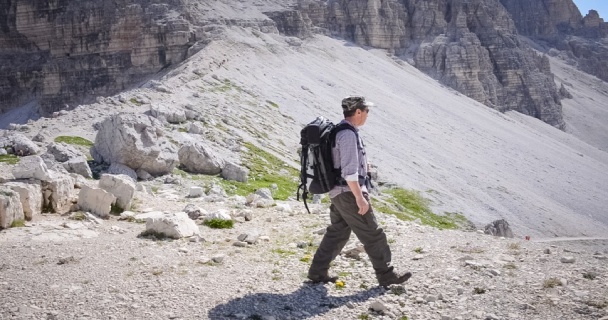} &      
      \includegraphics[width=0.162\linewidth]{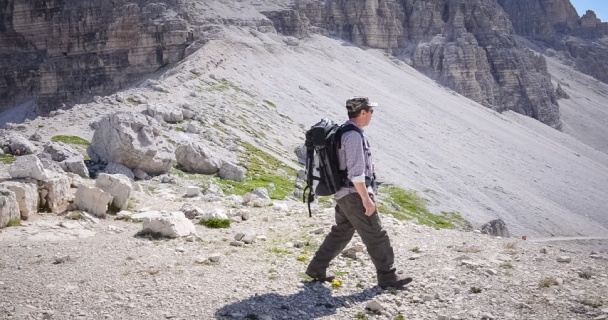} &    
       \includegraphics[width=0.162\linewidth]{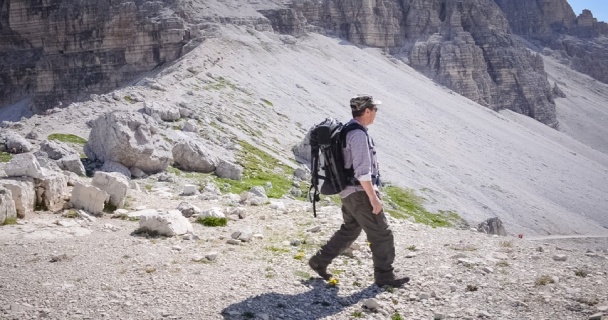} &    
       \includegraphics[width=0.162\linewidth]{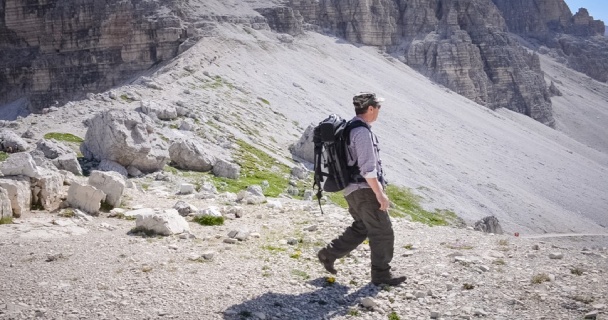} & 
       \includegraphics[width=0.162\linewidth]{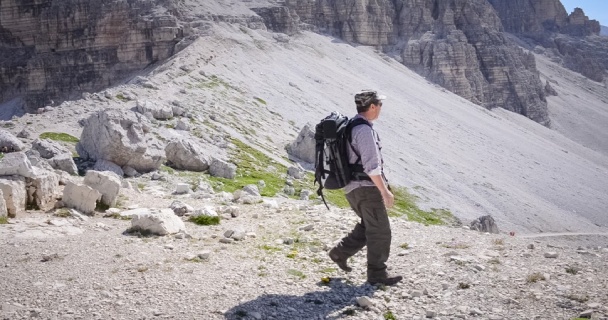} & 
       \includegraphics[width=0.162\linewidth]{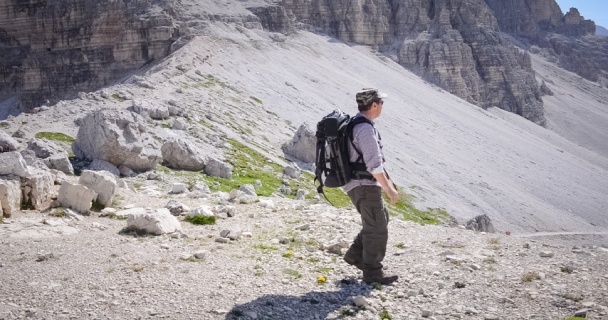} \\
       \rotatebox{90}{Predicted mask}&\includegraphics[width=0.162\linewidth]{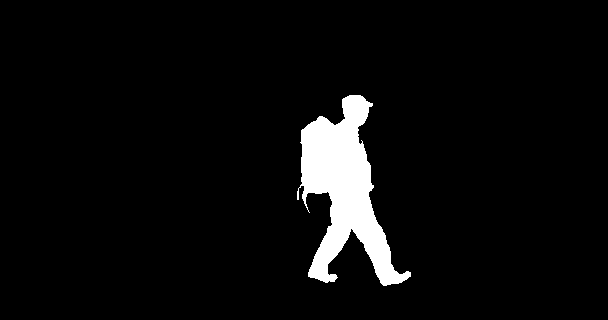} &   
       \includegraphics[width=0.162\linewidth]{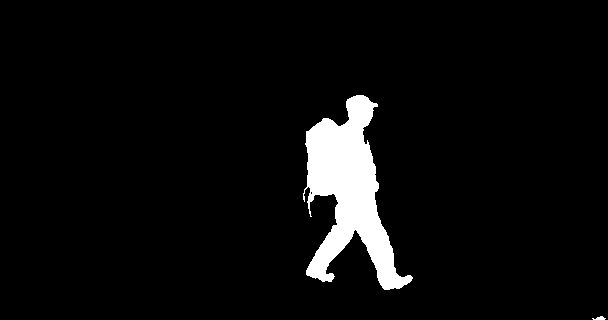} &  
       \includegraphics[width=0.162\linewidth]{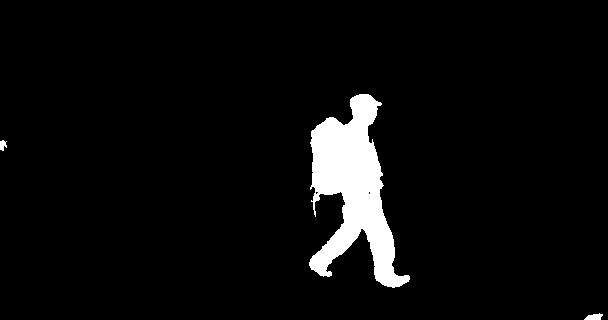} &  
      \includegraphics[width=0.162\linewidth]{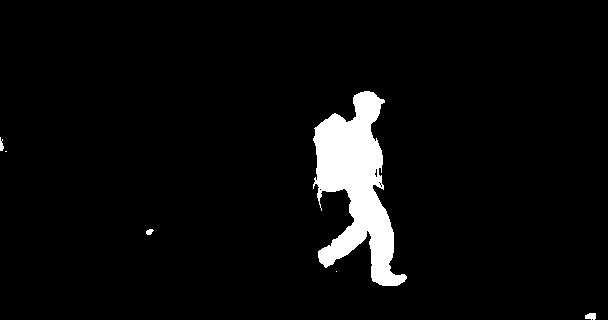} &  
       \includegraphics[width=0.162\linewidth]{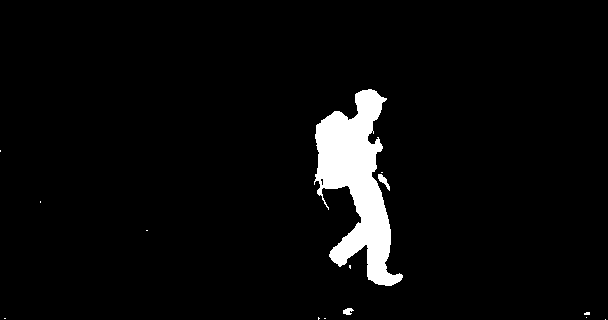} &  
       \includegraphics[width=0.162\linewidth]{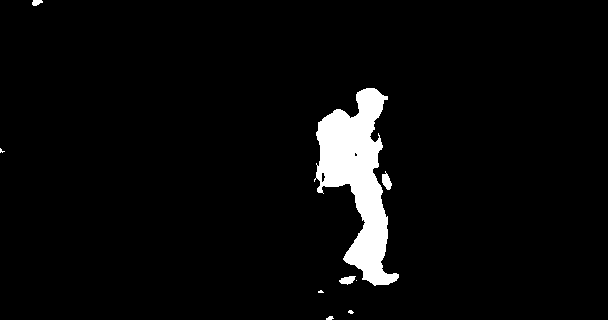}  \\
       
     \rotatebox{90}{~Input frame}& \includegraphics[width=0.162\linewidth]{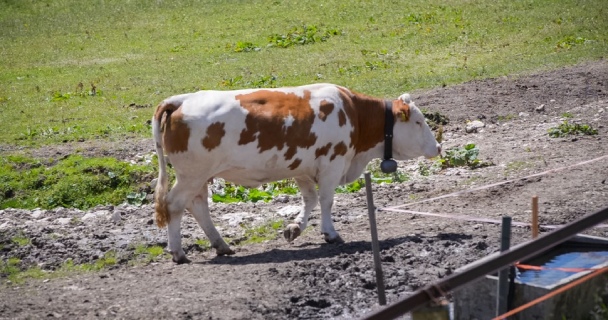} &      
      \includegraphics[width=0.162\linewidth]{latex/fig/extension/seg_more/cow/img/7.jpg} &    
       \includegraphics[width=0.162\linewidth]{latex/fig/extension/seg_more/cow/img/13.jpg} &    
       \includegraphics[width=0.162\linewidth]{latex/fig/extension/seg_more/cow/img/19.jpg} & 
       \includegraphics[width=0.162\linewidth]{latex/fig/extension/seg_more/cow/img/25.jpg} & 
       \includegraphics[width=0.162\linewidth]{latex/fig/extension/seg_more/cow/img/31.jpg} \\
       \rotatebox{90}{Predicted mask}&\includegraphics[width=0.162\linewidth]{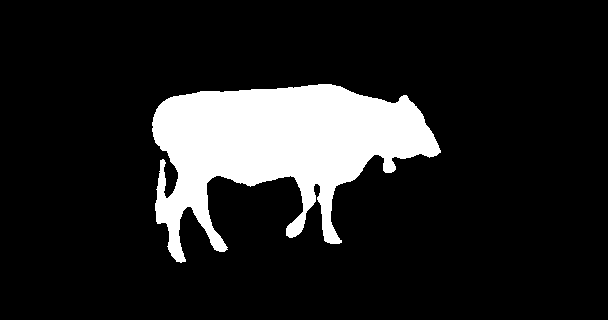} &   
       \includegraphics[width=0.162\linewidth]{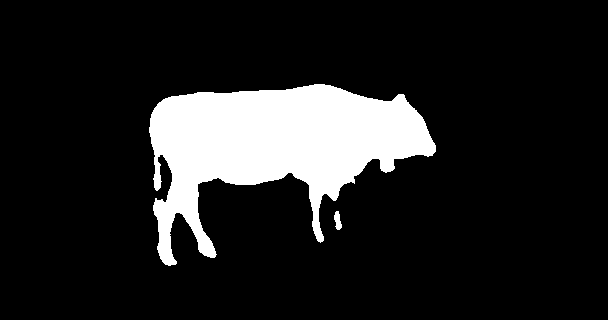} &  
       \includegraphics[width=0.162\linewidth]{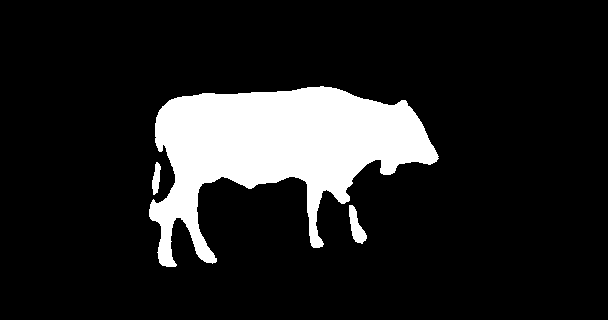} &  
      \includegraphics[width=0.162\linewidth]{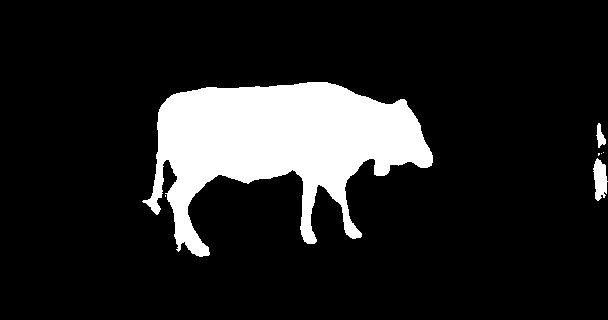} &  
       \includegraphics[width=0.162\linewidth]{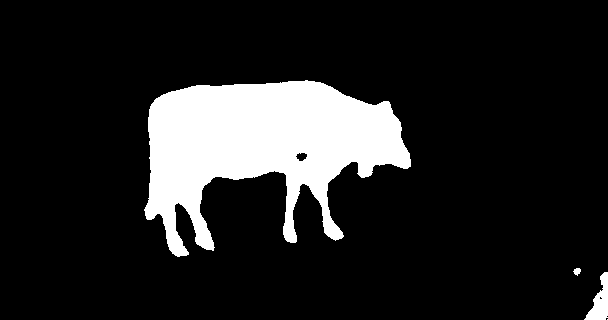} &  
       \includegraphics[width=0.162\linewidth]{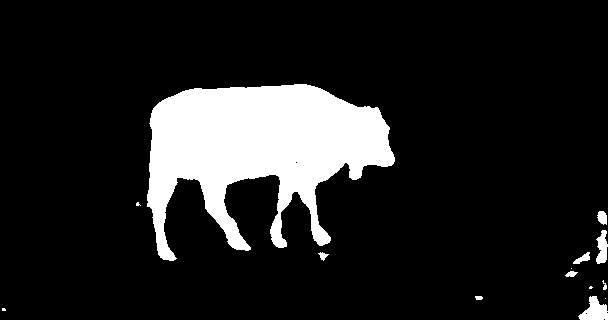}  \\  

     \rotatebox{90}{~Input frame}& \includegraphics[width=0.162\linewidth]{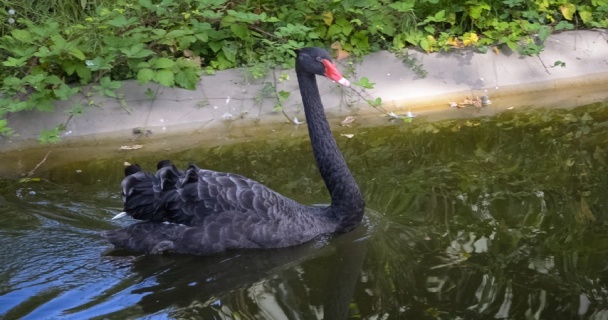} &      
      \includegraphics[width=0.162\linewidth]{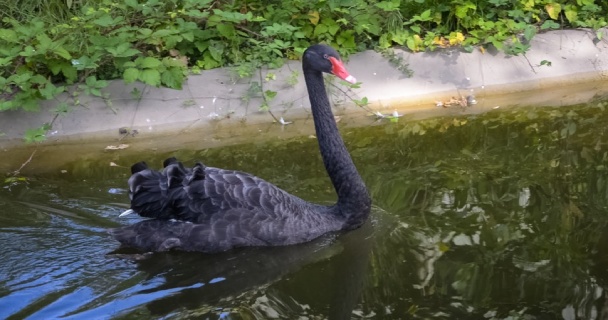} &    
       \includegraphics[width=0.162\linewidth]{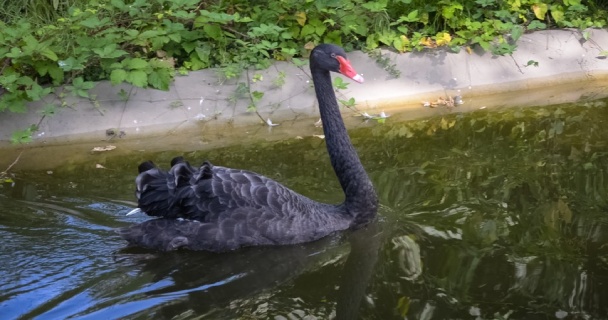} &    
       \includegraphics[width=0.162\linewidth]{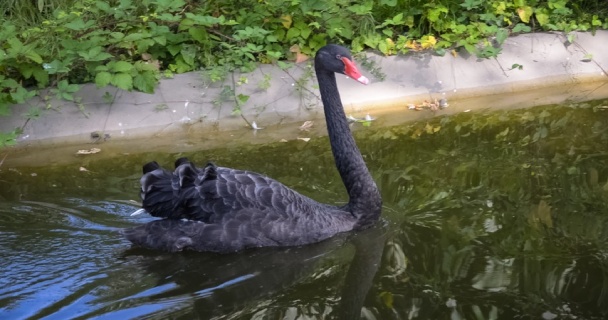} & 
       \includegraphics[width=0.162\linewidth]{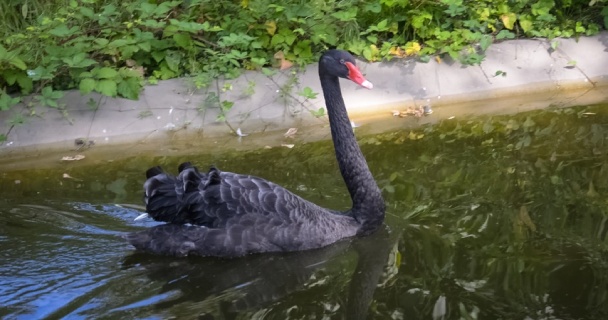} & 
       \includegraphics[width=0.162\linewidth]{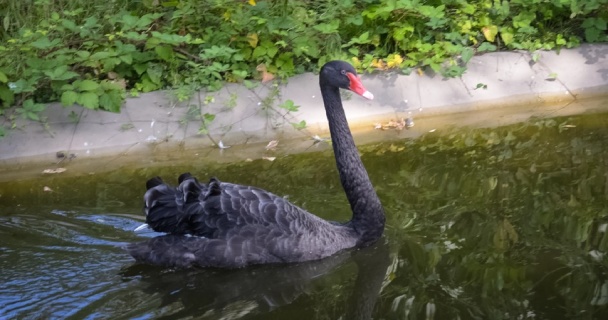} \\
       \rotatebox{90}{Predicted mask}&\includegraphics[width=0.162\linewidth]{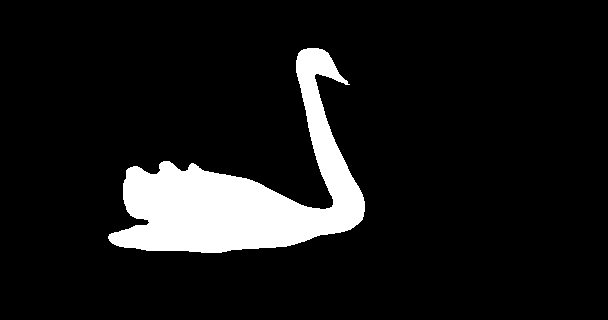} &   
       \includegraphics[width=0.162\linewidth]{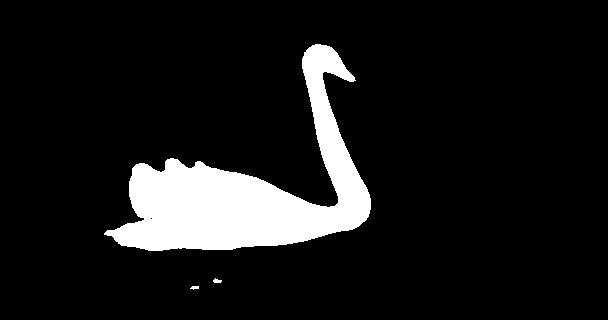} &  
       \includegraphics[width=0.162\linewidth]{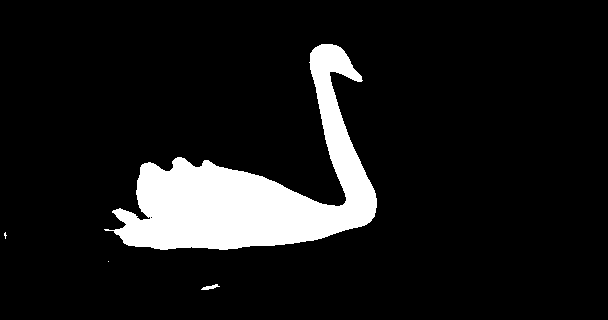} &  
      \includegraphics[width=0.162\linewidth]{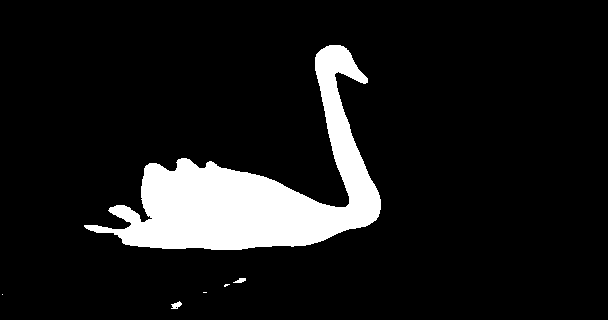} &  
       \includegraphics[width=0.162\linewidth]{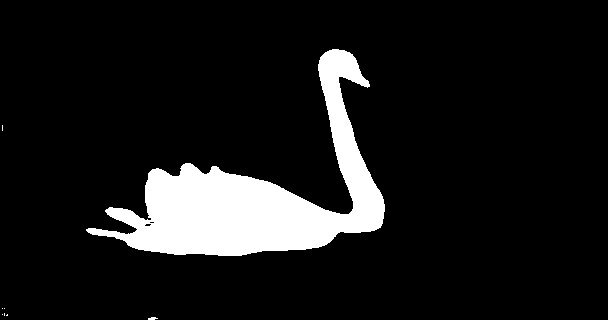} &  
       \includegraphics[width=0.162\linewidth]{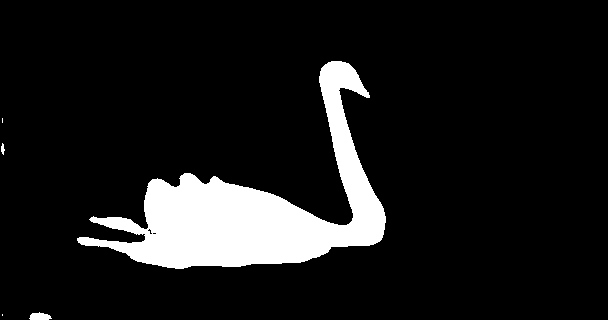}  \\   
    \end{tabular}
   \caption{Predicted masks of our implicit propagation approach. Given the mask of the first frame, we predict the masks of other frames by our method. These predicted masks can then be used for foreground object removal.}
    \label{fig:masks}
\end{figure*}

\textbf{Predicted masks} 
In Fig.~\ref{fig:masks}, we show our results on video mask prediction. Given a single mask of only one frame, our method can propagate it to other frames of the whole sequence automatically. Note that our internal method usually includes slightly more pixels to ensure that all objects' pixels are correctly classified. It performs stably on non-rigid deformation masks and obtains promising results. \\

\begin{figure*}
    \centering
    \begin{tabular}{@{}c @{\hspace{1mm}}c@{}}
         \includegraphics[width=0.49\linewidth]{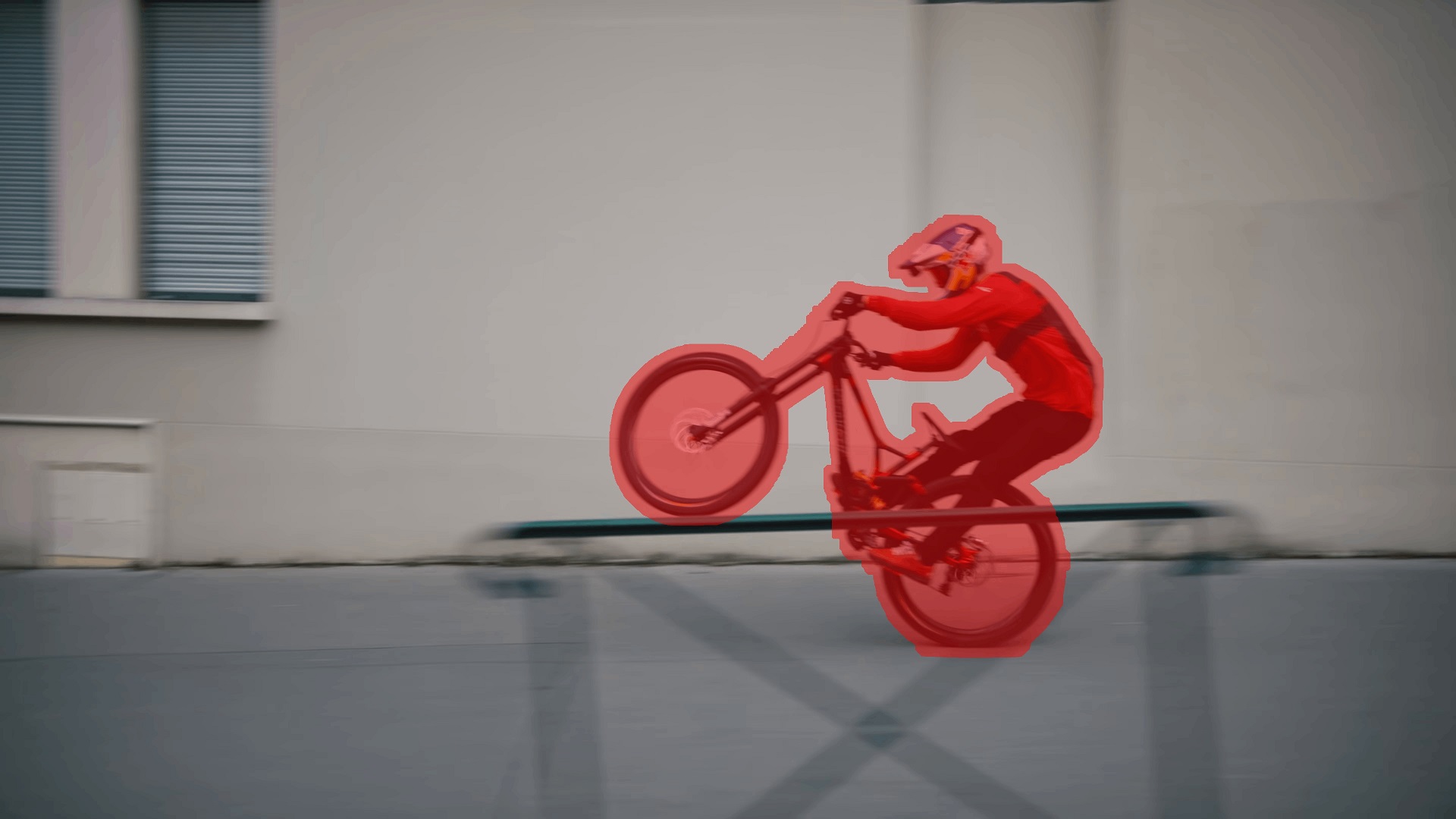} &    
         \includegraphics[width=0.49\linewidth]{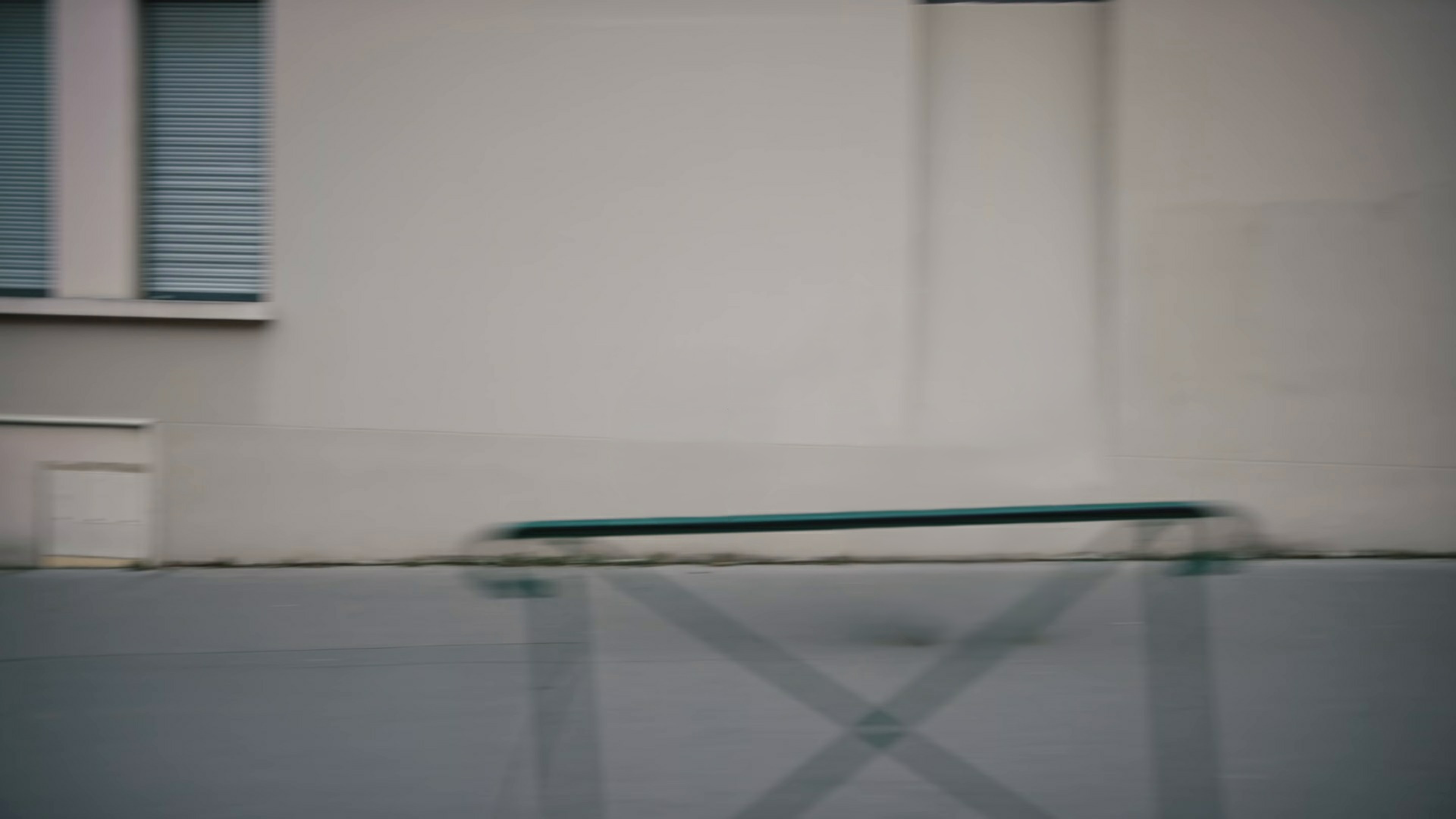}\\
         \includegraphics[width=0.49\linewidth]{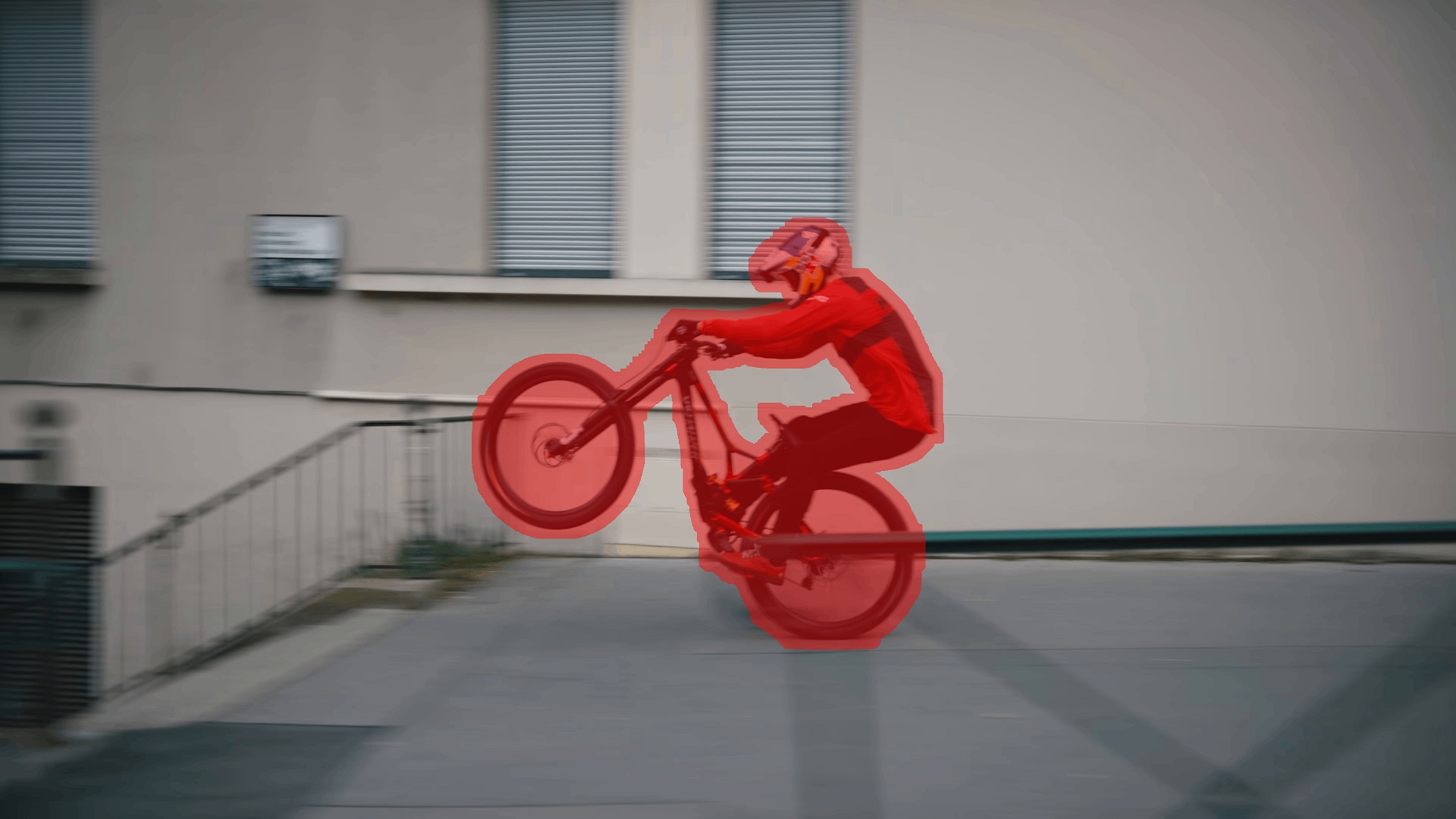} &    
         \includegraphics[width=0.49\linewidth]{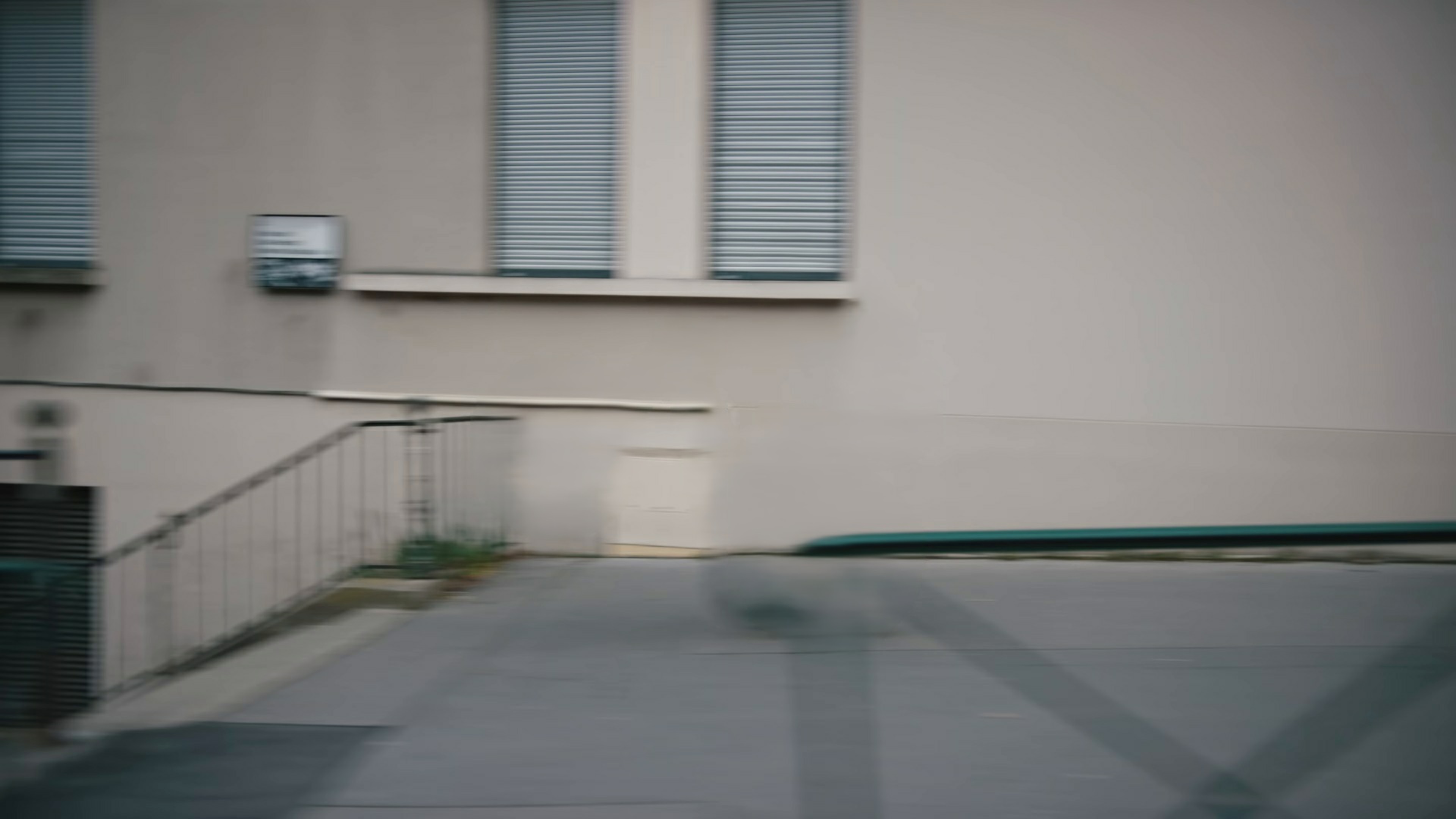}\\
         \includegraphics[width=0.49\linewidth]{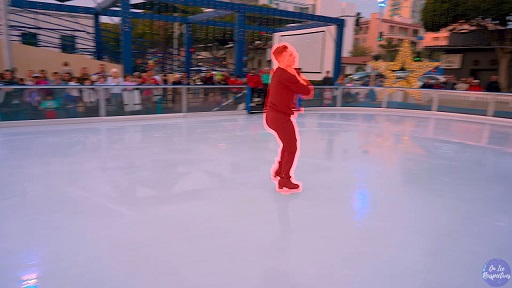} &    
         \includegraphics[width=0.49\linewidth]{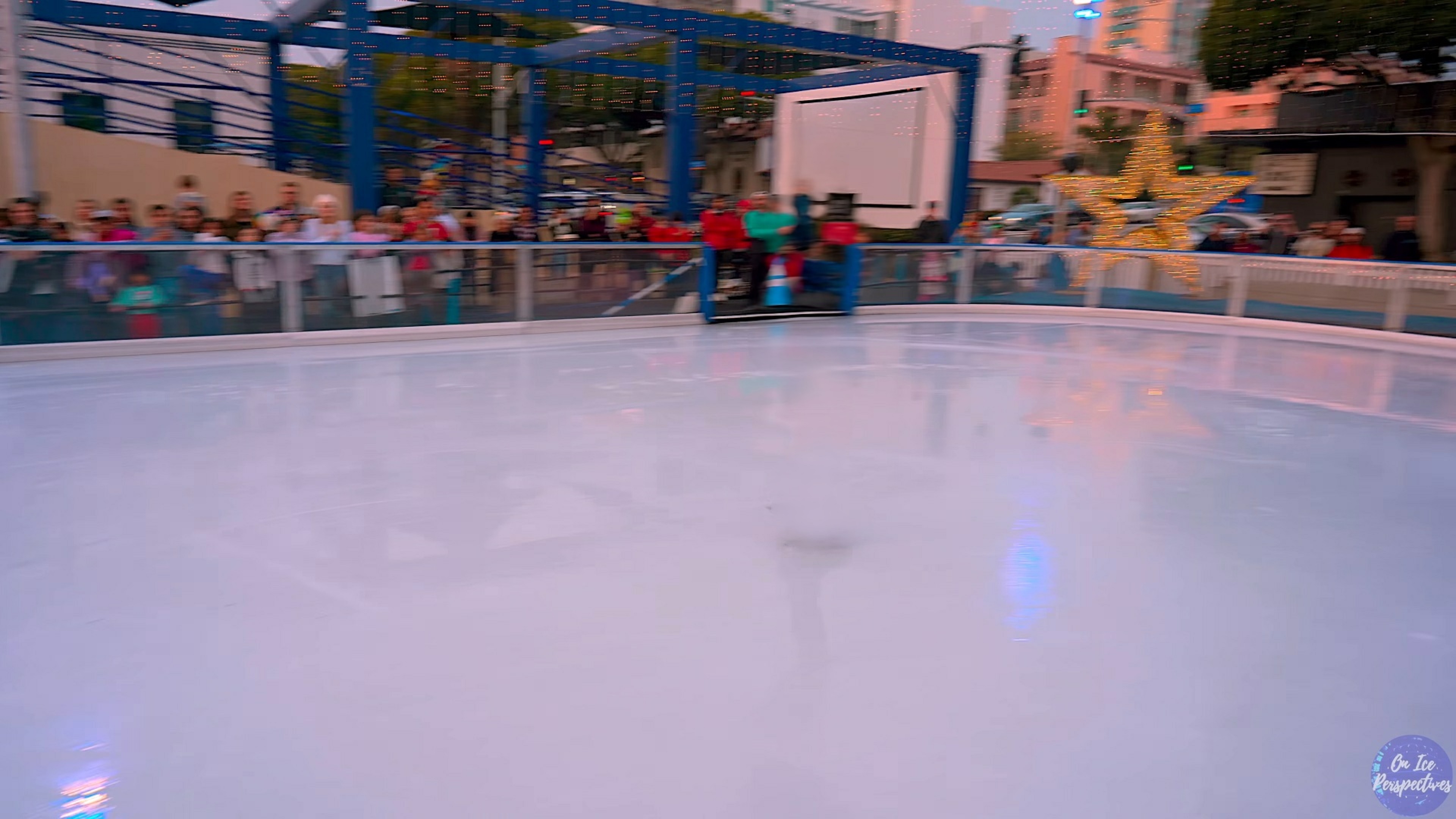}\\
         \includegraphics[width=0.49\linewidth]{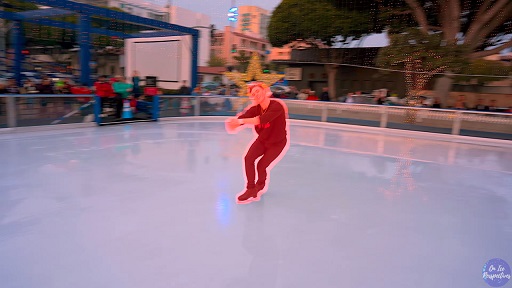} &    
         \includegraphics[width=0.49\linewidth]{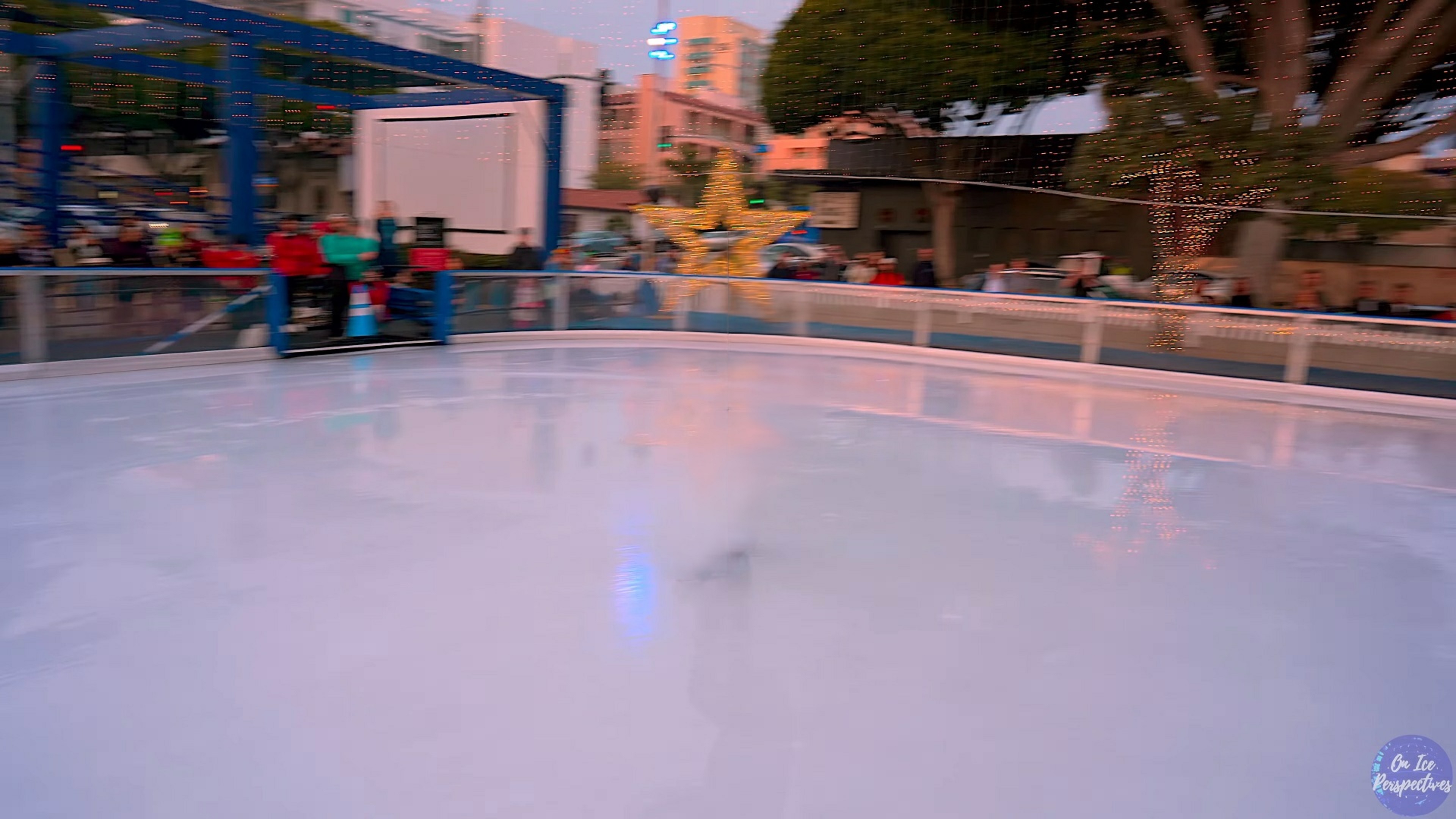}\\         
         Input & Ours
    \end{tabular}
    \caption{Visual results of 4K videos inpainting by our progressive inpainting approach. The size of output images is $3840\times2160$.}
    \label{fig:4k}
\end{figure*}

\textbf{4K videos} 
As shown in Fig.~\ref{fig:4k},  we further extend our approach to a progressive scheme for inpainting 4K videos. To the best of our knowledge, we are the first to perform video inpainting on ultra high-resolution videos.

\section{Discussions}
\textbf{Weights of the regularization terms}
With the default settings, our model generates plausible contents for most of the DAVIS sequences for object removal. However, we also observe that although the anti-ambiguity loss benefits generating high-frequency details, it may risk degrading the temporal consistency. Simultaneously, the stabilization loss tends to reduce the sharpness in the generated regions. How to find the best weights for each video sequence remains to be explored.\\

\textbf{Inference time}
We report the inference time on a 90-frame video sequence. With single NVIDIA RTX 2080Ti, our method takes  5 hours for training and 8.2s for inference. InterVI takes 12 hours for training and inference. FGVC takes 11.6 min for inference. CAP takes 2.0 min for inference. Note that prior approaches typically need days of training on large video   datasets or fine-tuning on pretrained image/video inpainting models.  \\

\begin{figure*}
    \centering 
    \small
    \begin{tabular}{@{}c@{\hspace{0.3mm}}c@{\hspace{0.3mm}}c@{\hspace{0.3mm}}c@{\hspace{0.3mm}}c@{\hspace{0.3mm}}c@{}}
     \includegraphics[width=0.156\linewidth]{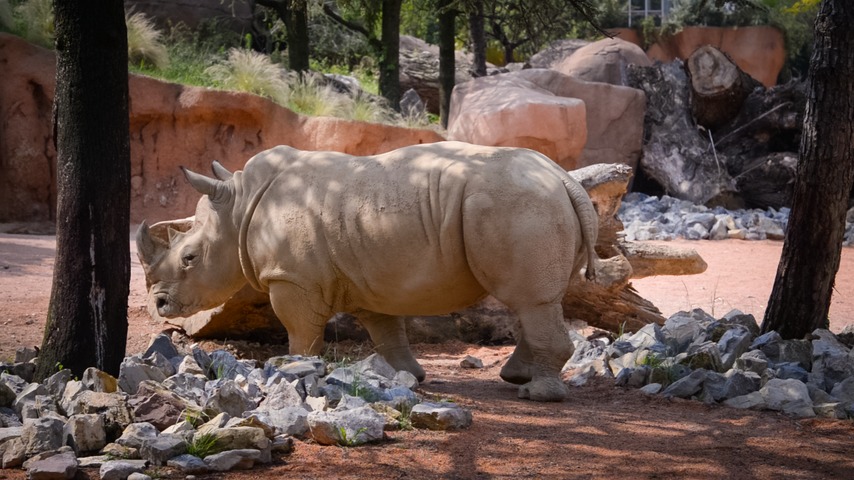} &      
     \includegraphics[width=0.162\linewidth]{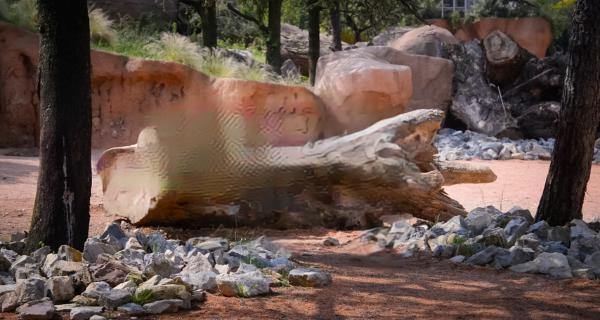} &    
     \includegraphics[width=0.162\linewidth]{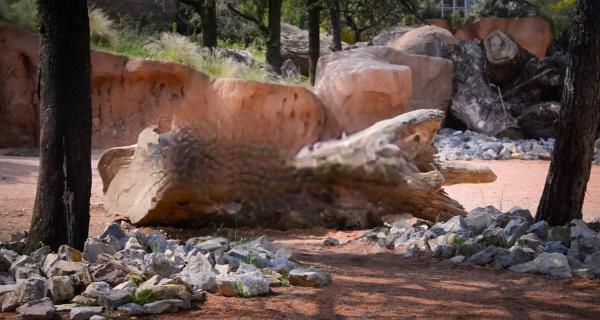} &       
     \includegraphics[width=0.156\linewidth]{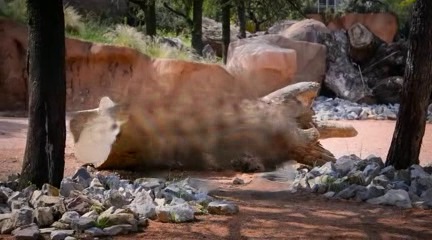} &     
     \includegraphics[width=0.162\linewidth]{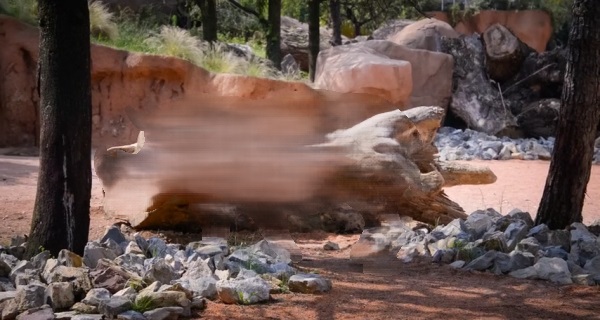}&      
     \includegraphics[width=0.162\linewidth]{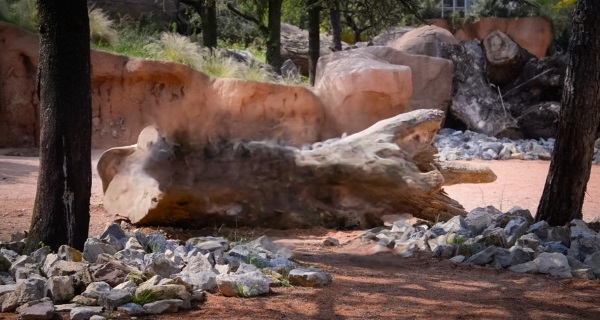} \\

     \includegraphics[width=0.156\linewidth]{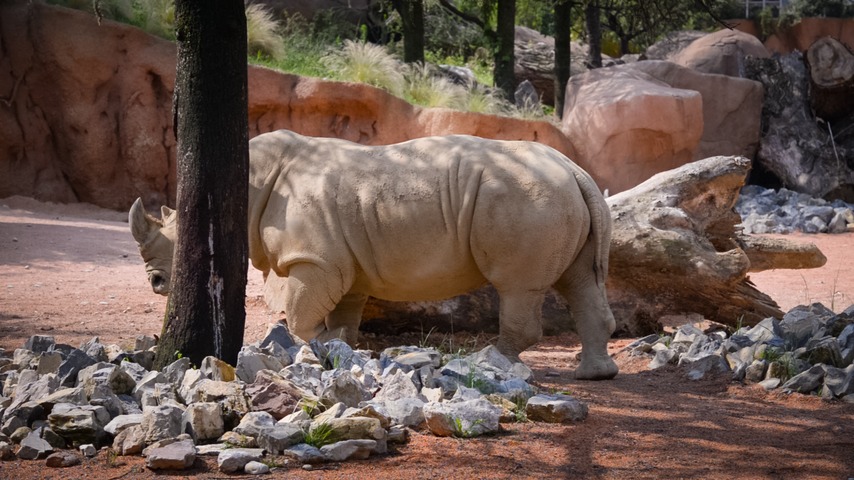} & 
     \includegraphics[width=0.162\linewidth]{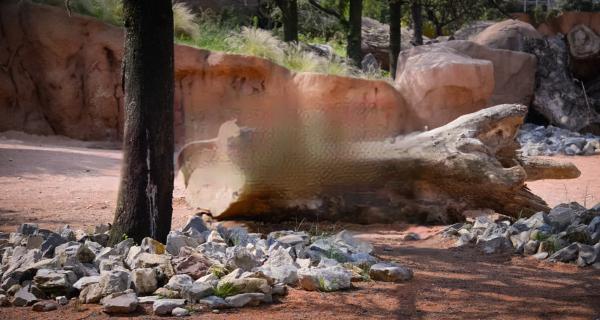} & 
     \includegraphics[width=0.162\linewidth]{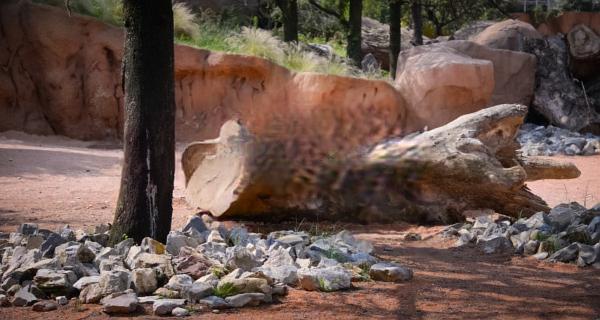} & 
     \includegraphics[width=0.156\linewidth]{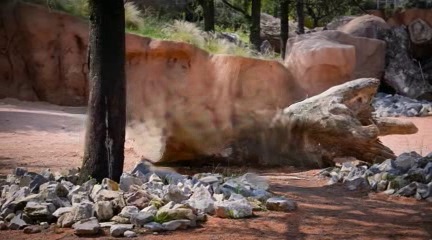} & 
     \includegraphics[width=0.162\linewidth]{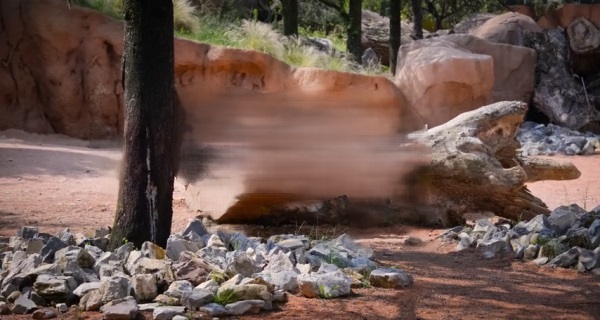} & 
     \includegraphics[width=0.162\linewidth]{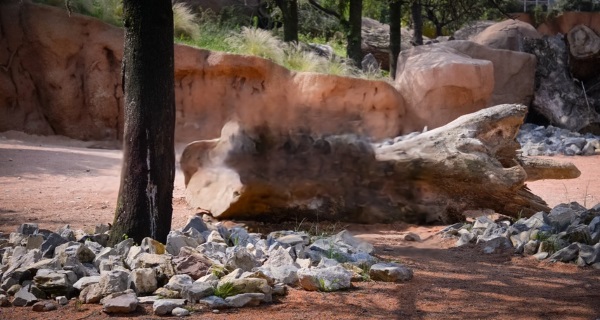} \\
     
     \includegraphics[width=0.156\linewidth]{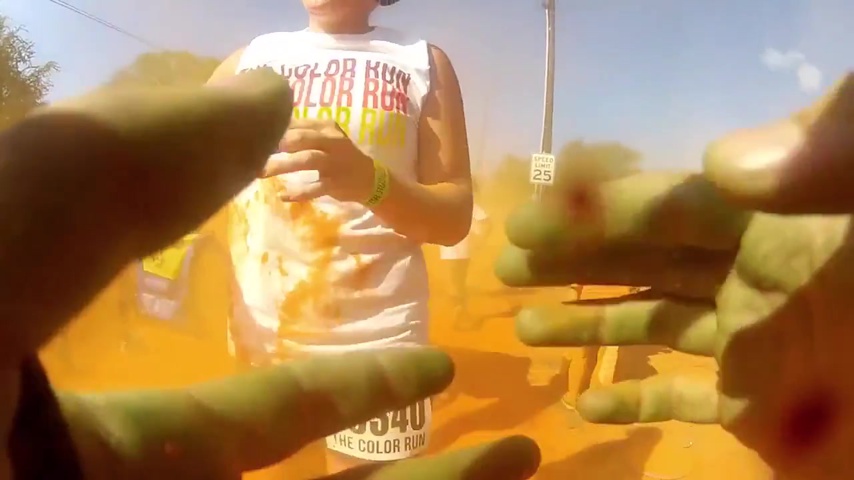} &      
     \includegraphics[width=0.162\linewidth]{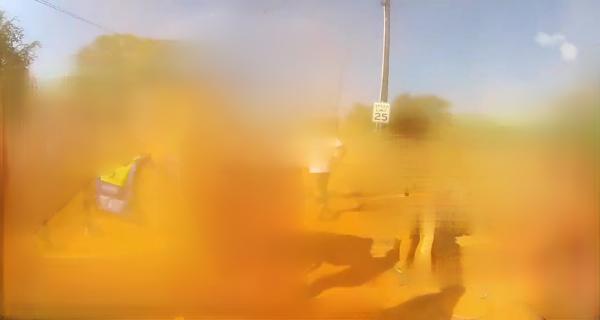} &    
     \includegraphics[width=0.162\linewidth]{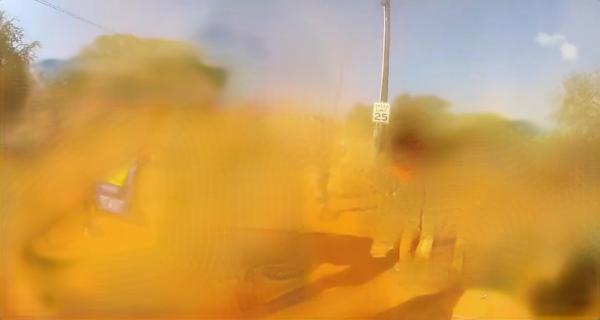} &       
     \includegraphics[width=0.156\linewidth]{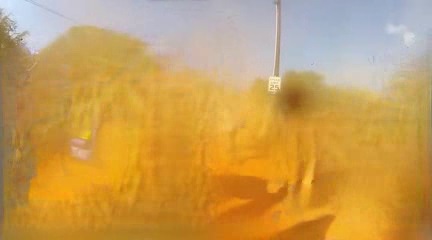} &     
     \includegraphics[width=0.162\linewidth]{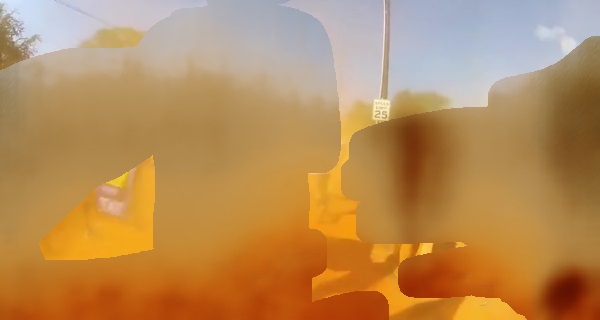}&      
     \includegraphics[width=0.162\linewidth]{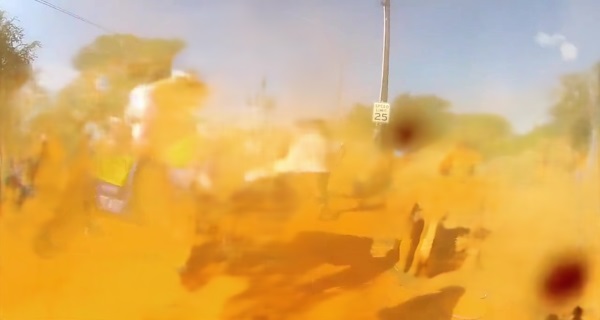} \\

     \includegraphics[width=0.156\linewidth]{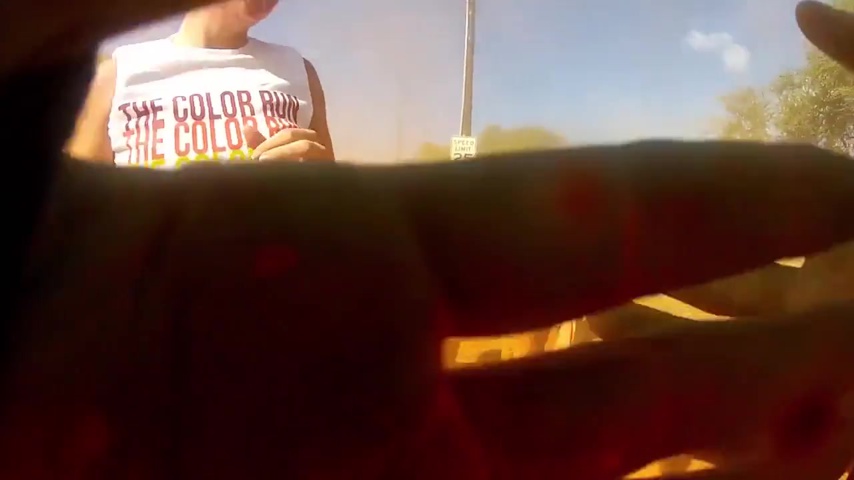} & 
     \includegraphics[width=0.162\linewidth]{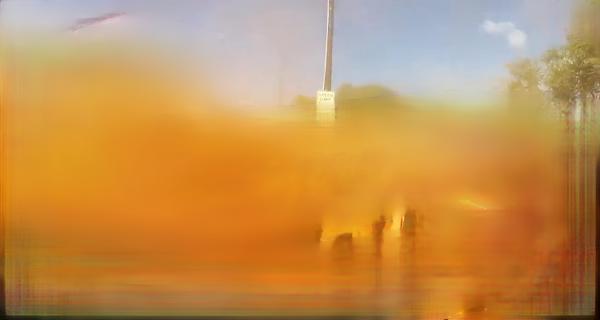} & 
     \includegraphics[width=0.162\linewidth]{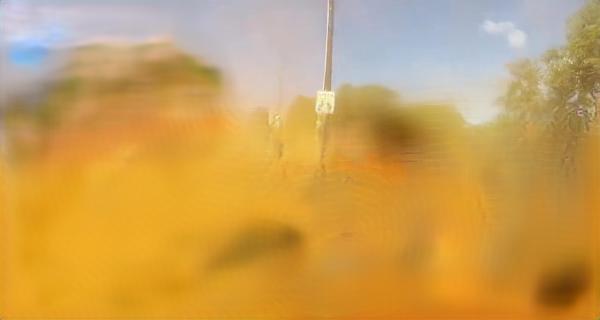} & 
     \includegraphics[width=0.156\linewidth]{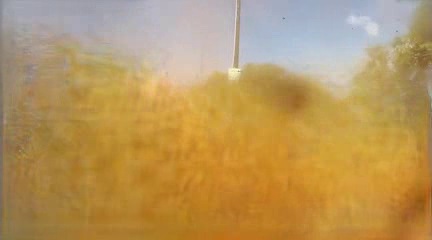} & 
     \includegraphics[width=0.162\linewidth]{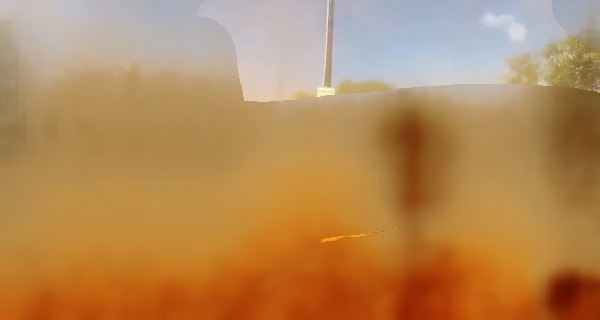} & 
     \includegraphics[width=0.162\linewidth]{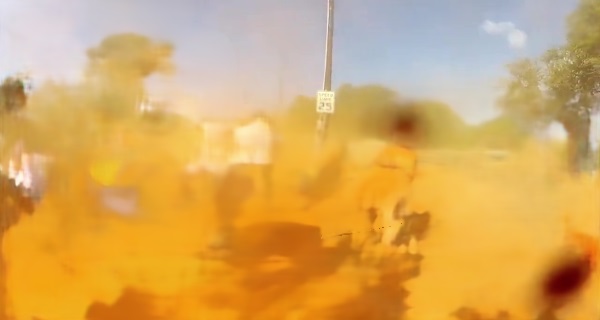} \\     
     Input&CAP &OPN &STTN &InterVI &Ours
    \end{tabular}
   \caption{Failure cases.}
    \label{fig:failure}
\end{figure*}

\textbf{Failure cases}
We demonstrate two failure cases of our method in Fig.~\ref{fig:failure}. In the first example, our model generates semantically incorrect details for the trunk's left part as it is always occluded in all frames. It remains to be solved how to integrate the external natural image priors (e.g., the knowledge of the shape of the trunk) efficiently to our formulation.\\

\end{document}